\documentclass[runningheads]{llncs}
\usepackage{graphicx}
\usepackage{comment}
\usepackage{amsmath,amssymb} % define this before the line numbering.
\usepackage{color}

\usepackage[]{algorithm2e}
\SetAlFnt{\scriptsize}

\usepackage{booktabs}
\usepackage{multirow, makecell}
\usepackage[dvipsnames]{xcolor}
\usepackage{float}
\usepackage{microtype}

\usepackage{hyperref}

\newcommand{\PAR}[1]{\vskip4pt \noindent{\bf #1~}}

\newlength{\tabcolsepdefault}
\DeclareMathOperator*{\argmax}{arg\,max}

\newcommand{\etal}{\textit{et al}.}
\newcommand{\ie}{\textit{i}.\textit{e}.}
\newcommand{\eg}{\textit{e}.\textit{g}.}

\begin{document}
% \linenumbers
\pagestyle{headings}
\mainmatter
\def\ECCVSubNumber{}  % Insert your submission number here

\title{Multi-View Optimization of\\Local Feature Geometry} % Replace with your title

\begin{comment}
\titlerunning{ECCV-20 submission ID \ECCVSubNumber} 
\authorrunning{ECCV-20 submission ID \ECCVSubNumber} 
\author{Anonymous ECCV submission}
\institute{Paper ID \ECCVSubNumber}
\end{comment}
%******************

% \begin{comment}
\titlerunning{Multi-View Optimization of Local Feature Geometry}
\author{Mihai Dusmanu\inst{1} \and
Johannes L. Sch\"onberger\inst{2} \and
Marc Pollefeys\inst{1,2}}
\authorrunning{M. Dusmanu et al.}
\institute{\mbox{Department of Computer Science, ETH Z\"urich \and Microsoft}}
% \end{comment}
%******************
\maketitle

\begin{abstract}
In this work, we address the problem of refining the geometry of local image features from multiple views without known scene or camera geometry. Current approaches to local feature detection are inherently limited in their keypoint localization accuracy because they only operate on a single view. This limitation has a negative impact on downstream tasks such as Structure-from-Motion, where inaccurate keypoints lead to large errors in triangulation and camera localization. Our proposed method naturally complements the traditional feature extraction and matching paradigm. We first estimate local geometric transformations between tentative matches and then optimize the keypoint locations over multiple views jointly according to a non-linear least squares formulation. Throughout a variety of experiments, we show that our method consistently improves the triangulation and camera localization performance for both hand-crafted and learned local features.
\keywords{3D reconstruction, local features}
\end{abstract}

\section{Introduction}

Local image features are one of the central blocks of many computer vision systems with numerous applications ranging from image matching and retrieval to visual localization and mapping.
Predominantly, local feature extraction and matching are the first stages in these systems with high impact on their final performance in terms of accuracy and completeness \cite{Schonberger2017Comparative}.
The main advantages of local features are their robustness, scalability, and efficient matching, thereby enabling large-scale 3D reconstruction~\cite{Heinly2015Reconstructing} and localization~\cite{Li2012Worldwide}.

\begin{figure}
	\centering
	\includegraphics[width=\columnwidth]{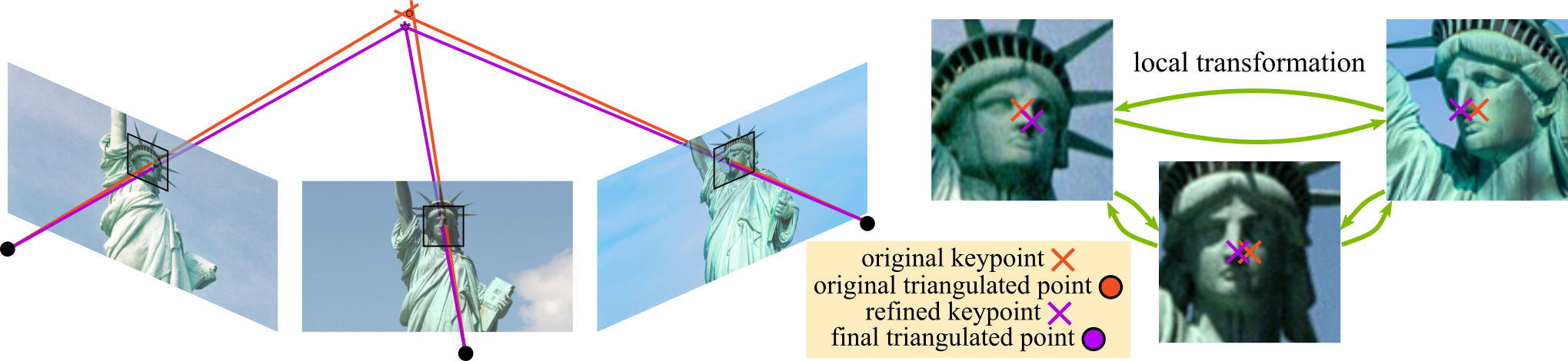}
	\caption{{\bf Multi-view keypoint refinement.} The proposed method estimates local transformations between multiple tentative views of a same feature and uses them to refine the 2D keypoint location, yielding more accurate and complete point clouds.}
	\label{fig:teaser}
\end{figure}

Handcrafted local feature approaches generally focus on low-level structures for detection~\cite{Harris1988Combined,Lowe2004Distinctive}.
Despite the typically accurate keypoint localization of these methods, they are easily perturbed by appearance variations such as day-to-night or seasonal changes, as shown by Sattler \etal~\cite{Sattler2017Benchmarking}.
To achieve a better robustness against viewpoint and appearance changes, recent methods turned to convolutional neural networks (CNNs) for local feature detection and description~\cite{Noh2017Largescale,Detone2018SuperPoint,Dusmanu2019D2,Revaud2019R2D2}.
However, this comes at the cost of a poorer keypoint localization, mainly caused by relying on larger receptive fields and feature map down-sampling through pooling or strided convolutions. 

Moreover, both traditional and CNN-based methods only exploit a single view, as feature detection and description is run independently on each image.
Even for low-level detectors, there is no inherent reason why detections would be consistent across multiple views, especially under strong viewpoint or appearance changes.
While some recent works~\cite{Han2015MatchNet,Hartmann2017Learned,Yao2018MVSNet} consider multiple views to improve the feature matching step, to the best of our knowledge, no prior work exploits multiple views to improve the feature detection stage for more accurate keypoints.

In this paper, we propose a method for optimizing the geometry of local features by exploiting multiple views without any prior knowledge about the camera geometry or scene structure.
Our proposed approach first uses a patch-alignment CNN between tentative matches to obtain an accurate two-view refinement of the feature geometry.
The second stage aggregates all the two-view refinements in a multi-view graph of relative feature geometry constraints and then globally optimizes them jointly to obtain the refined geometry of the features.
The proposed two-stage approach is agnostic to the type of local features and easily integrates into any application relying on local feature matching.
Numerous experiments demonstrate the superior performance of our approach for various local features on the tasks of image matching, triangulation, camera localization, and end-to-end 3D reconstruction from unstructured imagery.
The source code of our entire method and of the evaluation pipeline will be released as open source.

\section{Related work}

\begin{figure}[t]
	\centering
	\includegraphics[width=\textwidth]{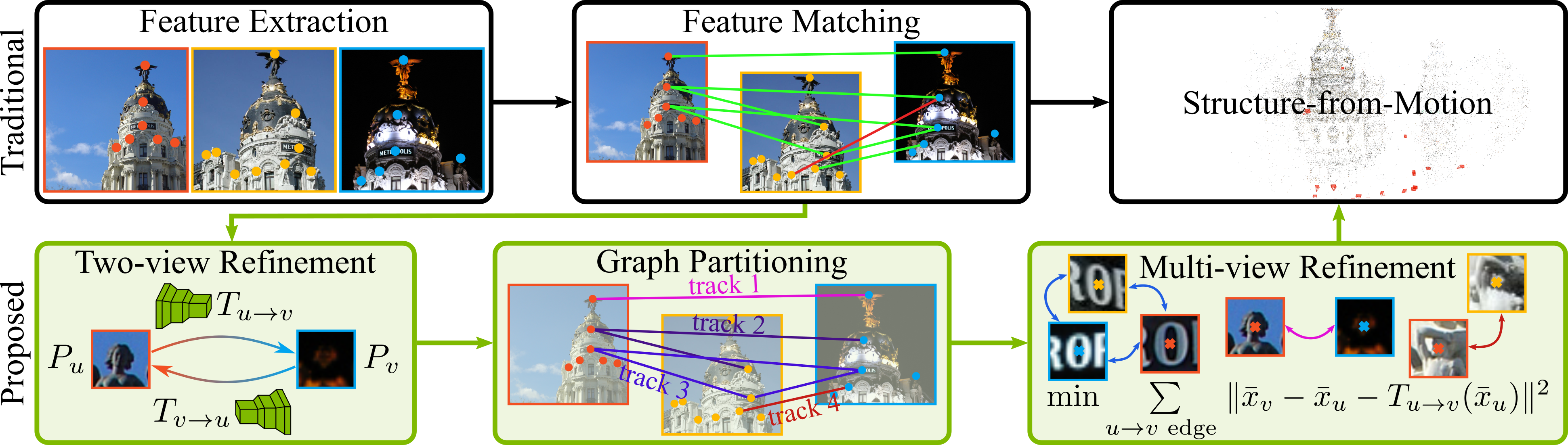}
	\caption{{\bf Overview of the proposed method.} Our method operates on the tentative matches graph (with patches as nodes $P_u, P_v$ and matches as edges) without knowledge of scene and camera geometry. A neural network is used to annotate the edges of this graph with local geometric transformations ($T_{u \rightarrow v}, T_{v \rightarrow u}$). Next, the graph is partitioned into tracks, each track containing at most one patch from each image. Finally, the keypoint locations $x_u, x_v$ are refined using a global optimization over all edges.}
	\label{fig:overview}
\end{figure}

Our method is directly related to local features as well as patch description and matching.
The two-view alignment network borrows concepts from recent advances in the field of image alignment and visual flow.
In this section, we provide an overview of the state of the art in these research directions.

\PAR{Local features.}
Traditional local feature extractors can be split into two main stages: first, feature detection finds interesting regions in the image using low-level statistics (\eg, Difference-of-Gaussians~\cite{Lowe2004Distinctive} or the Harris score~\cite{Harris1988Combined}), typically followed by the estimation of local feature geometry (\eg, scale, orientation, affine shape) for the detected interest points to achieve viewpoint invariance.
Second, feature description then normalizes the local image region around interest points to a canonical frame using the detected feature geometry and finally extracts an illumination invariant, compact numerical representation from the normalized patch (\eg, SIFT~\cite{Lowe2004Distinctive}, Root-SIFT~\cite{Arandjelovic2012Three}, BRIEF~\cite{Calonder2010BRIEF}).
More recently, researchers have developed trainable counterparts that either replace individual parts of the pipeline -- learned detectors~\cite{Verdie2015TILDE,Savinov2017Quad,Barroso-Laguna2019KeyNet} and descriptors~\cite{Balntas2016Learning,Mishchuk2017Working} -- or reformulate the entire pipeline in an end-to-end trainable manner~\cite{Yi2016LIFT,Ono2019LFNet}.

Lately, methods have moved away from the detect-then-describe methodology to a describe-then-detect approach, mainly due to the sensitivity of detections to changes in image statistics.
These methods start by using a CNN as a dense feature extractor and afterwards either train a classifier for detection on top~\cite{Noh2017Largescale}, use it as a shared encoder that splits into two decoders for detection and description respectively~\cite{Detone2018SuperPoint,Revaud2019R2D2}, or directly use non-maxima suppression on the deep feature maps~\cite{Dusmanu2019D2}.
However, these approaches have another issue: due to their large receptive field and feature map down-sampling, the obtained keypoints are generally not well localized when compared to their hand-crafted, low-level counterparts.
This is the case even for methods~\cite{Detone2018SuperPoint} explicitly trained to detect corners.
In this paper, we address the limited accuracy of feature detections for both hand-crafted as well as learned features.
Our approach only requires images as input and achieves superior detection accuracy by considering multiple views jointly, which is in contrast to existing local feature approaches.

\PAR{Patch description and matching.}
CNNs have been successfully used to learn local descriptors offering better robustness to viewpoint and illumination changes using different triplet losses~\cite{Balntas2016Learning}, hard-negative mining techniques~\cite{Mishchuk2017Working}, and geometric similarity for training~\cite{Luo2018Geodesc}.
Likewise, in Multi-View Stereo, hand-crafted similarity metrics~\cite{Goesele2006Multi} and descriptors~\cite{Tola2009Daisy} traditionally used for patch matching were replaced by learned counterparts~\cite{Luo2016Efficient,Han2015MatchNet,Hartmann2017Learned,Yao2018MVSNet}.
Closer to our approach are methods bypassing description and directly considering multiple views to decide whether two points correspond~\cite{Zagoruyko2015Learning,Zbontar2016Stereo,Han2015MatchNet}.
While these approaches focus on the second part of the local feature pipeline, we focus our attention on the detection stage.
However, the intrinsic motivation is the same: exploiting multiple views facilitates a more informed decision for better results.

\PAR{Geometric alignment and visual flow.}
Recent advances in semantic alignment~\cite{Rocco2017Convolutional,Rocco2018Neighbourhood} and image matching~\cite{Rocco2018Neighbourhood} as well as flow estimation~\cite{Dosovitskiy2015FlowNet} use a Siamese network followed by a feature matching layer.
Our patch alignment network uses the correlation normalization introduced in~\cite{Rocco2017Convolutional}. The matching results are processed by a sequence of convolutional and fully connected layers for prediction.
Contrary to visual flow, which is generally targeted at temporally adjacent video frames, where pixel displacements remain relatively low and appearance is similar, our method must handle large deformations and drastic illumination changes.

\PAR{Refinement from known geometry or poses.}
Closer to our method, Eichhardt \etal~\cite{Eichhardt2019Optimal} recently introduced an approach for local affine frame refinement. While they similarly formulate the problem as a constrained, multi-view least squares optimization, their method assumes known two-view camera geometries and does not consider visual cues from two views jointly to compute the patch alignment. Furthermore, they need access to ground-truth feature tracks (computed by an initial Structure-from-Motion process). In contrast, not requiring known camera geometry and feature tracks makes our approach amenable to a much wider range of practical applications, \eg, Structure-from-Motion or visual localization. Moreover, the two methods are in fact complementary -- our procedure can improve the quality of Structure-from-Motion, which can then be further refined using their approach.

\section{Method}

The generic pipeline for multi-view geometry estimation, illustrated in Figure~\ref{fig:overview}, starts from a set of input images $\mathcal{I} = \{I_1, \dots, I_N\}$ and first runs feature extraction on each image $I_i$ independently yielding keypoints $p_i$ with associated local descriptors $d_i$.
Feature matching next computes tentative feature correspondences $\mathcal{M}_{i, j} = \{(k, l)$ such that $d_{i, k}$ matches $d_{j, l} \}$ between image pairs $(I_i, I_j)$ based on nearest neighbors search in descriptor space (usually alongside filtering techniques).
The output of this step can be interpreted as a tentative matches graph $G = (V, E)$ with keypoints as nodes ($V = \cup_i p_i$) and matches as edges ($E = \cup_{i, j} \mathcal{M}_{i, j}$), optionally weighted (\eg, by the cosine similarity of descriptors).
In the last step, the specific application (\eg, a Structure-from-Motion~\cite{Schoenberger2016Structure} or visual localization pipeline~\cite{Sattler2011Fast}) takes the tentative matches graph as input and estimates camera or scene geometry as the final output. 

In this paper, we propose a further geometric refinement of the nodes $V$ in the tentative matches graph, as shown in the bottom part of Figure~\ref{fig:overview}.
This intermediate processing step naturally fits into any generic multi-view geometry pipeline.
As demonstrated in experiments, our method significantly improves the geometric accuracy of the keypoints and thereby also  the later processing steps, such as triangulation and camera pose estimation.

\subsection{Overview}

Our proposed method operates in a two-stage approach.
First, for each edge, we perform a two-view refinement using a patch alignment network that, given local patches $P_u, P_v$ around the corresponding initial keypoint locations $u, v \in \mathbb{R}^2$, predicts the flow $d_{u \rightarrow v}$ of the central pixel from one patch in the other and vice versa as $d_{v \rightarrow u}$.
This network is used to annotate the edges of the tentative matches graph with geometric transformations $T_{u \rightarrow v}$, $T_{v \rightarrow u}$.
In the second step, we partition the graph into components (\ie, features tracks) and find a global consensus by optimizing a non-linear least squares problem over the keypoint locations, given the estimated two-view transformations.

\begin{figure}[t]
	\centering
	\begin{minipage}{0.45\textwidth}
		\includegraphics[width=1.0\columnwidth]{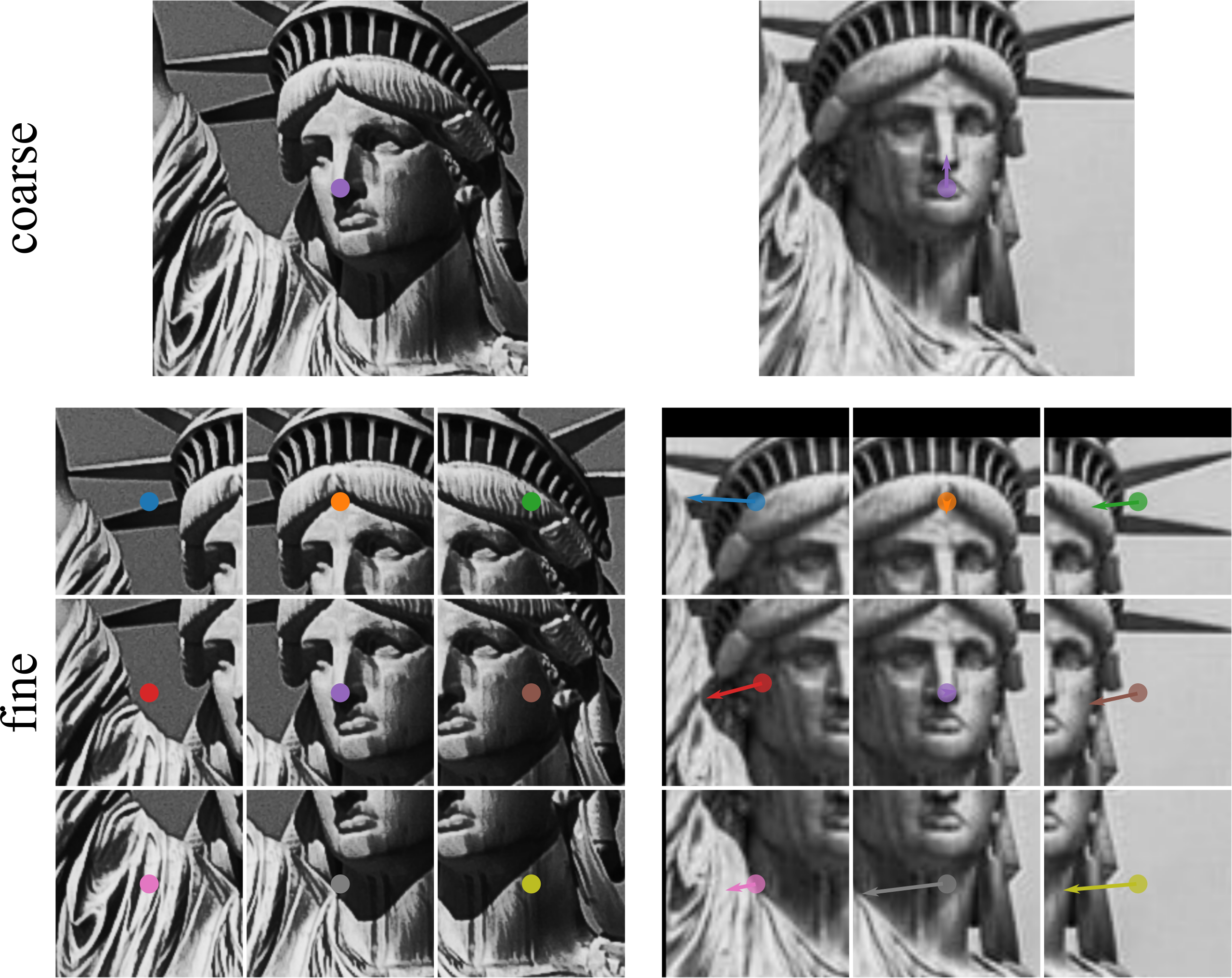}
	\end{minipage}
	~
	\begin{minipage}{0.45\textwidth}
		\begin{tabular}{c c c}
			\multirow{2}{*}{\bf{Source}} & \multirowcell{2}{Warped\\Source $\leftarrow$ Target} & \multirow{2}{*}{\bf{Target}} \\ 
			& & \\
			\includegraphics[width=0.275\columnwidth]{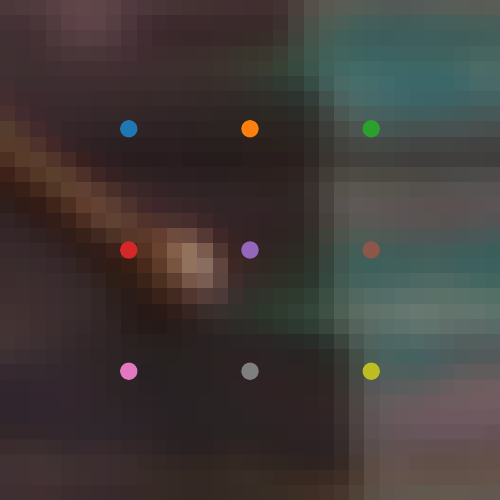} & \includegraphics[width=0.275\columnwidth]{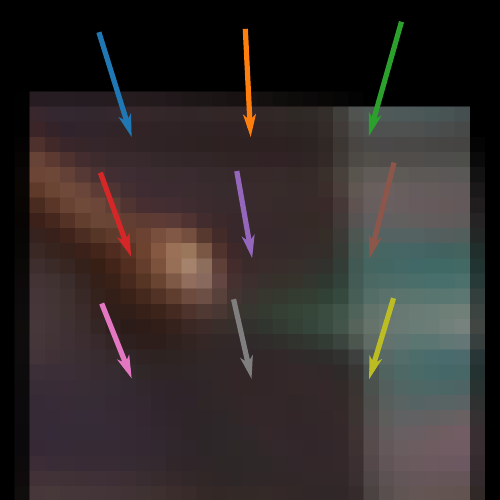} & \includegraphics[width=0.275\columnwidth]{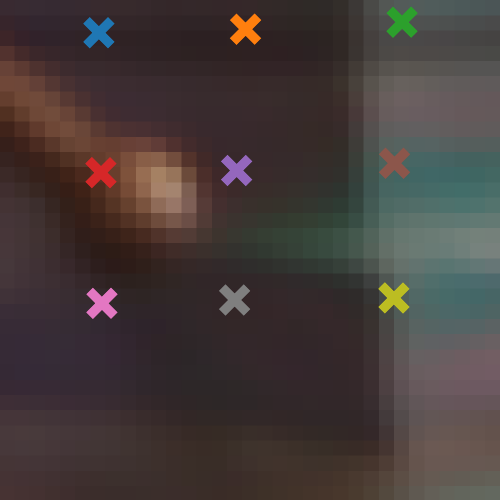} \\
			\includegraphics[width=0.275\columnwidth]{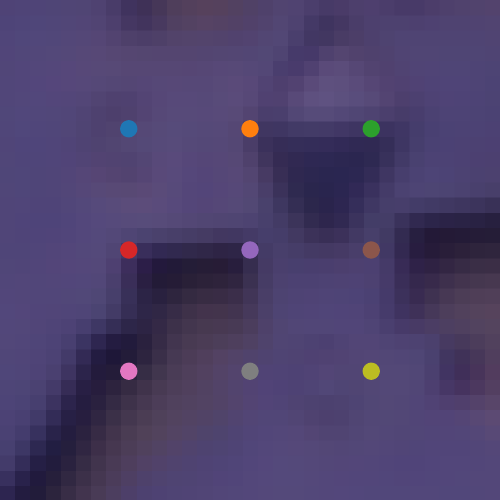} & \includegraphics[width=0.275\columnwidth]{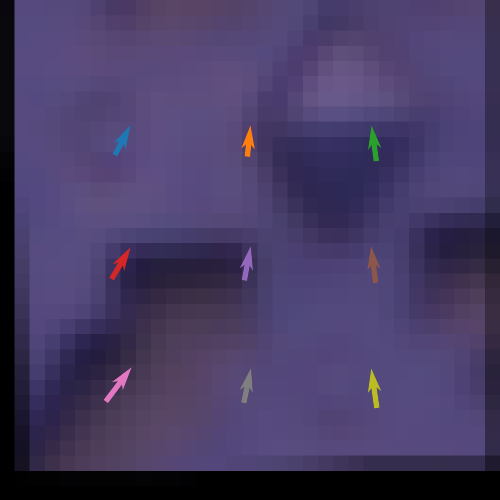} & \includegraphics[width=0.275\columnwidth]{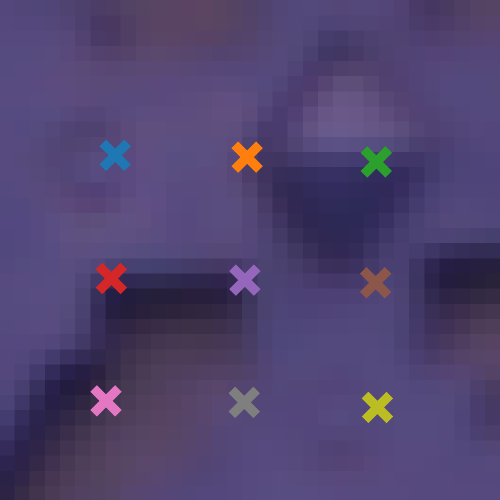} \\
			\includegraphics[width=0.275\columnwidth]{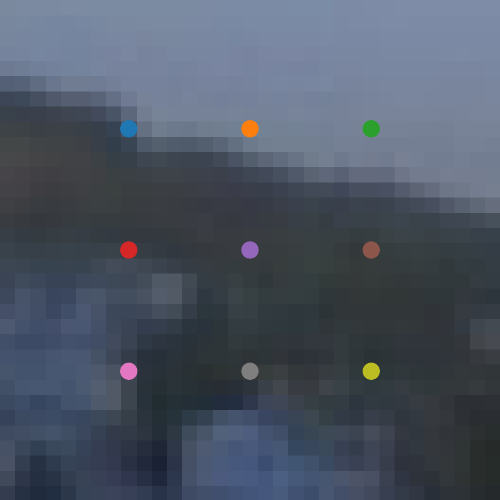} & \includegraphics[width=0.275\columnwidth]{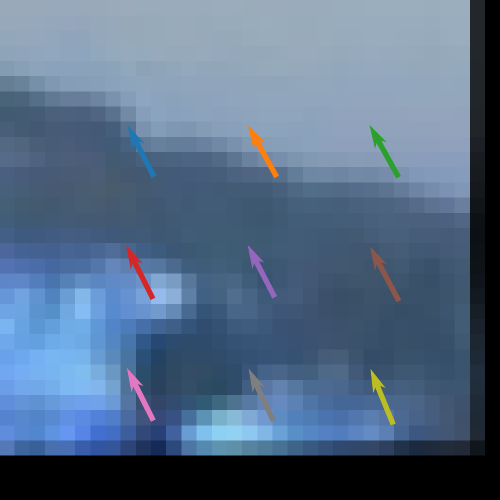} & \includegraphics[width=0.275\columnwidth]{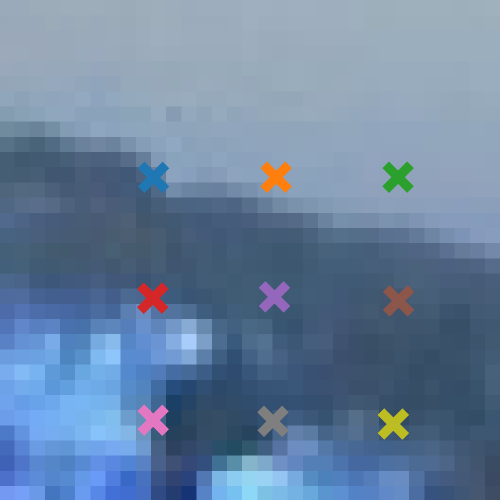} \\
		\end{tabular}
	\end{minipage}
	\caption{{\bf Coarse-to-fine refinement and qualitative examples.} \textit{Left:} We start by a coarse alignment at feature extraction resolution taking into account only the central point, followed by a fine refinement on sub-patches corresponding to each grid point. \textit{Right:} The first and last columns show the source and the target patch, respectively. The $3 \times 3$ regular grid is plotted with circles. For the target patch, we plot the deformed grid predicted by the coarse-to-fine refinement with crosses. The middle column shows the warped target patch using bisquare interpolation in between grid locations.}
	\label{fig:coarse_to_fine}\label{fig:grid_qualitative}
\end{figure}

\subsection{Two-view refinement}
Our method starts by computing a two-view refinement for every edge in the graph.
Similarly to previous works in the field of CNNs for semantic alignment \cite{Rocco2017Convolutional,Rocco2018End}, image matching \cite{Rocco2018Neighbourhood}, and visual flow \cite{Dosovitskiy2015FlowNet}, we employ a Siamese architecture for feature extraction followed by a correlation layer. The final flow is predicted by a succession of convolutional and fully connected layers.

\PAR{Feature extraction and correlation.}
The architecture first densely extracts features in both patches ($P_u$, $P_v$) with a standard CNN architecture.
The output is two 3D tensors $F_u, F_v \in \mathbb{R}^{h \times w \times d}$, each of which can be interpreted as a set of $d$-dimensional descriptors associated to a $h \times w$ spatial grid in their corresponding patches $\mathbf{f}_u(i, j), \mathbf{f}_v(i, j) \in \mathbb{R}^d$. Before matching the descriptors using dot-product correlation, we perform L2-normalization as $\hat{\mathbf{f}}(i, j) = \frac{\mathbf{f}(i, j)}{\lVert \mathbf{f}(i, j) \rVert_2}$.

Dense matching can be implemented using a correlation layer yielding a 4D tensor $c \in \mathbb{R}^{h \times w \times h \times w}$ defined by $c(i_1, j_1, i_2, j_2) = \hat{\mathbf{f}}_u(i_1, j_1)^T \hat{\mathbf{f}}_v(i_2, j_2)$.
This volume can be interpreted as a 3D tensor $m \in \mathbb{R}^{h \times w \times (h \cdot w)}$, where each channel is associated to a different grid position in the opposite patch: $m(i_1, j_1)_k = c(i_1, j_1, i_2, j_2)$ where $k = i_2 \cdot w + j_2$.

Following the methodology proposed by \cite{Rocco2017Convolutional}, we use L2-normalization across the channel dimension to lower the values of ambiguous matches
\begin{equation}
	\hat{m}(i, j) = \frac{\text{ReLU}(m(i, j))}{\lVert \text{ReLU}(m(i, j)) \rVert_2} \enspace ,
\end{equation}
when the opposite patch contains more than one similar descriptor.

\PAR{Regression.}
The final matching result $\hat{m}$ is post-processed by a CNN to aggregate local information.
Finally, to enforce a patch-level consistency, a sequence of fully connected layers predicts the final output $d_{u \rightarrow v}$.
Please refer to the supplementary material for more details regarding the architecture.

\subsection{Multi-view refinement}

In a two-view scenario, the network described in the previous section is sufficient: ($u$ and $v + d_{u \rightarrow v}$) or ($u + d_{v \rightarrow u}$ and $v$) can directly be used as the refined keypoint locations.
However, given that our final goal is to perform optimization over multiple views, there are several challenges we need to overcome.

Firstly, since corresponding features are generally observed from different viewpoints and looking at non-planar scene structures, the computed displacement vector is only valid for the central pixel and not constant within the patch (\ie, $\frac{\delta}{\delta u} d_{u \rightarrow v} \neq \mathbf{0}_{2, 2}$).
Thus, when refining keypoint locations $u, v, w, \ldots$ over multiple views, consistent results can only be produced by forming displacement chains (\eg, $d_{u \rightarrow v} + d_{(v + d_{u \rightarrow v}) \rightarrow w} + \ldots$) without loops.
However, such an approach does not consider all possible edges in the graph and quickly accumulate errors along the chain.
Another possible way to perform the refinement is to predict new displacements every time the keypoint locations are updated during the multi-view optimization.
The main downside of this approach is its run-time, since the two-view network would have to be run for each edge after each optimization step.
Therefore, to refine the keypoints over the entire graph and also achieve practical run-times, we use the two-view network to estimate local flow fields $T_{u \rightarrow v}$ prior to multi-view refinement and then efficiently interpolate displacements within the patch during the optimization. Some qualitative examples are shown in Figure~\ref{fig:grid_qualitative}~(right).

Secondly, the connected components of $G$ generally contain feature tracks of different scene points, as the graph topology is purely based on appearance and feature matching is imperfect despite various filtering constraints -- a single incorrect match can merge two tracks.
As such, we partition the connected components into smaller, more reliable subsets based on the descriptor cosine similarity $s_{u, v}$ between patch pairs $(u, v)$.

Thirdly, predicting the reverse flow or loops in the graph does not necessarily produce a consistent result (\eg, $T_{v \rightarrow u} \circ T_{u \rightarrow v} \neq \textbf{id}$, $T_{w \rightarrow u} \circ T_{v \rightarrow w} \circ T_{u \rightarrow v} \neq \textbf{id}$) due to wrong matches or noisy network predictions.
We tackle this by formulating a joint robust optimization of all tentatively matching keypoint locations considering all the edges over multiple views, analogous to Pose Graph Optimization~\cite{Olson2006Fast}.
In the following paragraphs, we detail our solutions to the issues mentioned above.

\PAR{Flow field prediction.}
To facilitate the multi-view optimization of the keypoint locations, we use repeated forward passes of the central flow network to predict a local flow field $T_{u \rightarrow v}$ around the initial keypoint location $u$.
Note that this prediction is directionally biased and, as such, we always also predict the inverse flow field $T_{v \rightarrow u}$.
For further space and time efficiency considerations, we approximate the full flow field between two patches by a $3\times3$ displacement grid and use bi-square interpolation with replicate padding in between the grid points.
Assuming locally smooth flow fields, we can efficiently chain the transformations from any node $u$ to another node $w$ without any additional forward-passes of the two-view network.
To obtain correspondences for all points of the $3\times3$ grid, we first predict a coarse alignment $d_{u \rightarrow v}^c$ using patches around matched features $u, v$ at original keypoint extraction resolution. Subsequently, we further refine the coarse flow at a finer resolution using sub-patches around each $3\times3$ grid position $g$, $d_{u + g \rightarrow v + d_{u \rightarrow v}^c + g}^f$. The final transformation is given by: $T_{u \rightarrow v}(g) = d_{u \rightarrow v}^c + d_{u + g \rightarrow v + d_{u \rightarrow v}^c + g}^f$.
This process is illustrated in Figure~\ref{fig:coarse_to_fine}~(left).

\PAR{Match graph partitioning.}
To address the second issue, our multi-view refinement starts by partitioning the tentative matches graph into disjoint components called tracks.
A track is defined as a subset of the nodes $V$ containing at most one node (patch) from each image.
This is similar to a 3D feature track (\ie, the set of 2D keypoints corresponding to the same 3D point).
For each node $u \in V$, we denote $t_u$ the track containing $u$. For a subset $S$ of $V$, we define $\mathbf{I}_S$ as the set of images in which the features (nodes) of $S$ were extracted (\ie, $\mathbf{I}_S = \{I \in \mathcal{I} | \exists u \in S \text{ s.t. } u \in I \}$).

The proposed algorithm for track separation follows a greedy strategy and is closely related to Kruskal's minimum-spanning-tree algorithm~\cite{Kruskal1956On}.
The edges $(u \rightarrow v) \in E$ are processed in decreasing order of their descriptor similarity $s_{u \rightarrow v}$.
Given an edge $u \rightarrow v$ linking two nodes from different tracks (\ie,  $t_u \neq t_v$), the two tracks are joined only if their patches come from different images (\ie, $\mathbf{I}_{t_u} \cap \mathbf{I}_{t_v} = \O$).
The pseudo-code of this algorithm is defined in Figure~\ref{alg:track}~(left).

Another challenge commonly arising due to repetitive scene structures are very large connected components in the tentative matches graph.
These large components are generally caused by a small number of low-similarity edges and lead to excessively large optimization problems.
To prevent these large components from slowing down the optimization, we use recursive normalized graph-cuts (GC) on the meta-graph of tracks $\mathcal{G} = (\mathcal{V}, \mathcal{E})$ until each remaining connected component has fewer nodes than the number of images $N$.
The nodes of $\mathcal{G}$ correspond to tracks ($\mathcal{V} = \{t_u | u \in V\}$) and its edges aggregate over the edges of $G$, $\mathcal{E} = \{(t_u, t_v, w_{t_u, t_v}) | (u \rightarrow v) \in E, w_{t_u, t_v} = \sum_{(u^\prime \rightarrow v^\prime) \in E \text{ s.t. } t_{u^\prime} = t_u, t_{v^\prime} = t_v} s_{u^\prime \rightarrow v^\prime}\}$.
The $G$-cardinality of a subset $\mathcal{A} \subseteq \mathcal{V}$ is defined as: $\lvert \mathcal{A} \rvert_G = \lvert \{u \in V | t_u \in \mathcal{A}\} \rvert$.
The pseudo-code is detailed in Figure~\ref{alg:graph-cut}~(right). This step returns a pair-wise disjoint family of sets $\mathcal{S}$ corresponding to the final connected components of $\mathcal{G}$.

Given the track assignments and a set of tracks $\mathcal{A} \in \mathcal{S}$, we define the set of intra-edges connecting nodes within a track as $E_\text{intra}^\mathcal{A} = \{ (u \rightarrow v) \in E | t_u = t_v, t_u \in \mathcal{A} \}$ and the set of inter-edges connecting nodes of different tracks as $E_\text{inter}^\mathcal{A} = \{ (u \rightarrow v) \in E | t_u \neq t_v, t_u \in \mathcal{A}, t_v \in \mathcal{A} \}$.
In the subsequent optimization step, the intra-edges are considered more reliable and prioritized, since they correspond to more confident matches.

\begin{figure}[t]
	\centering
	\begin{minipage}[c]{0.48\textwidth}
		\centering
		\begin{algorithm}[H]
			\KwIn{Graph $G = (V, E)$}
			\KwOut{Track assignments $t_u, \forall u \in V$}
			\BlankLine
			\For{$u \in V$} {
				$t_u \leftarrow$ new track $\{u\}$\;
			}
			$F \leftarrow E$ sorted by decreasing similarity\;
			\For{$(u, v) \in F$}{
				\If{$\mathbf{I}_{t_u} \cap \mathbf{I}_{t_v} = \O$}{
					merge $t_u$ and $t_v$\;
				}
			}
		\end{algorithm}
	\end{minipage}
	~
	\begin{minipage}[c]{0.48\textwidth}
		\centering
		\begin{algorithm}[H]
			\KwIn{Meta-graph $\mathcal{G} = (\mathcal{V}, \mathcal{E})$}
			\KwOut{Family of sets $\mathcal{S}$}
			\SetKwProg{Fn}{Function}{}{}
			\BlankLine
			$\mathcal{S} \leftarrow \{\}$\;
			\For{$\mathcal{C}$ connected component of $\mathcal{G}$}{
				RecursiveGraphCut($\mathcal{C}$)\;
			}
			\Fn{RecursiveGraphCut($\mathcal{C}$)}{
				\uIf{$\lvert \mathcal{C} \rvert_G > N$}{
					$\mathcal{A}, \mathcal{B} \leftarrow$ NormalizedGC($\mathcal{C}$)\;
					RecursiveGraphCut($\mathcal{A}$)\;
					RecursiveGraphCut($\mathcal{B}$)\;
				}\Else{
					$\mathcal{S} \leftarrow \mathcal{S} \cup \{\mathcal{C}\}$;
				}
			}
		\end{algorithm}
	\end{minipage}
	\caption{{\bf Algorithms.} \textit{Left -- track separation algorithm:} the tentative matches graph is partitioned into tracks following a greedy strategy. Each track contains at most one patch from each image. \textit{Right -- recursive graph cut:} we remove edges until having connected components of size at most $N$ - the number of images. This algorithm yields a pair-wise disjoint family of sets $\mathcal{S}$, each set representing an ensemble of tracks.}
	\label{alg:track}\label{alg:graph-cut}
\end{figure}

\PAR{Graph optimization.}
Given the tentative matches graph augmented by differentiable flow fields $T$ for all edges, the problem of optimizing the keypoint locations $x_p$ can be formulated independently for each set of tracks $\mathcal{A} \in \mathcal{S}$ as the bounded non-linear least squares problem
\begin{equation}
	\begin{split}
		\min_{\{x_p | t_p \in \mathcal{A}\}} & \sum_{(u \rightarrow v) \in E_\text{intra}^\mathcal{A}} s_{u \rightarrow v} \rho(\lVert \bar{x}_v - \bar{x}_u - T_{u \rightarrow v}(\bar{x}_u) \rVert^2) + \\
		& \sum_{(u \rightarrow v) \in E_\text{inter}^\mathcal{A}} s_{u \rightarrow v} \psi(\lVert \bar{x}_v - \bar{x}_u - T_{u \rightarrow v}(\bar{x}_u) \rVert^2) \\
		\text{s.t.} & \lVert \bar{x}_p \rVert_1 = \lVert x_p - x_p^0 \rVert_1 \leq K, \forall p \enspace ,
	\end{split}
	\label{eq:least_squares}
\end{equation}
where $x_\cdot^0$ are the initial keypoint locations, $\rho$ is a soft, unbounded robust function for intra-edges, $\psi$ is a stronger, bounded robust function for inter-edges, and $K$ is the degree of liberty of each keypoint (in pixels).
Finally, $s_{u \rightarrow v}$ is the cosine similarity between descriptors of nodes $u$ and $v$; thus, closer matches in descriptor space are given more confidence during the optimization.

The inter-edges are essential since most features detectors in the literature sometimes fire multiple times for the same visual feature despite non-max suppression (at multiple scales or with different orientations).
Without inter-edges, given our definition of a track as only containing at most one feature from each image,  these detections would be optimized separately.
With inter-edges, the optimization can merge different tracks for higher estimation redundancy if the deviations from the intra-track solutions are not too high.

Note that this problem can have multiple local minima corresponding to different scene points observed in all the patches of a track.
For robust convergence of the optimization to a good local minimum, we fix the keypoint location of the node $r_\tau$ with the highest connectivity score\footnote{The connectivity score of a node $u$ is defined as the similarity-weighted degree of the intra-edges $\gamma(u) = \sum_{\{(u \rightarrow v) | t_u = t_v\}} s_{u \rightarrow v}$.} in each track $\tau$, $r_\tau = \argmax\limits_{\{u | t_u = \tau\}} \gamma(u)$.

\section{Implementation details}

This section describes the loss and dataset used for training the patch alignment network in a supervised manner, as well as details regarding the graph optimization algorithm, hyperparameters, and runtime.

\PAR{Training loss.} 
For training the network, we use a squared L2 loss: $\mathcal{L} = \sum_{P_1, P_2} \lVert d_{1 \rightarrow 2} - d^{\text{gt}}_{1 \rightarrow 2} \rVert_2^2$,
where $d$ and $d^{\text{gt}}$ are the predicted and ground-truth displacements for the central pixel from patch $1$ to patch $2$, respectively.

\PAR{Training dataset.}
We use the MegaDepth dataset~\cite{Li2018MegaDepth} consisting of $196$ different scenes reconstructed from internet images using COLMAP~\cite{Schoenberger2016Structure,Schoenberger2016Pixelwise} to generate training data.
Given the camera intrinsics, extrinsics, and depth maps of each image, a random triangulated SIFT keypoint is selected as reference and reprojected to a matching image to generate a corresponding patch pair.
We enforce depth consistency to ensure that the reference pixel is not occluded in the other view.
We discarded $16$ scenes due to inconsistencies between sparse and dense reconstructions. 
The extracted patch pairs are centered around the SIFT keypoint in the reference view and its reprojected correspondence in the target view respectively (\ie, the ground-truth flow is $\mathbf{0}$). Random homographies are used on the target view to obtain varied ground-truth central point flow.
While the MegaDepth dataset provides training data across a large variety of viewpoint and illumination conditions, the ground-truth flow is sometimes not perfectly sub-pixel accurate due to errors in the dense reconstruction.
Therefore, we synthesize same-condition patch pairs with perfect geometric flow annotation using random warping of reference patches to generate a synthetic counterpart.

\PAR{Feature extraction CNN.}
As the backbone architecture for feature extraction, we use the first two blocks of VGG16 \cite{Simonyan2014Very} (up to \texttt{conv2\_2}) pretrained on ImageNet \cite{Deng2009ImageNet}.
To keep the features aligned with input patch pixels, we replace the $2 \times 2$ max-pooling with stride $2$ by a $3\times3$ max-pooling with stride $2$ and zero padding.

\PAR{Training methodology.}
We start by training the regression head for $5$ epochs.
Afterwards, the entire network is trained end-to-end for $30$ epochs, with the learning rate divided by $10$ every $10$ epochs.
Adam~\cite{Kingma2014Adam} serves as the optimizer with an initial learning rate of $10^{-3}$ and a batch size of $32$.
To counter scene imbalance, $100$ patch pairs are sampled from every scene during each epoch.

\PAR{Graph optimization.} During the optimization, keypoints are allowed to move a maximum of $K=16$ pixels in any direction. We initialize $x_p$ to the initial keypoint locations $x_p^0$. Empirically, we model the soft robust function $\rho$ as Cauchy scaled at $4$ pixels, and the strong one $\psi$ as Tukey scaled at $1$ pixel. We solve the problems from Eq.~\ref{eq:least_squares} for each connected component $\mathcal{A} \in \mathcal{S}$ independently using Ceres~\cite{AgarwalCeres} with sparse Cholesky factorization on the normal equations.

\PAR{Runtime.} The coarse-to-fine patch transformation prediction processes 1-4 image pairs per second on a modern GPU depending on the number of matches. The average runtime of the graph optimization across all methods on the ETH3D scenes is $3.0$s (median runtime $1.0$s) on a CPU with 16 logical processors.

\section{Experimental evaluation}

Despite being trained on SIFT keypoints, our method can be used with a variety of different feature detectors.
To validate this, we evaluate our approach in conjunction with two well-known hand-crafted features (SIFT~\cite{Lowe2004Distinctive} and SURF~\cite{Bay2006SURF}), one learned detector combined with a learned descriptor (Key.Net~\cite{Barroso-Laguna2019KeyNet} with HardNet~\cite{Mishchuk2017Working}), and three learned ones (SuperPoint~\cite{Detone2018SuperPoint} denoted SP, D2-Net~\cite{Dusmanu2019D2}, and R2D2~\cite{Revaud2019R2D2}).
For all methods, we resize the images before feature extraction such that the longest edge is at most $1600$ pixels (lower resolution images are kept unchanged).
We use the default parameters as released by their authors in the associated public code repositories.
Our refinement protocol takes exactly the same input as the feature extraction.
The main objective is not to compare these methods against each other, but rather to show that each of them independently significantly improves when coupled with our refinement procedure.

First, we evaluate the performance with and without refinement on a standard image matching task containing sequences with illumination and viewpoint changes.
Then, we present results in the more complex setting of Structure-from-Motion.
In particular, we demonstrate large improvements on the tasks of multi-view triangulation, camera localization, as well as their combination in an end-to-end image-based 3D reconstruction scenario.

For the Structure-from-Motion evaluations, we use the following matching protocol: for SIFT and SURF, we use a symmetric second nearest neighbor ratio test (with the standard threshold of $0.8$) and mutual nearest neighbors filtering. For Key.Net+HardNet, we use the same protocol with a threshold of $0.9$.
For the remaining methods, we use mutual nearest neighbors filtering with different similarity thresholds - $0.755$ for SuperPoint, $0.8$ for D2-Net, and $0.9$ for R2D2.\footnote{The thresholds for the learned methods were determined following the methodology of~\cite{Lowe2004Distinctive}. Please refer to the supplementary material for more details.}

\subsection{Image matching}

In this experiment, we evaluate the effect of our refinement procedure on the full image sequences from the well-known HPatches dataset~\cite{Balntas2017HPatches}.
This dataset consists of 116 sequences of 6 images with changes in either illumination or viewpoint.
We follow the standard evaluation protocol introduced by~\cite{Dusmanu2019D2} that discards 8 of the sequences due to resolution considerations.
The protocol reports the mean matching accuracy per image pair of a mutual nearest neighbors matcher while varying the pixel threshold up to which a match is considered to be correct.

\begin{figure}[t]
	\centering
	\includegraphics[width=\textwidth]{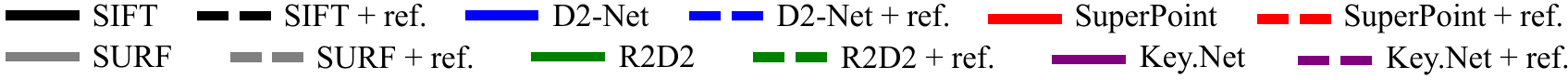} \\
	\begin{minipage}{0.60\textwidth}
		\centering
		\includegraphics[width=\textwidth]{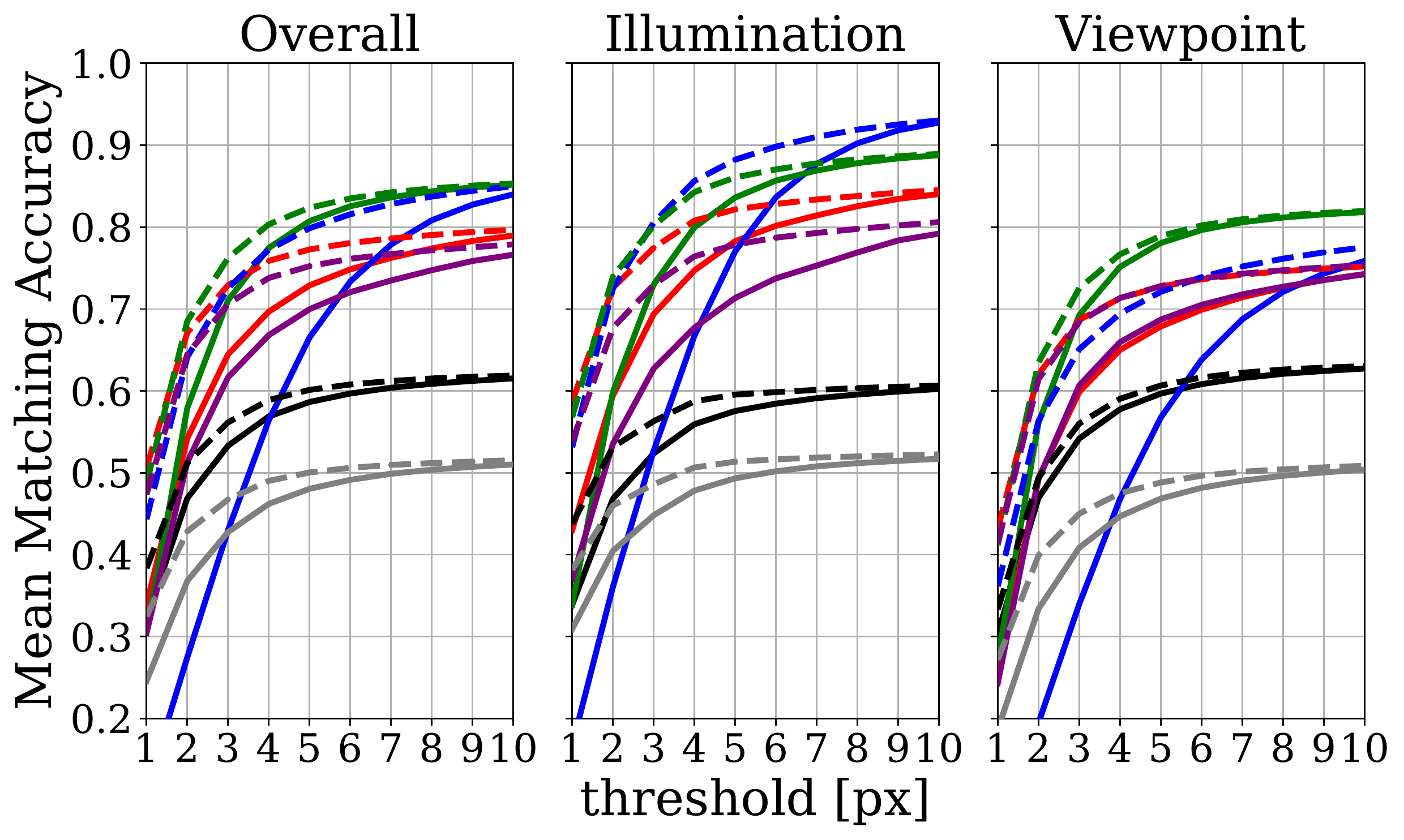}
	\end{minipage}
	~
	\begin{minipage}{0.375\textwidth}
		\tiny
		\centering
		\setlength{\tabcolsep}{1.5pt}
		\begin{tabular}{l | c c c}
			\toprule
			\multirow{2}{*}{Method} & \multicolumn{3}{c}{AUC} \\
			& $2$px & $5$px & $10$px \\ \midrule
			SIFT & $39.49\%$ & $49.57\%$ & $55.15\%$ \\
			SIFT + ref. & $\mathbf{44.78\%}$ & $\mathbf{52.95\%}$ & $\mathbf{57.19\%}$ \\ \midrule
			SURF & $30.69\%$ & $39.68\%$ & $44.96\%$ \\
			SURF + ref. & $\mathbf{37.67\%}$ & $\mathbf{44.23\%}$ & $\mathbf{47.68\%}$ \\ \midrule
			D2-Net & $19.36\%$ & $40.93\%$ & $60.35\%$ \\
			D2-Net + ref. & $\mathbf{54.24\%}$ & $\mathbf{67.62\%}$ & $\mathbf{75.57\%}$ \\ \midrule
			R2D2 & $44.21\%$ & $63.54\%$ & $73.83\%$ \\
			R2D2 + ref. & $\mathbf{58.62\%}$ & $\mathbf{71.23\%}$ & $\mathbf{77.90\%}$ \\ \midrule
			SP & $44.08\%$ & $59.04\%$ & $68.09\%$ \\
			SP + ref. & $\mathbf{58.88\%}$ & $\mathbf{68.76\%}$ & $\mathbf{73.86\%}$ \\ \midrule
			Key.Net & $40.87\%$ & $56.04\%$ & $65.30\%$ \\
			Key.Net + ref. & $\mathbf{55.91\%}$ & $\mathbf{66.29\%}$ & $\mathbf{71.69\%}$ \\ \bottomrule
		\end{tabular}
		\setlength{\tabcolsep}{\tabcolsepdefault}
	\end{minipage}
	\caption{{\bf Matching evaluation.} We plot the mean matching accuracy on HPatches Sequences at different thresholds for illumination and viewpoint sequences, as well as overall. We also report the area under the overall curve (AUC) up to $2$, $5$, and $10$ pixels. All methods have their performance improved by the proposed refinement procedure.}
	\label{fig:hpatches_sequences}
\end{figure}

Figure~\ref{fig:hpatches_sequences} shows the results for illumination-only, viewpoint-only, as well as overall for features with and without refinement.
As expected, our method greatly improves upon learned features under either condition.
Note that the evaluated learned methods represent the state of the art on this benchmark already and we further improve their results.
For SIFT~\cite{Lowe2004Distinctive}, while the performance remains roughly the same under viewpoint changes, our method significantly improves the results under illumination sequences, where low-level changes in image statistics perturb the feature detector.
It is also worth noting that, especially in the viewpoint sequences for learned features, our refinement procedure improves the results for coarse thresholds by correcting wrong, far-away correspondences.

\subsection{Triangulation}

Next, we evaluate the triangulation quality with known ground-truth camera poses and intrinsics on the ETH3D benchmark~\cite{Schoeps2017A}.
Originally, this benchmark was proposed for multi-view stereo methods and provides highly accurate ground-truth camera poses and dense 3D point-clouds.
Nevertheless, the same evaluation protocol also applies to our scenario -- we want to evaluate the impact of refined keypoint locations on the completeness and accuracy of sparse multi-view triangulation.
For each method, we run the multi-view triangulator of COLMAP~\cite{Schoenberger2016Structure} with fixed camera intrinsics and extrinsics.
Given the sparse point cloud, we run the ETH3D evaluation code to report the accuracy (\% of triangulated points) and completeness (\% of ground-truth triangulated points) at different real-world thresholds.
We refer to the original paper for more details about the evaluation.

Table~\ref{tab:eth3d} compares the different local feature approaches with their refined counterparts.
Our proposed keypoint refinement procedure improves the results across the board for all methods.
Once again, the learned keypoints that suffer from poor localization due to downsampling and large receptive field are drastically improved for both indoor and outdoor scenarios.
Even though the performance gain is smaller in the case of SIFT, this experiment shows that exploiting multi-view information is beneficial for very well localized features as well.
The increase in completeness for all local features shows that our approach does not trim the 3D models to only contain accurate points, but rather improves the overall quality by yielding more triangulated points which are also more precise.
Please refer to the supplementary material for results on each dataset.

\setlength{\tabcolsep}{1.5pt}
\begin{table}[t]
	\centering
	\caption{\textbf{Triangulation evaluation.} We report the accuracy (\% of triangulated points) and completeness (\% of ground-truth triangulated points) at $1$cm, $2$cm, and $5$cm. The refined versions outperform their raw counterparts in both metrics.}
	\label{tab:eth3d}
	\tiny
	\begin{tabular}{c | l | c c c | c c c | l | c c c | c c c}
		\toprule
		\multirow{2}{*}{Dataset} & \multirow{2}{*}{Method} & \multicolumn{3}{c |}{Comp. (\%)} & \multicolumn{3}{c |}{Accuracy (\%)} & \multirow{2}{*}{Method} & \multicolumn{3}{c |}{Comp. (\%)} & \multicolumn{3}{c}{Accuracy (\%)} \\
		& & $1$cm & $2$cm & $5$cm & $1$cm & $2$cm & $5$cm & & $1$cm & $2$cm & $5$cm & $1$cm & $2$cm & $5$cm\\ \midrule
		\multirowcell{8}{\textit{Indoors}\\$7$ scenes} & SIFT & $0.20$ & $0.86$ & $3.61$ & $75.74$ & $84.77$ & $92.26$ & SURF & $0.08$ & $0.41$ & $1.97$ & $66.37$ & $79.05$ & $89.61$ \\
		& SIFT + ref. & $\mathbf{0.24}$ & $\mathbf{0.96}$ & $\mathbf{3.88}$ & $\mathbf{81.06}$ & $\mathbf{88.64}$ & $\mathbf{94.61}$ & SURF + ref. & $\mathbf{0.12}$ & $\mathbf{0.52}$ & $\mathbf{2.26}$ & $\mathbf{76.28}$ & $\mathbf{85.30}$ & $\mathbf{92.36}$ \\ \cmidrule{2-15}
		& D2-Net & $0.46$ & $1.83$ & $7.00$ & $46.95$ & $64.91$ & $83.25$ & R2D2 & $0.53$ & $2.04$ & $8.53$ & $66.70$ & $79.26$ & $90.04$ \\
		& D2-Net + ref. & $\mathbf{1.44}$ & $\mathbf{4.53}$ & $\mathbf{12.97}$ & $\mathbf{78.53}$ & $\mathbf{86.46}$ & $\mathbf{93.05}$ & R2D2 + ref. & $\mathbf{0.66}$ & $\mathbf{2.32}$ & $\mathbf{9.08}$ & $\mathbf{77.56}$ & $\mathbf{85.74}$ & $\mathbf{92.54}$ \\ \cmidrule{2-15}
		& SP & $0.59$ & $2.21$ & $8.86$ & $75.26$ & $85.27$ & $93.30$ & Key.Net & $0.16$ & $0.68$ & $3.01$ & $66.51$ & $80.44$ & $91.61$ \\
		& SP + ref. & $\mathbf{0.71}$ & $\mathbf{2.51}$ & $\mathbf{9.55}$ & $\mathbf{86.03}$ & $\mathbf{91.91}$ & $\mathbf{95.83}$ & Key.Net + ref. & $\mathbf{0.21}$ & $\mathbf{0.81}$ & $\mathbf{3.36}$ & $\mathbf{80.51}$ & $\mathbf{89.24}$ & $\mathbf{94.73}$ \\ \midrule
		\multirowcell{8}{\textit{Outdoors}\\$6$ scenes} & SIFT & $0.06$ & $0.34$ & $2.44$ & $58.31$ & $73.13$ & $86.24$ & SURF & $0.03$ & $0.17$ & $1.22$ & $44.21$ & $63.11$ & $79.71$ \\
		& SIFT + ref. & $\mathbf{0.07}$ & $\mathbf{0.41}$ & $\mathbf{2.75}$ & $\mathbf{61.61}$ & $\mathbf{76.89}$ & $\mathbf{88.96}$ & SURF + ref. & $\mathbf{0.05}$ & $\mathbf{0.26}$ & $\mathbf{1.68}$ & $\mathbf{62.88}$ & $\mathbf{74.67}$ & $\mathbf{87.10}$ \\ \cmidrule{2-15}
		& D2-Net & $0.03$ & $0.19$ & $1.80$ & $21.35$ & $35.08$ & $56.75$ & R2D2 & $0.11$ & $0.55$ & $3.61$ & $48.75$ & $65.74$ & $82.81$ \\
		& D2-Net + ref. & $\mathbf{0.21}$ & $\mathbf{1.09}$ & $\mathbf{6.13}$ & $\mathbf{59.07}$ & $\mathbf{72.34}$ & $\mathbf{85.62}$ & R2D2 + ref. & $\mathbf{0.16}$ & $\mathbf{0.71}$ & $\mathbf{4.08}$ & $\mathbf{63.85}$ & $\mathbf{78.10}$ & $\mathbf{90.09}$ \\ \cmidrule{2-15}
		& SP & $0.09$ & $0.54$ & $3.86$ & $49.67$ & $64.57$ & $80.79$ & Key.Net & $0.01$ & $0.09$ & $0.75$ & $39.25$ & $54.57$ & $72.30$\\
		& SP + ref. & $\mathbf{0.15}$ & $\mathbf{0.77}$ & $\mathbf{4.91}$ & $\mathbf{65.23}$ & $\mathbf{77.50}$ & $\mathbf{88.37}$ & Key.Net + ref. & $\mathbf{0.02}$ & $\mathbf{0.13}$ & $\mathbf{0.91}$ & $\mathbf{55.62}$ & $\mathbf{69.41}$ & $\mathbf{85.56}$ \\ \bottomrule
	\end{tabular}
\end{table}
\setlength{\tabcolsep}{\tabcolsepdefault}

\subsection{Camera localization}

We also evaluate the camera localization performance under strict thresholds on the ETH3D dataset~\cite{Schoeps2017A}. 
For each scene, we randomly sample $10$ images that will be treated as queries ($130$ query images in total).
For each query, a partial 3D model is built without the query image and its 2 closest neighbors in terms of co-visibility in the reference model (released with the dataset); 2D-3D correspondences are inferred from the tentative matches between the query image and all (partial) 3D model images; finally, absolute pose estimation with non-linear refinement from COLMAP is used to obtain the camera pose.
The partial models are built independently, \ie, multi-view optimization is only run on the views that are part of each partial model (without the query and holdout images).
For the query keypoints, central point flow is predicted from the reprojected locations of 3D scene points in the matching views to the query view.
To obtain a single 2D coordinate for each matching 3D point, we compute the similarity-weighted average of the flow for each track, which is equivalent to solving Eq.~\ref{eq:least_squares}, where nodes of keypoints in the 3D model are connected through a single edge to matching query keypoints.

\begin{figure}[t]
	\centering
	\includegraphics[width=\textwidth]{images/legend.pdf} \\
	\begin{minipage}{0.26\textwidth}
		\centering
		\includegraphics[width=\textwidth]{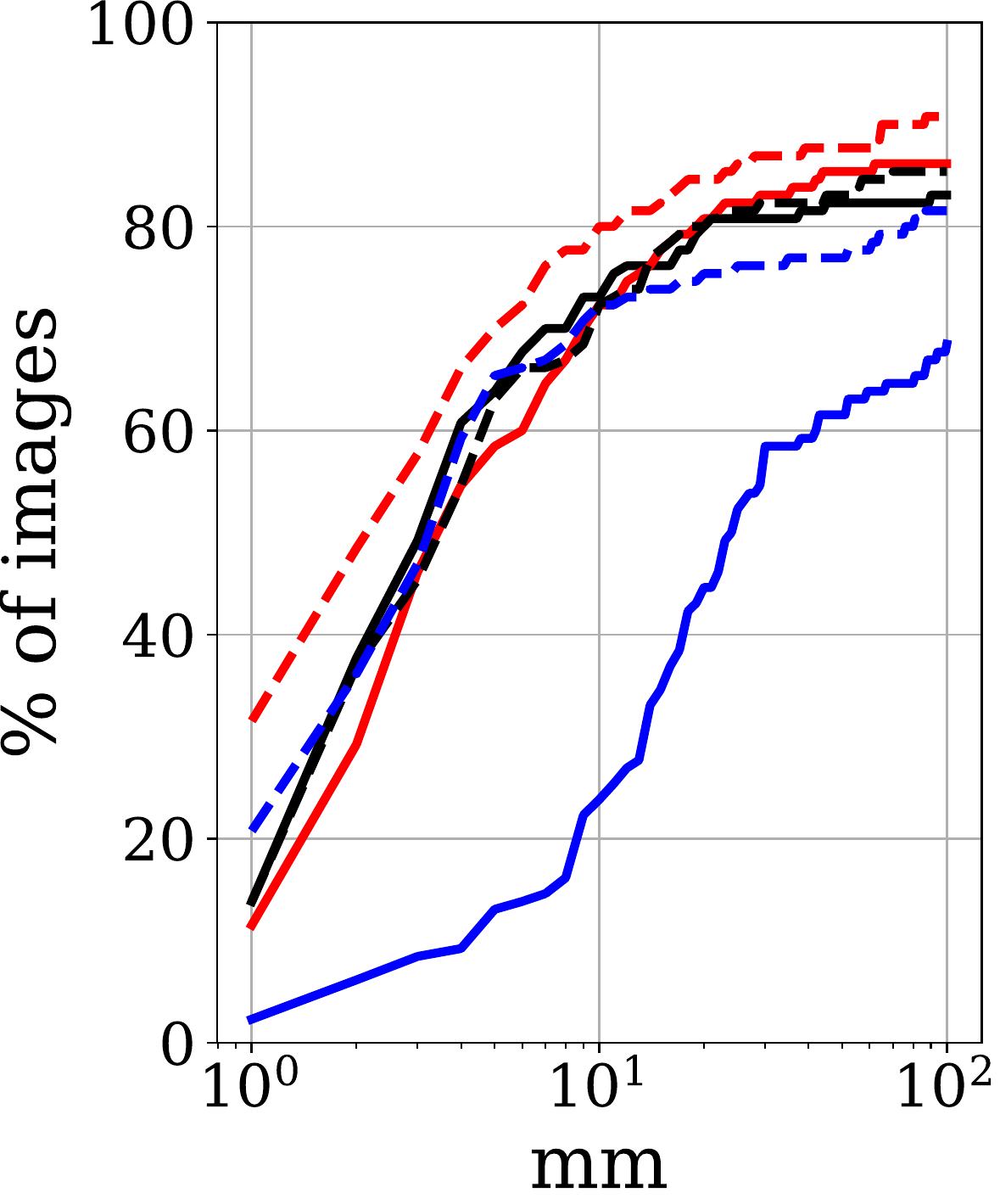}
	\end{minipage}
	~
	\begin{minipage}{0.40\textwidth}
		\tiny
		\centering
		\setlength{\tabcolsep}{1.5pt}
		\begin{tabular}{l | c c c}
			\toprule
			\multirow{2}{*}{Method} & \multicolumn{3}{c}{AUC} \\
			& $1$mm & $1$cm & $10$cm \\ \midrule
			SIFT & $\mathbf{13.85\%}$ & $\mathbf{55.40\%}$ & $79.13\%$ \\
			SIFT + ref. & $\mathbf{13.85\%}$ & $52.82\%$ & $\mathbf{80.30\%}$ \\ \midrule
			SURF & $4.62\%$ & $28.37\%$ & $63.18\%$ \\
			SURF + ref. & $\mathbf{7.69\%}$ & $\mathbf{35.40\%}$ & $\mathbf{65.68\%}$ \\ \midrule
			D2-Net & $2.31\%$ & $12.24\%$ & $54.58\%$ \\
			D2-Net + ref. & $\mathbf{20.77\%}$ & $\mathbf{55.29\%}$ & $\mathbf{76.00\%}$ \\ \midrule
			R2D2 & $10.77\%$ & $52.52\%$ & $81.72\%$ \\
			R2D2 + ref. & $\mathbf{23.08\%}$ & $\mathbf{60.93\%}$ & $\mathbf{82.93\%}$ \\ \midrule
			SP & $11.54\%$ & $51.00\%$ & $81.03\%$ \\
			SP + ref. & $\mathbf{31.54\%}$ & $\mathbf{64.02\%}$ & $\mathbf{85.79\%}$ \\ \midrule
			Key.Net & $3.85\%$ & $29.08\%$ & $68.03\%$ \\
			Key.Net + ref. & $\mathbf{14.62\%}$ & $\mathbf{48.18\%}$ & $\mathbf{72.61\%}$ \\ \bottomrule
		\end{tabular}
		\setlength{\tabcolsep}{\tabcolsepdefault}
	\end{minipage}
	~
	\begin{minipage}{0.26\textwidth}
		\centering
		\includegraphics[width=\textwidth]{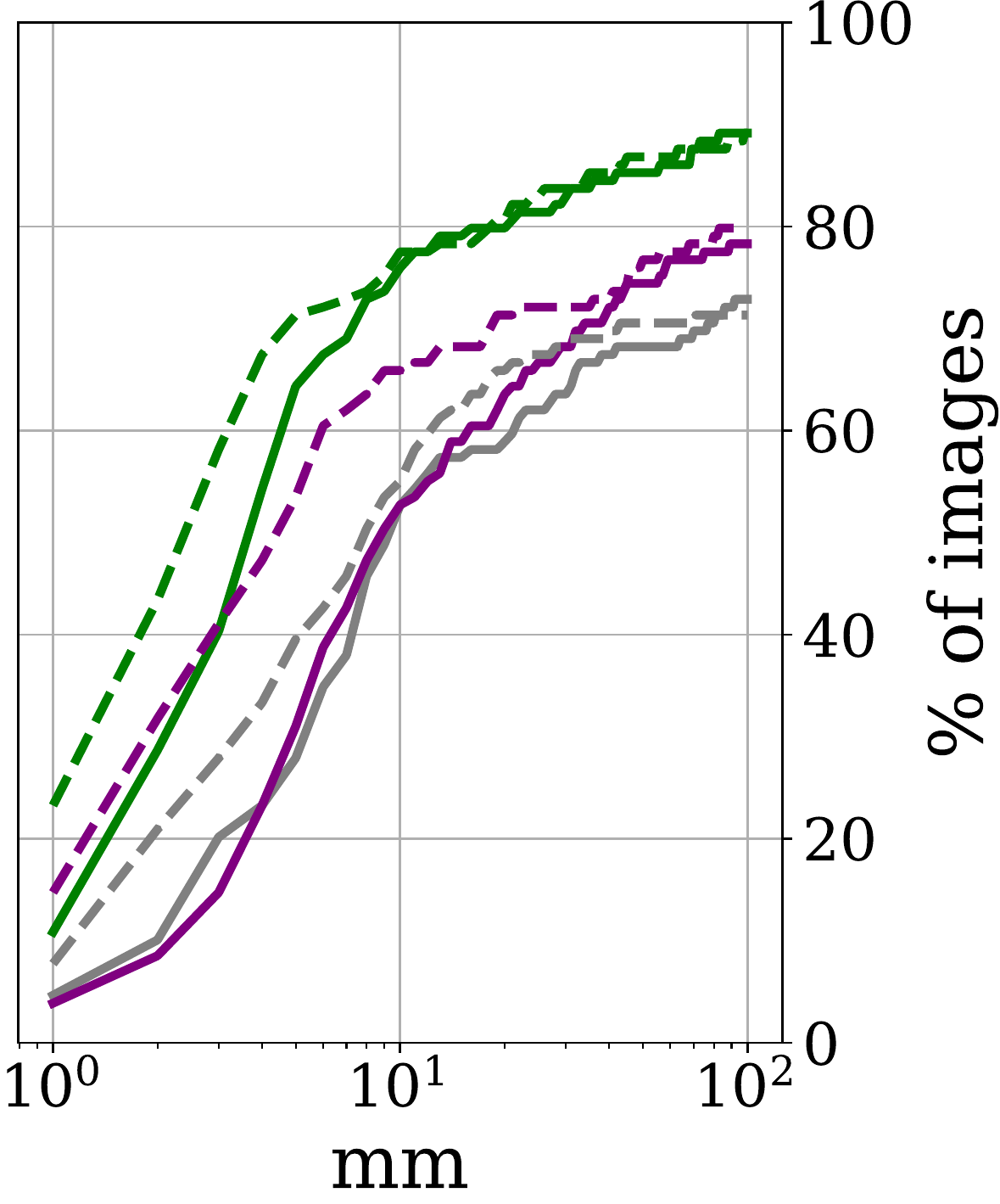}
	\end{minipage}
	\caption{{\bf Camera localization evaluation.} We report the percentage of localized images at different camera position error thresholds as well as the area under the curve (AUC) up to $1$mm, $1$cm and $10$cm. The performance of SIFT remains similar on this task. All other features show greatly improved camera pose accuracy after refinement.}
	\label{fig:eth3d_loc}
\end{figure}

The results of this experiment are presented in Figure~\ref{fig:eth3d_loc}.
The performance of SIFT~\cite{Lowe2004Distinctive} after refinement is on par with the unrefined version despite the increase in point-cloud accuracy and completeness; this suggests that the method has nearly saturated on this localization task.
All the other features have their performance greatly improved by the proposed refinement.
It is worth noting that the refined versions of SuperPoint~\cite{Detone2018SuperPoint} and R2D2~\cite{Revaud2019R2D2} drastically outperform SIFT especially on the finer thresholds ($1$mm and $1$cm).

\subsection{Structure-from-Motion}

Finally, we evaluate our refinement procedure on the scenario of end-to-end 3D reconstruction from unstructured imagery on the benchmark introduced in~\cite{Schonberger2017Comparative}.
For the internet datasets (Madrid Metropolis, Gendarmenmarkt, and Tower of London), instead of exhaustively matching all images, we use NetVLAD~\cite{Arandjelovic2016NetVLAD} to retrieve top $20$ related views for each image and only match against these.
Due to the wide range of resolutions in internet images, we impose the use of multi-scale features if available and not active by default (\ie, for D2-Net~\cite{Dusmanu2019D2}).

After matching and feature refinement, we run COLMAP~\cite{Schoenberger2016Structure} to obtain sparse 3D reconstructions.
Finally, different reconstruction statistics taking into account only the images registered both with and without refinement are reported in Table~\ref{tab:lfe}.
For independent results, please refer to the supplementary material.

Overall, the results with refined keypoints achieve significantly better statistics than their original counterparts.
On the small datasets all refined methods apart from R2D2 have sub-pixel keypoint accuracy (\ie, a reprojection error lower than $0.5$).
SuperPoint and Key.Net, despite being targeted at low-level features, are still largely behind SIFT in terms of reprojection error without refinement.
The refinement lowers this gap while also improving their already significant track length.
For SIFT, the main improvement is in terms of reprojection error showing that it is possible to refine even features with accurate, sub-pixel keypoint localization.
For R2D2 and SURF, on the large datasets, we see a tendency to very slightly decrease the track length to improve the reprojection error. This points to the fact that loosely grouped features during SfM are split into multiple, but more accurate feature tracks.
The results on the large internet datasets notably show the robustness of the multi-view refinement to incorrect matches, repeated structures, drastic illumination changes, and large, complex graphs with as much as 5 million nodes and more than 1 million tracks.

\setlength{\tabcolsep}{1.5pt}
\begin{table}[t]
	\centering
	\caption{\textbf{Local Feature Evaluation Benchmark.} A 3D model is built for each method and different reconstruction statistics are reported. For the large datasets, we report the statistics on the common images only.}
	\label{tab:lfe}
	\tiny
	\begin{tabular}{c | l | c c c c | l | c c c c}
		\toprule
		\multirow{2}{*}{Dataset} & \multirow{2}{*}{Method} & \multirowcell{2}{Reg. \\ images} & \multirowcell{2}{Num. \\ obs.} & \multirowcell{2}{Track \\ length} & \multirowcell{2}{Reproj. \\ error} & \multirow{2}{*}{Method} & \multirowcell{2}{Reg. \\ images} & \multirowcell{2}{Num. \\ obs.} & \multirowcell{2}{Track \\ length} & \multirowcell{2}{Reproj. \\ error} \\
		& & & & & & & & & \\ \midrule
		\multirowcell{8}{\textit{Herzjesu} \\ $8$ images} & SIFT & \multirow{2}{*}{8} & 15.9K & 4.10 & 0.59 & SURF & \multirow{2}{*}{8} & 5.0K & 3.64 & 0.70 \\
		& SIFT + ref. & & \bf{16.2K} & \bf{4.16} & \bf{0.29} & SURF + ref. & & \bf{5.2K} & \bf{3.70} & \bf{0.30} \\ \cmidrule{2-11}
		& D2-Net & \multirow{2}{*}{8} & 38.5K & 3.36 & 1.32 & R2D2 & \multirow{2}{*}{8} & 21.1K & 5.84 & 1.08 \\
		& D2-Net + ref. & & \bf{47.7K} & \bf{4.06} & \bf{0.41} & R2D2 + ref. & & \bf{21.6K} & \bf{6.04} & \bf{0.57} \\ \cmidrule{2-11}
		& SP & \multirow{2}{*}{8} & 17.2K & 4.54 & 1.00 & Key.Net & \multirow{2}{*}{8} & 5.0K & 4.29 & 1.00 \\
		& SP + ref. & & \bf{17.9K} & \bf{4.72} & \bf{0.36} & Key.Net + ref. & & \bf{5.3K} & \bf{4.46} & \bf{0.42} \\ \midrule
		\multirowcell{8}{\textit{Fountain} \\ $11$ images} & SIFT & \multirow{2}{*}{11} & 27.0K & 4.51 & 0.55 & SURF & \multirow{2}{*}{11} & 5.6K & 3.91 & 0.64 \\
		& SIFT + ref. & & \bf{27.4K} & \bf{4.56} & \bf{0.26} & SURF + ref. & & \bf{5.7K} & \bf{3.95} & \bf{0.30} \\ \cmidrule{2-11}
		& D2-Net & \multirow{2}{*}{11} & 62.0K & 3.51 & 1.36 & R2D2 & \multirow{2}{*}{11} & 33.0K & 7.11 & 1.10 \\
		& D2-Net + ref. & & \bf{77.4K} & \bf{4.47} & \bf{0.40} & R2D2 + ref. & & \bf{33.6K} & \bf{7.47} & \bf{0.62} \\ \cmidrule{2-11}
		& SP & \multirow{2}{*}{11} & 21.5K & 4.93 & 1.06 & Key.Net & \multirow{2}{*}{11} & 8.4K & 5.53 & 1.00 \\
		& SP + ref. & & \bf{22.4K} & \bf{5.19} & \bf{0.43} & Key.Net + ref. & & \bf{8.7K} & \bf{5.70} & \bf{0.44} \\ \midrule
		\multirowcell{8}{\textit{Madrid} \\ \textit{Metropolis} \\ $1344$ images} & SIFT &  \multirow{2}{*}{379} & 187.2K & 6.83 & 0.70 & SURF & \multirow{2}{*}{268} & \bf{116.0K} & \bf{6.25} & 0.76 \\
		& SIFT + ref. & & \bf{187.7K} & \bf{6.86} & \bf{0.66} & SURF + ref. & & 115.2K & \bf{6.25} & \bf{0.66} \\ \cmidrule{2-11}
		& D2-Net & \multirow{2}{*}{372} & 668.8K & 6.00 & 1.47 & R2D2 & \multirow{2}{*}{410} & 355.2K & \bf{10.20} & 0.90 \\ 
		& D2-Net + ref. & & \bf{752.5K} & \bf{7.28} & \bf{0.96} & R2D2 + ref. & & \bf{356.8K} & 10.17 & \bf{0.76} \\ \cmidrule{2-11}
		& SP & \multirow{2}{*}{414} & 269.7K & 7.64 & 0.98 & Key.Net & \multirow{2}{*}{304} & 111.9K & 9.18 & 0.94 \\
		& SP + ref. & & \bf{277.7K} & \bf{8.20} & \bf{0.72} & Key.Net + ref. & & \bf{114.5K} & \bf{9.31} & \bf{0.75} \\ \midrule
		\multirowcell{8}{\textit{Gendarmen-} \\ \textit{markt} \\ $1463$ images} & SIFT & \multirow{2}{*}{874} & 440.3K & 6.33 & 0.82 & SURF & \multirow{2}{*}{472} & 163.9K & \bf{5.45} & 0.90 \\
		& SIFT + ref. & & \bf{441.4K} & \bf{6.42} & \bf{0.75} & SURF + ref. & & \bf{164.8K} & 5.43 & \bf{0.78} \\ \cmidrule{2-11}
		& D2-Net & \multirow{2}{*}{858} & 1.479M & 5.33 & 1.44 & R2D2 & \multirow{2}{*}{929} & \bf{1.043M} & \bf{10.09} & 0.99 \\
		& D2-Net + ref. & & \bf{1.665M} & \bf{6.37} & \bf{1.04} & R2D2 + ref. & & \bf{1.043M} & 10.05 & \bf{0.89} \\ \cmidrule{2-11}
		& SP & \multirow{2}{*}{911} & 626.9K & 6.84 & 1.05 & Key.Net & \multirow{2}{*}{810} & 253.3K & 7.08 & 0.99 \\
		& SP + ref. & & \bf{648.0K} & \bf{7.10} & \bf{0.89} & Key.Net + ref. & & \bf{258.6K} & \bf{7.25} & \bf{0.86} \\ \midrule
		\multirowcell{8}{\textit{Tower of} \\ \textit{London} \\ $1576$ images} & SIFT & \multirow{2}{*}{561} & 447.8K & 7.90 & 0.69 & SURF & \multirow{2}{*}{430} & 212.0K & \bf{5.94} & 0.70 \\
		& SIFT + ref. & & \bf{449.0K} & \bf{7.96} & \bf{0.59} & SURF + ref. & & \bf{212.7K} & 5.92 & \bf{0.58} \\ \cmidrule{2-11}
		& D2-Net & \multirow{2}{*}{635} & 1.408M & 5.96 & 1.48 & R2D2 & \multirow{2}{*}{689} & 758.0K & 13.44 & 0.92 \\ 
		& D2-Net + ref. & & \bf{1.561M} & \bf{7.63} & \bf{0.91} & R2D2 + ref. & & \bf{759.2K} & \bf{13.74} & \bf{0.76} \\ \cmidrule{2-11}
		& SP & \multirow{2}{*}{621} & 442.9K & 8.06 & 0.95 & Key.Net & \multirow{2}{*}{495} & 186.5K & 9.02 & 0.85 \\
		& SP + ref. & & \bf{457.6K} & \bf{8.55} & \bf{0.69} & Key.Net + ref. & & \bf{190.8K} & \bf{9.18} & \bf{0.65} \\ \bottomrule
	\end{tabular}
\end{table}
\setlength{\tabcolsep}{\tabcolsepdefault}

\section{Conclusion}
We have proposed a novel method for keypoint refinement from multiple views.
Our approach is agnostic to the type of local features and seamlessly integrates into the standard feature extraction and matching paradigm.
We use a patch alignment neural network for two-view flow prediction and formulate the multi-view refinement as a non-linear least squares optimization problem.
The experimental evaluation demonstrates drastically improved performance on the Structure-from-Motion tasks of triangulation and camera localization.
Throughout our experiments, we have shown that our refinement cannot only address the poor keypoint localization of recent learned feature approaches, but it can also improve upon SIFT -- the arguably most well-known handcrafted local feature with accurate sub-pixel keypoint refinement. 

{\noindent \textbf{Acknowledgements.} This work was supported by the Microsoft Mixed Reality \& AI Z\"urich Lab PhD scholarship.}

\appendix

\section*{Supplementary material}

This supplementary material provides the following information:
Section~\ref{sec:filtering} explains how we determined the match filtering thresholds for the learned methods.
Section~\ref{sec:results} contains the additional results mentioned in the main paper (\eg, ETH3D~\cite{Schoeps2017A} triangulation results on each individual dataset and independent results of each method on the Local Feature Evaluation Benchmark~\cite{Schonberger2017Comparative}) as well as some qualitative examples before and after refinement.
Section~\ref{sec:ablation} presents an ablation study for both the two-view and the multi-view refinement procedure.
Section~\ref{sec:query} details the query keypoint refinement protocol used for camera localization on the ETH3D dataset.
Section~\ref{sec:training} describes the filtering steps used during the generation of the two-view training dataset.

\section{Match filtering}
\label{sec:filtering}

\begin{figure}[t]
	\begin{minipage}{0.30\textwidth}
		\centering
		\includegraphics[width=\textwidth]{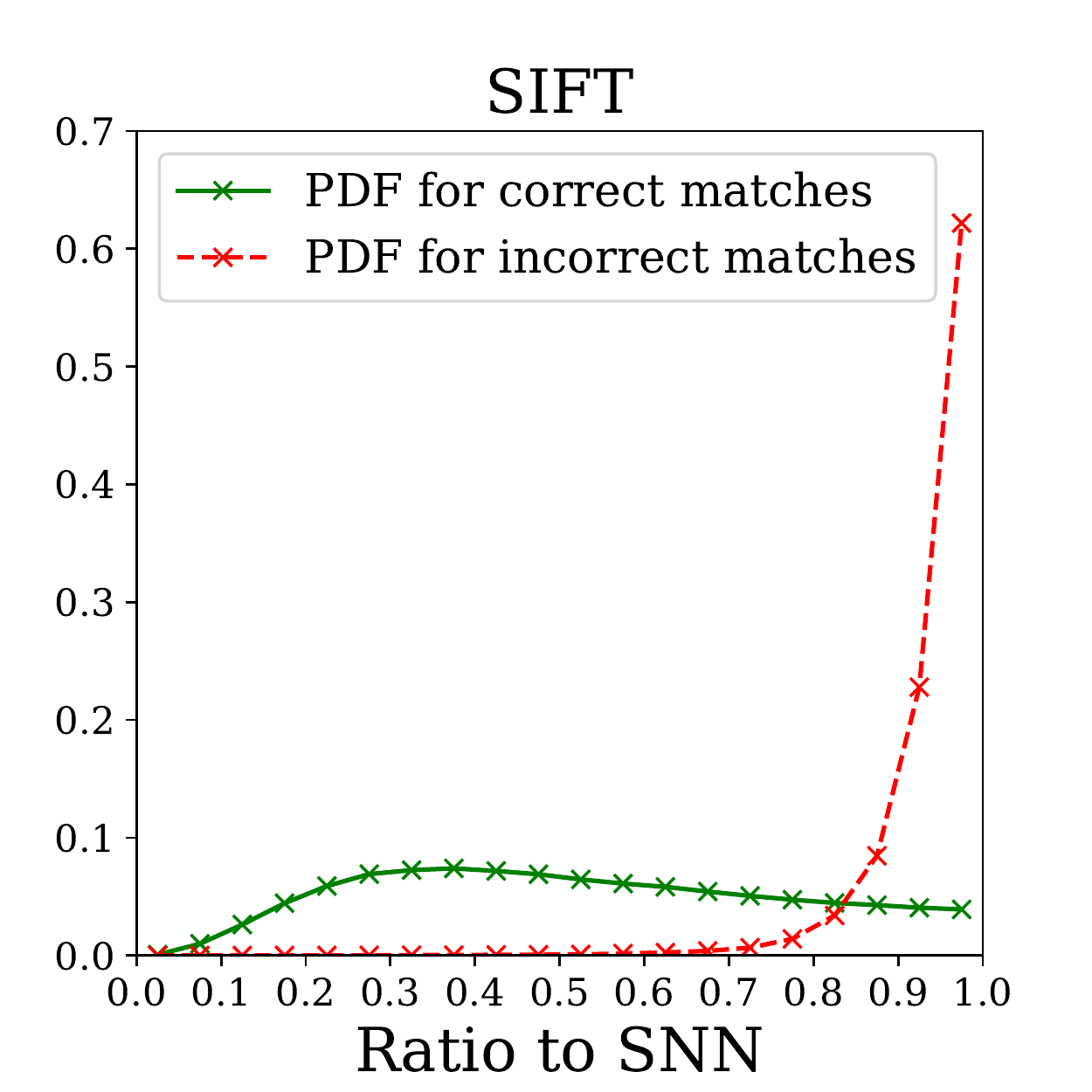}
	\end{minipage}
	~
	\begin{minipage}{0.35\textwidth}
		\centering
		\setlength{\tabcolsep}{1.5pt}
		\tiny
		\begin{tabular}{l | c | c c}
			\toprule
			\multirow{2}{*}{Method} & \multirow{2}{*}{Filtering} & \% correct & \% wrong \\
			& & \multicolumn{2}{c}{filtered out}\\ \midrule
			SIFT & Ratio 0.8 & 16.7\% & 96.8\% \\ \midrule
			KeyNey & Ratio 0.9 & 19.7\% & 87.9\% \\ \midrule
			D2-Net & Sim. 0.8 & 17.2\% & 62.7\% \\ \midrule
			R2D2 & Sim. 0.9 & 22.1\% & 83.0\% \\ \midrule
			SP & Sim. 0.755 & 12.8\% & 82.0\% \\
			\bottomrule
		\end{tabular}
		\setlength{\tabcolsep}{\tabcolsepdefault}
	\end{minipage}
	~
	\begin{minipage}{0.30\textwidth}
		\centering
		\includegraphics[width=\textwidth]{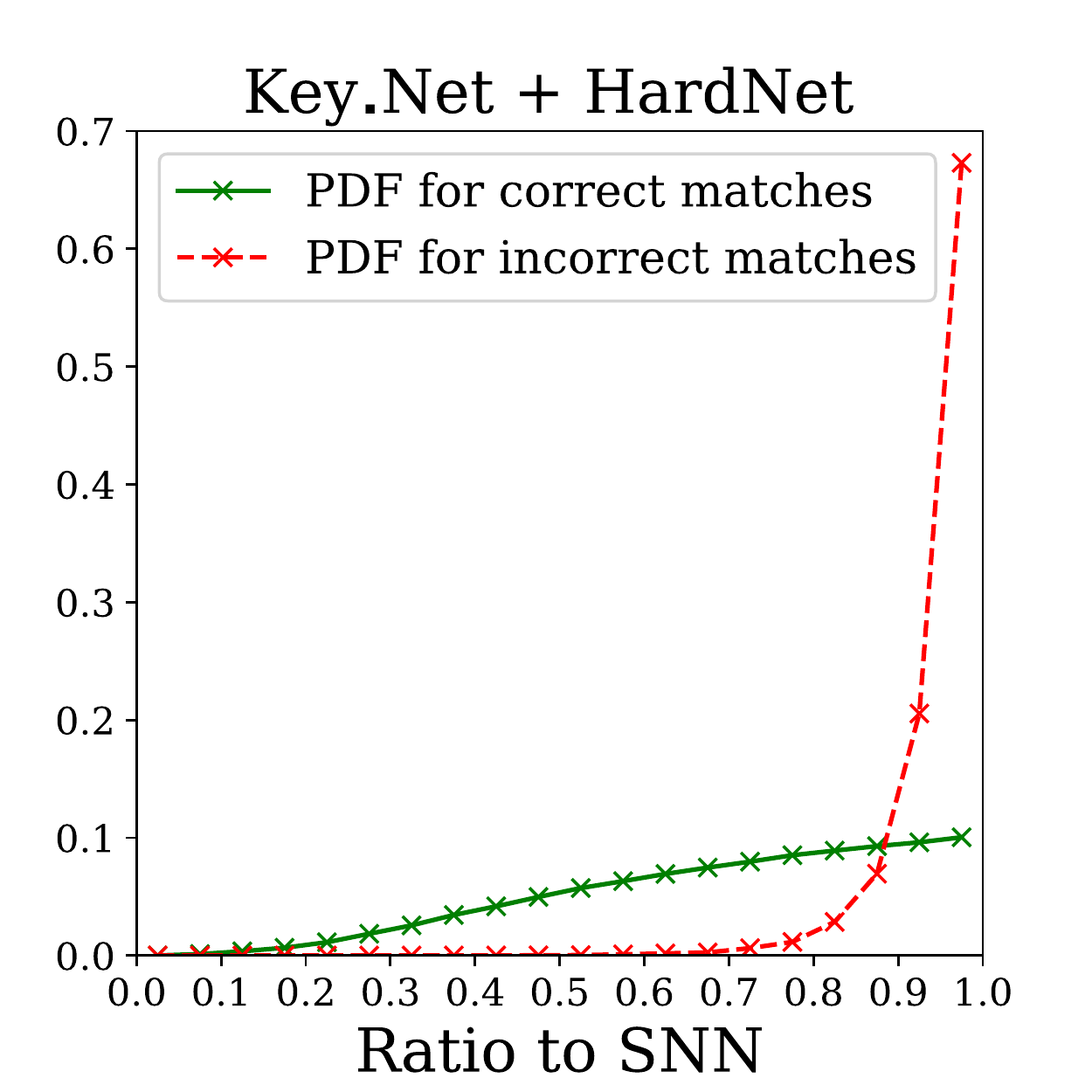}
	\end{minipage}\\
	\begin{minipage}{0.30\textwidth}
		\centering
		\includegraphics[width=\textwidth]{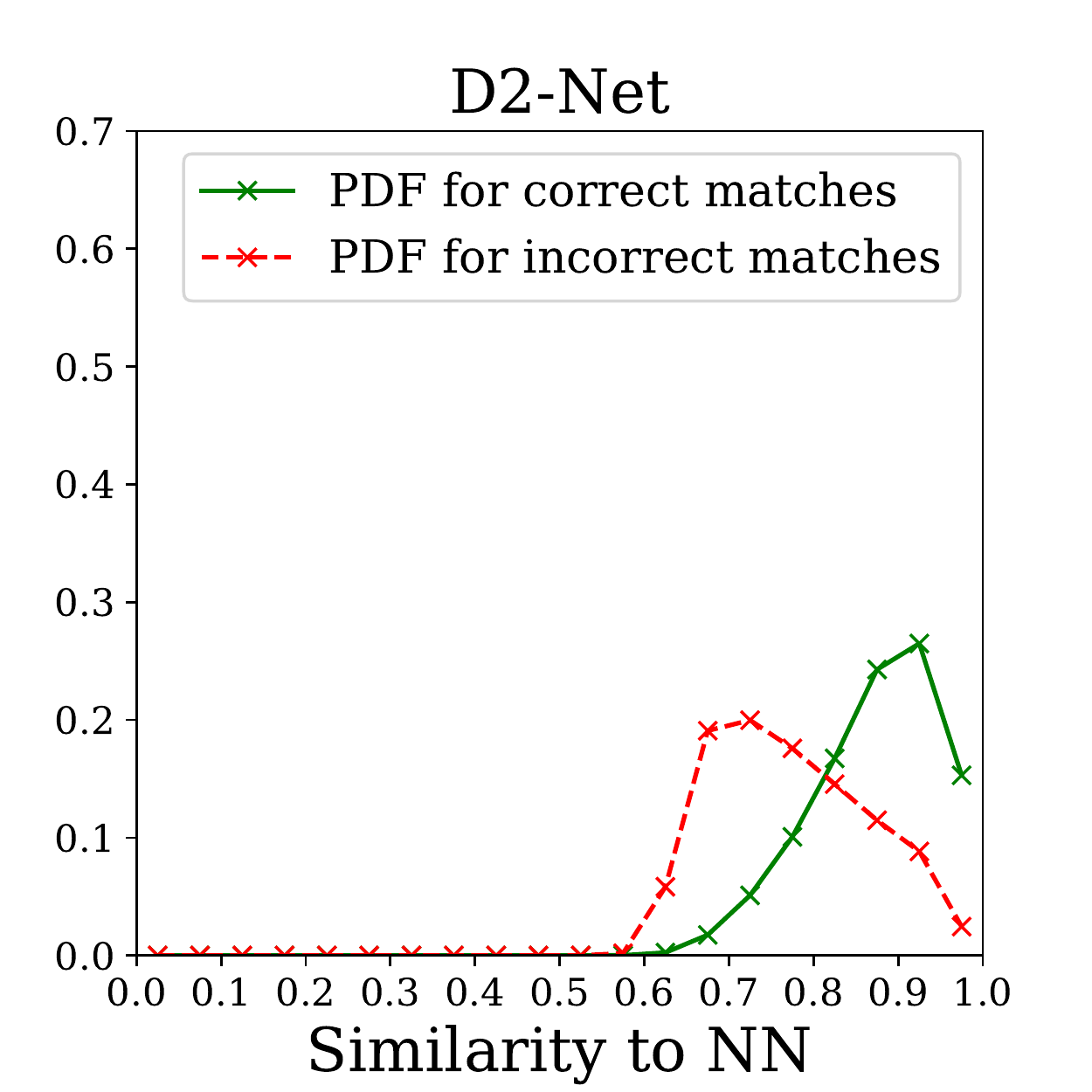}
	\end{minipage}
	~
	\begin{minipage}{0.35\textwidth}
		\centering
		\includegraphics[width=.857\textwidth]{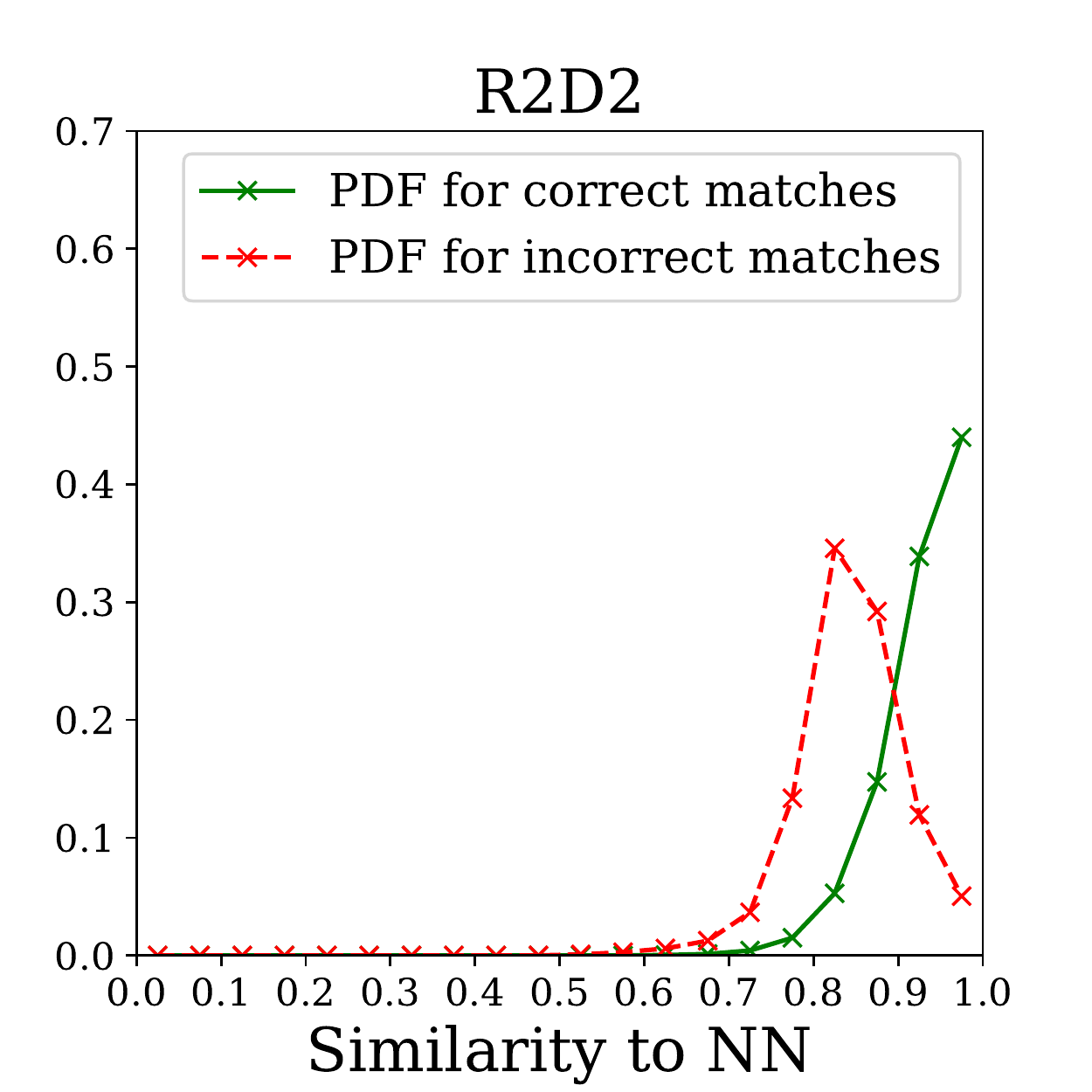}
	\end{minipage}
	~
	\begin{minipage}{0.30\textwidth}
		\centering
		\includegraphics[width=\textwidth]{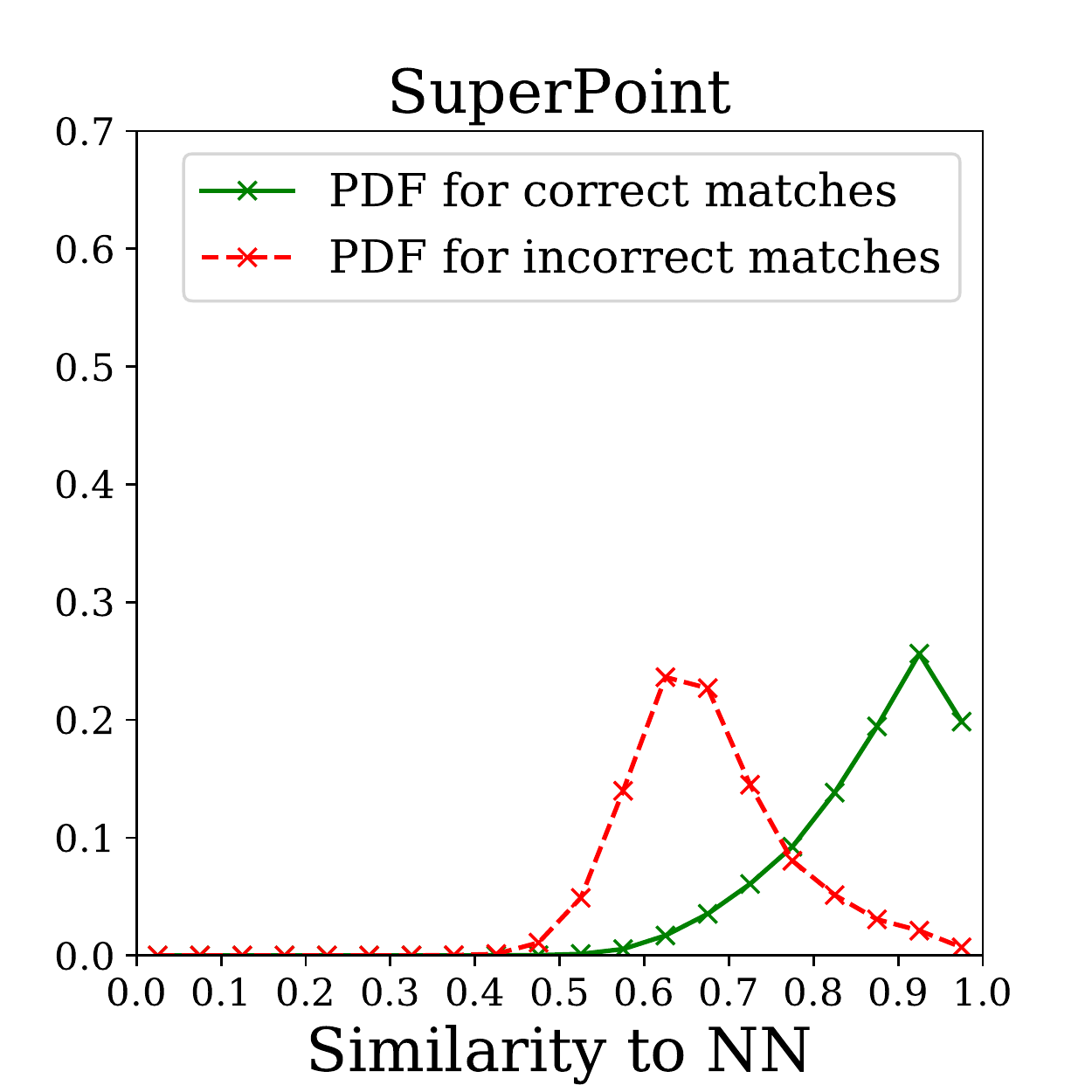}
	\end{minipage}
	\caption{{\bf Match filtering.} Following the protocol of Lowe~\cite{Lowe2004Distinctive}, we plot the probability distribution function (PDF) of the correct and incorrect mutual nearest neighbors matches. The horizontal axis represents either the ratio to the second nearest neighbor or the cosien similarity to the first nearest neighbor.}
	\label{fig:match-filtering}
\end{figure}

Match filtering is an essential step before large-scale SfM because it significantly reduces the number of wrong registrations due to repetitive structures and semantically similar scenes.
To determine a good threshold (either for similarity or ratio to the second nearest neighbor), we adopt the methodology suggested by Lowe~\cite{Lowe2004Distinctive} -- we plot the probability distribution functions for correct and incorrect mutual nearest neighbors matches on the sequences from the HPatches dataset~\cite{Balntas2017HPatches}.
A match is considered correct if its projection error, estimated using the ground-truth homographies, is below $4$ pixels.
To have a clear separation, the threshold for incorrect matches is set to 12 pixels. 
All matches with errors in-between are discarded.
Figure~\ref{fig:match-filtering} shows the plots for all learned methods as well as SIFT (used as reference).

For SIFT~\cite{Lowe2004Distinctive}, the ratio threshold traditionally used ($0.8$) filters out $16.7\%$ of correct matches and $96.8\%$ of wrong ones.
For SuperPoint~\cite{Detone2018SuperPoint}, we use the cosine similarity threshold suggested by the authors ($0.755$) which filters out $82.0\%$ of wrong matches.
For Key.Net~\cite{Barroso-Laguna2019KeyNet} and R2D2~\cite{Revaud2019R2D2}, we empirically determine thresholds with a similar filtering performance to the ones used for SIFT and SuperPoint.
The only method that is not compatible with either the ratio test or similarity thresholding is D2-Net~\cite{Dusmanu2019D2}. Thus, for it, we settle on a conservative similarity threshold of $0.8$, filtering out only $62.7\%$ of incorrect matches.

\section{Additional results}
\label{sec:results}

For the Local Feature Evaluation Benchmark~\cite{Schonberger2017Comparative}, the results reported in the main paper show the sparse 3D reconstruction statistics on the images registered by both the refined and unrefined versions of each feature - this was done in order to allow a fair comparison in terms of number of observations, track length, and reprojection error. Nevertheless, we also provide the independent results for each local feature in Table~\ref{tab:lfe-indep}.

\setlength{\tabcolsep}{1.5pt}
\begin{table}[t]
	\centering
	\caption{\textbf{Evaluation on the Local Feature Evaluation Benchmark.} We report the results for each method independently, instead of considering only the commonly registered images for refined and unrefined features.}
	\label{tab:lfe-indep}
	\tiny
	\begin{tabular}{c | l | c c c c | l | c c c c}
		\toprule
		\multirow{2}{*}{Dataset} & \multirow{2}{*}{Method} & \multirowcell{2}{Reg. \\ images} & \multirowcell{2}{Num. \\ obs.} & \multirowcell{2}{Track \\ length} & \multirowcell{2}{Reproj. \\ error} & \multirow{2}{*}{Method} & \multirowcell{2}{Reg. \\ images} & \multirowcell{2}{Num. \\ obs.} & \multirowcell{2}{Track \\ length} & \multirowcell{2}{Reproj. \\ error} \\
		& & & & & & & & & \\ \midrule
		\multirowcell{7}{\textit{Madrid} \\ \textit{Metropolis} \\ $1344$ images} & SIFT & \textbf{393} & 188.7K & 6.84 & 0.70 & SURF & 296 & \textbf{121.4K} & 6.22 & 0.76 \\
		& SIFT + ref. & 390 & \textbf{189.7K} & \textbf{6.90} & \textbf{0.66} & SURF + ref. & \textbf{274} & 116.6K & \textbf{6.26} & \textbf{0.66} \\ \cmidrule{2-11}
		& D2-Net & 392 & 683.6K & 6.01 & 1.46 & R2D2 & 422 & 357.2K & \textbf{10.17} & 0.90 \\ 
		& D2-Net + ref. & \textbf{405} & \textbf{773.4K} & \textbf{7.26} & \textbf{0.96} & R2D2 + ref. & \textbf{427} & \textbf{359.5K} & 10.15 & \textbf{0.76} \\ \cmidrule{2-11}
		& SP & 422 & 272.1K & 7.64 & 0.98 & Key.Net & 317 & 114.4K & 9.28 & 0.94 \\
		& SP + ref. & \textbf{425} & \textbf{279.9K} & \textbf{8.23} & \textbf{0.72} & Key.Net + ref. & \textbf{323} & \textbf{119.4K} & \textbf{9.39} & \textbf{0.75} \\ \midrule
		\multirowcell{7}{\textit{Gendarmen-} \\ \textit{markt} \\ $1463$ images} & SIFT & 879 & 440.7K & 6.34 & 0.82 & SURF & 475 & 164.1K & \textbf{5.45} & 0.90 \\
		& SIFT + ref. & \textbf{882} & \textbf{442.2K} & \textbf{6.41} & \textbf{0.75} & SURF + ref. & \textbf{483} & \textbf{165.6K} & 5.42 & \textbf{0.78} \\ \cmidrule{2-11}
		& D2-Net & 865 & 1.482M & 5.33 & 1.44 & R2D2 & \textbf{988} & \textbf{1.102M} & 9.94 & 0.98 \\
		& D2-Net + ref. & \textbf{959} & \textbf{1.805M} & \textbf{6.38} & \textbf{1.02} & R2D2 + ref. & 935 & 1.044M & \textbf{10.04} & \textbf{0.89} \\ \cmidrule{2-11}
		& SP & 919 & 627.4K & 6.84 & 1.05 & Key.Net & 817 & 253.9K & 7.08 & 0.99 \\
		& SP + ref. & \textbf{972} & \textbf{680.6K} & \textbf{7.07} & \textbf{0.88} & Key.Net + ref. & \textbf{828} & \textbf{260.5K} & \textbf{7.21} & \textbf{0.86} \\ \midrule
		\multirowcell{7}{\textit{Tower of} \\ \textit{London} \\ $1576$ images} & SIFT & 562 & 448.9K & 7.90 & 0.69 & SURF & \textbf{433} & 212.2K & \textbf{5.94} & 0.71 \\
		& SIFT + ref. & \textbf{566} & \textbf{449.6K} & \textbf{7.96} & \textbf{0.59} & SURF + ref. & 432 & \textbf{212.9K} & 5.92 & \textbf{0.58} \\ \cmidrule{2-11}
		& D2-Net & 653 & 1.417M & 5.93 & 1.48 & R2D2 & 693 & 758.2K & 13.44 & 0.92 \\ 
		& D2-Net + ref. & \textbf{661} & \textbf{1.568M} & \textbf{7.64} & \textbf{0.91} & R2D2 + ref. & \textbf{700} & \textbf{760.8K} & \textbf{13.73} & \textbf{0.76} \\ \cmidrule{2-11}
		& SP & 625 & 443.3K & 8.06 & 0.95 & Key.Net & \textbf{500} & 186.9K & 9.03 & 0.85 \\
		& SP + ref. & \textbf{633} & \textbf{458.9K} & \textbf{8.52} & \textbf{0.69} & Key.Net + ref. & 495 & \textbf{190.8K} & \textbf{9.18} & \textbf{0.66} \\ \bottomrule
	\end{tabular}
\end{table}
\setlength{\tabcolsep}{\tabcolsepdefault}

\begin{figure}
	\centering
	\begin{tabular}{c c c}
		SIFT & \hspace{2cm} & SURF \\
		\includegraphics[width=.35\textwidth]{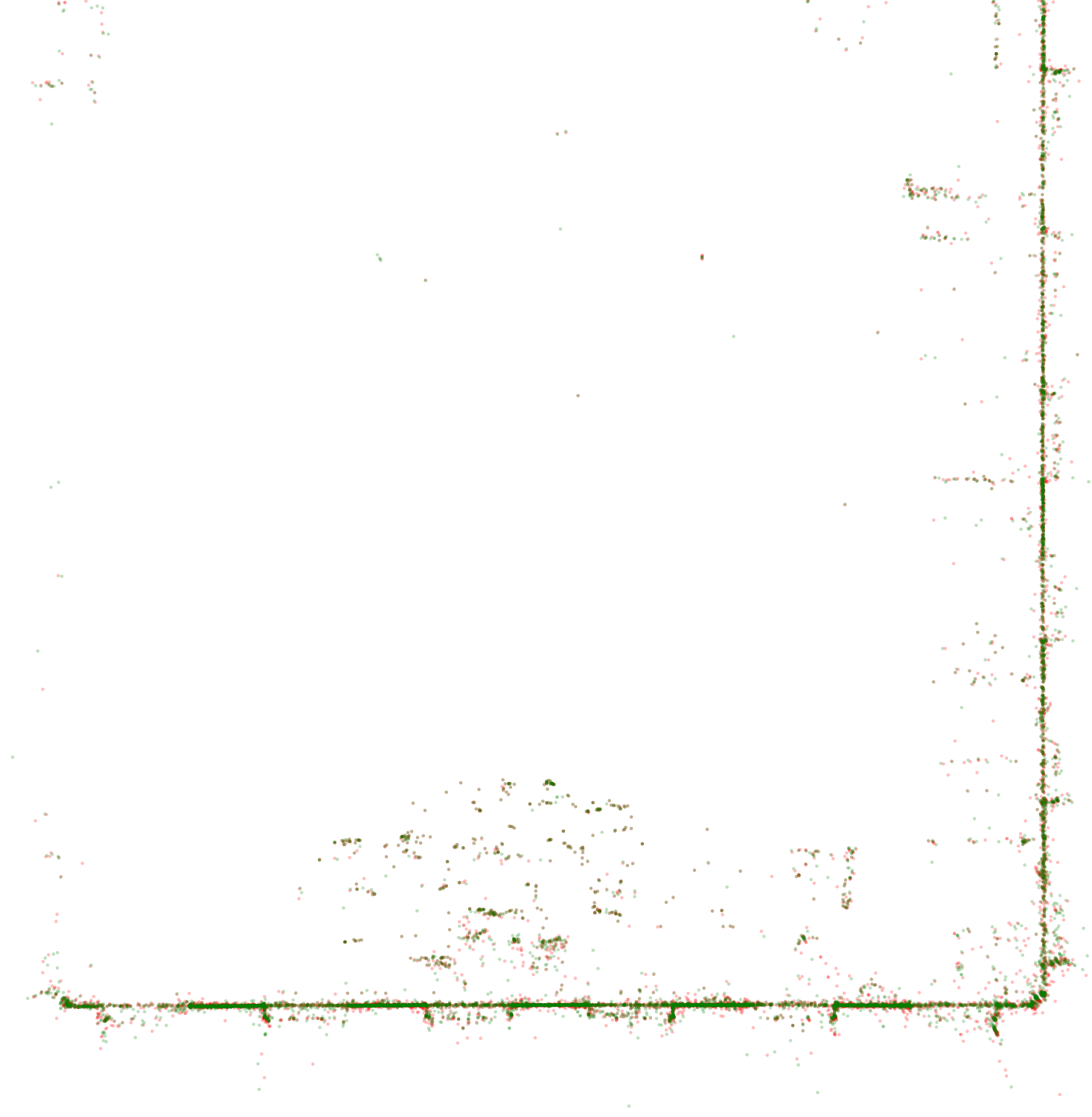} & & \includegraphics[width=.35\textwidth]{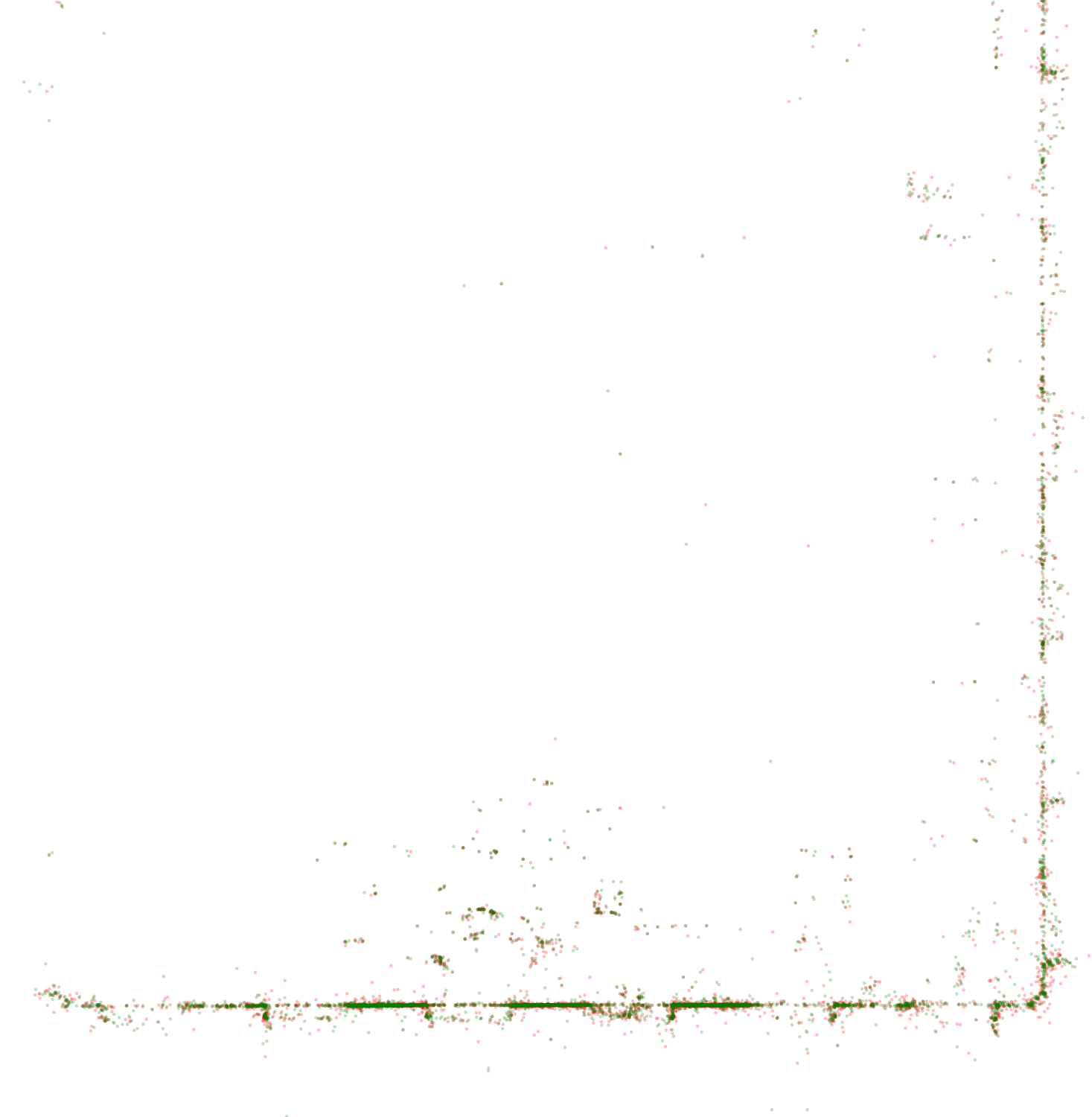} \\
		D2-Net & & R2D2 \\
		\includegraphics[width=.35\textwidth]{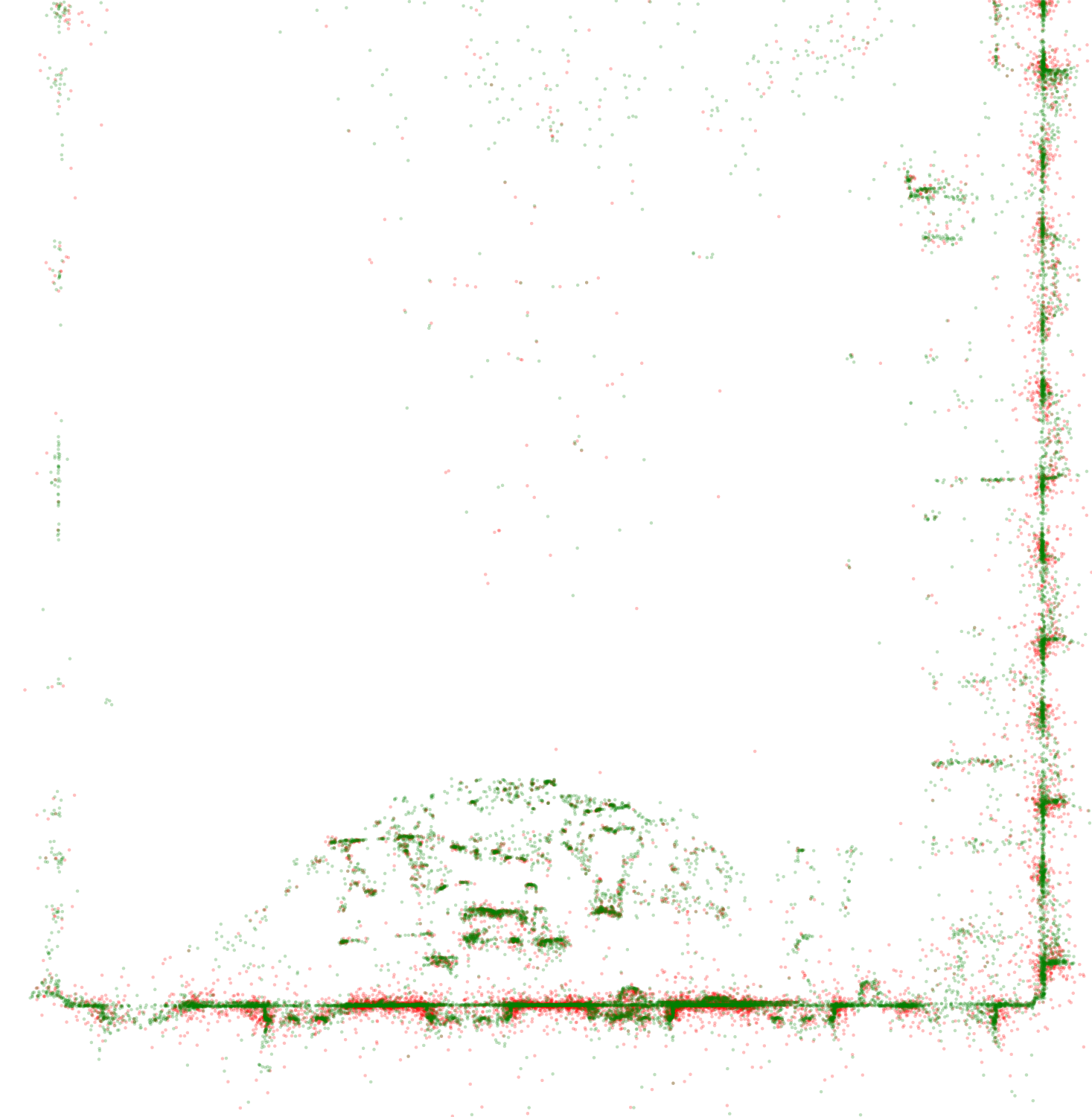} & & \includegraphics[width=.35\textwidth]{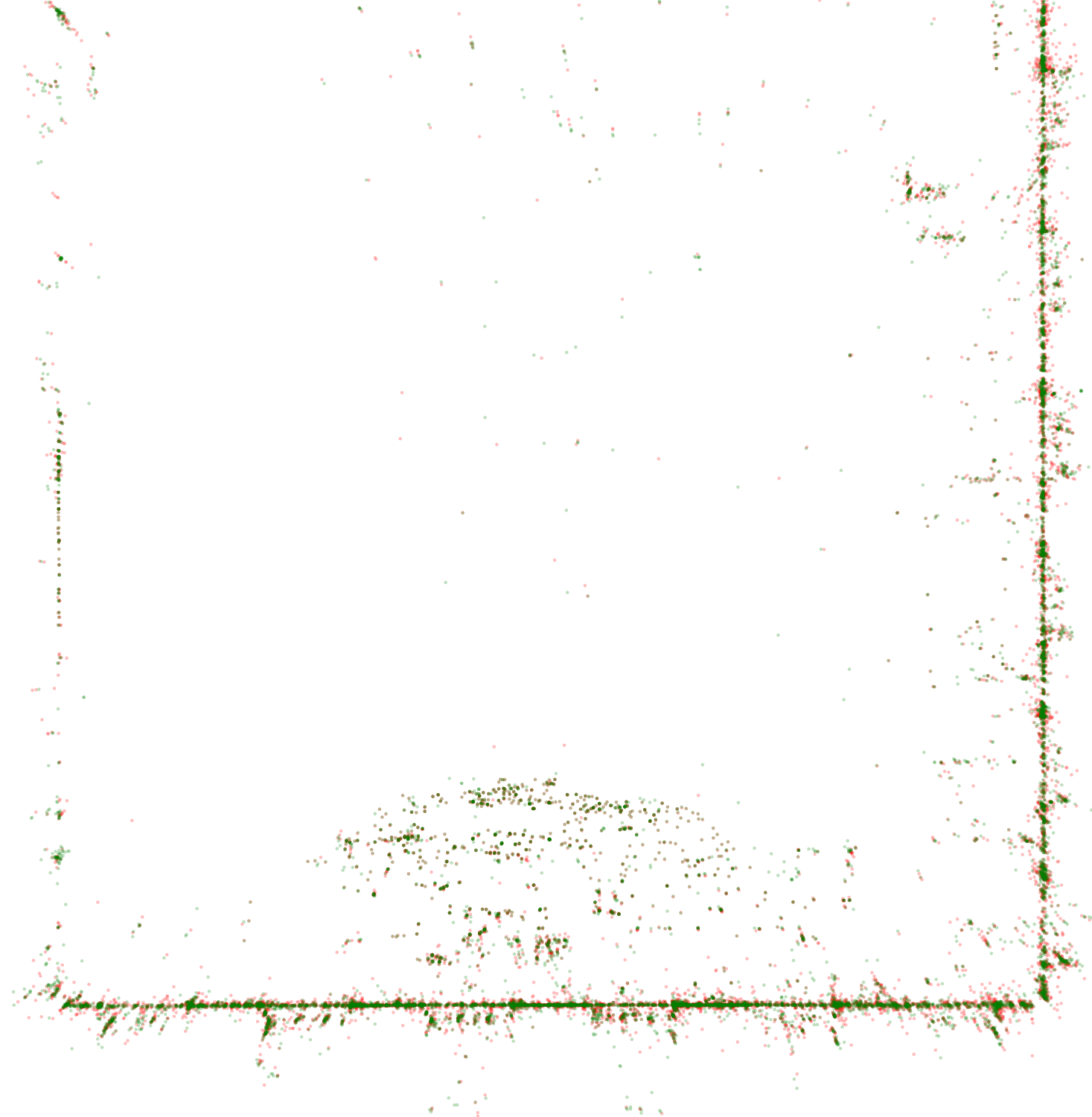} \\
		SuperPoint & & Key.Net \\
		\includegraphics[width=.35\textwidth]{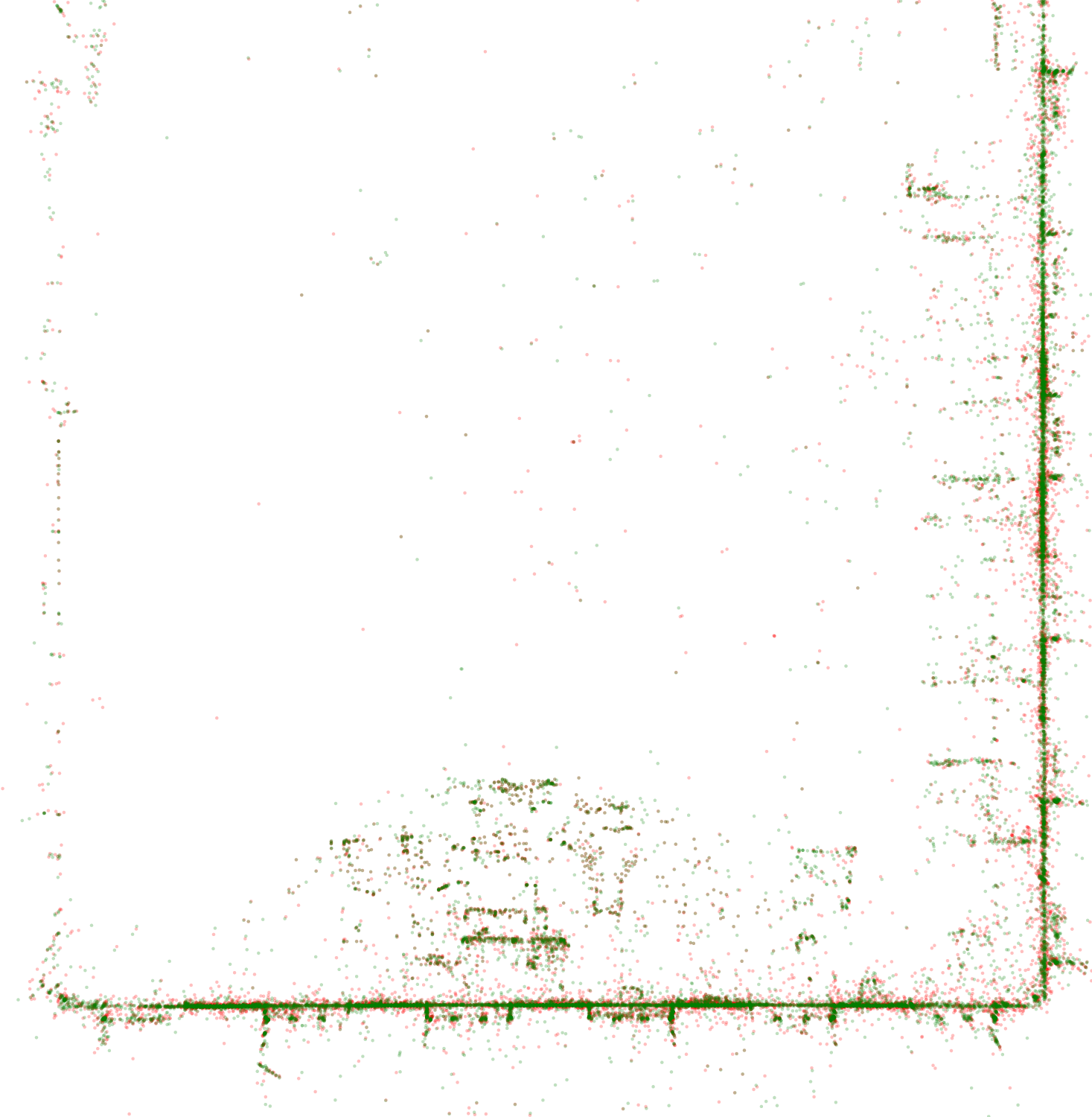} & & \includegraphics[width=.35\textwidth]{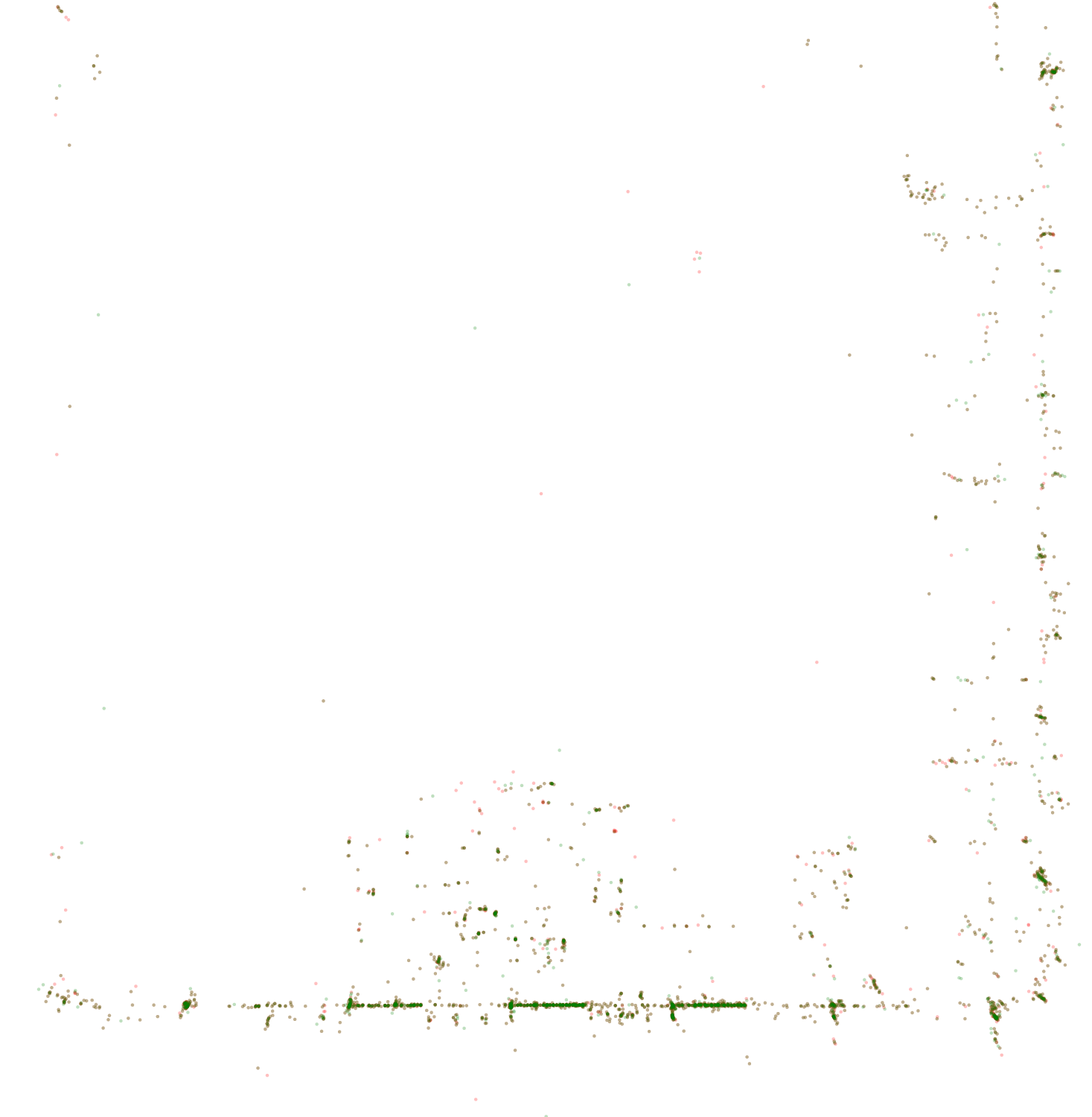} \\
	\end{tabular}
	\caption{{\bf Courtyard.} We show top-down partial views of point clouds triangulated on the Courtyard scene. We overlap the point-cloud obtained from \textcolor{green}{refined keypoints} and the point-cloud from \textcolor{red}{raw keypoints}. The noise levels are drastically reduced nearby planar surfaces. Best viewed on a monitor.}
	\label{fig:courtyard}
\end{figure}

\begin{figure}[p]
	\centering
	\begin{tabular}{c c c}
		SIFT & \hspace{2cm} & SURF \\
		\includegraphics[width=.30\textwidth]{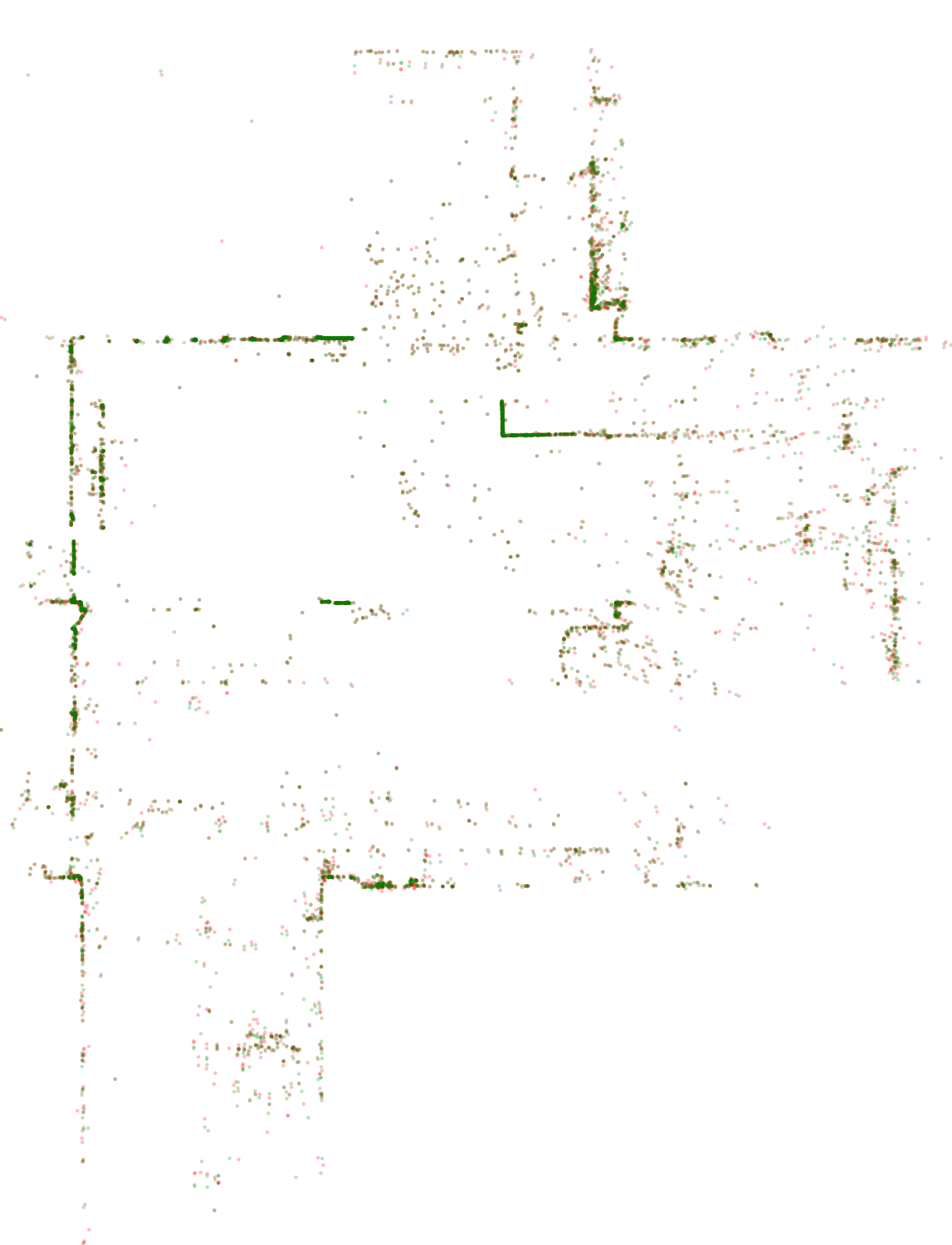} & & \includegraphics[width=.30\textwidth]{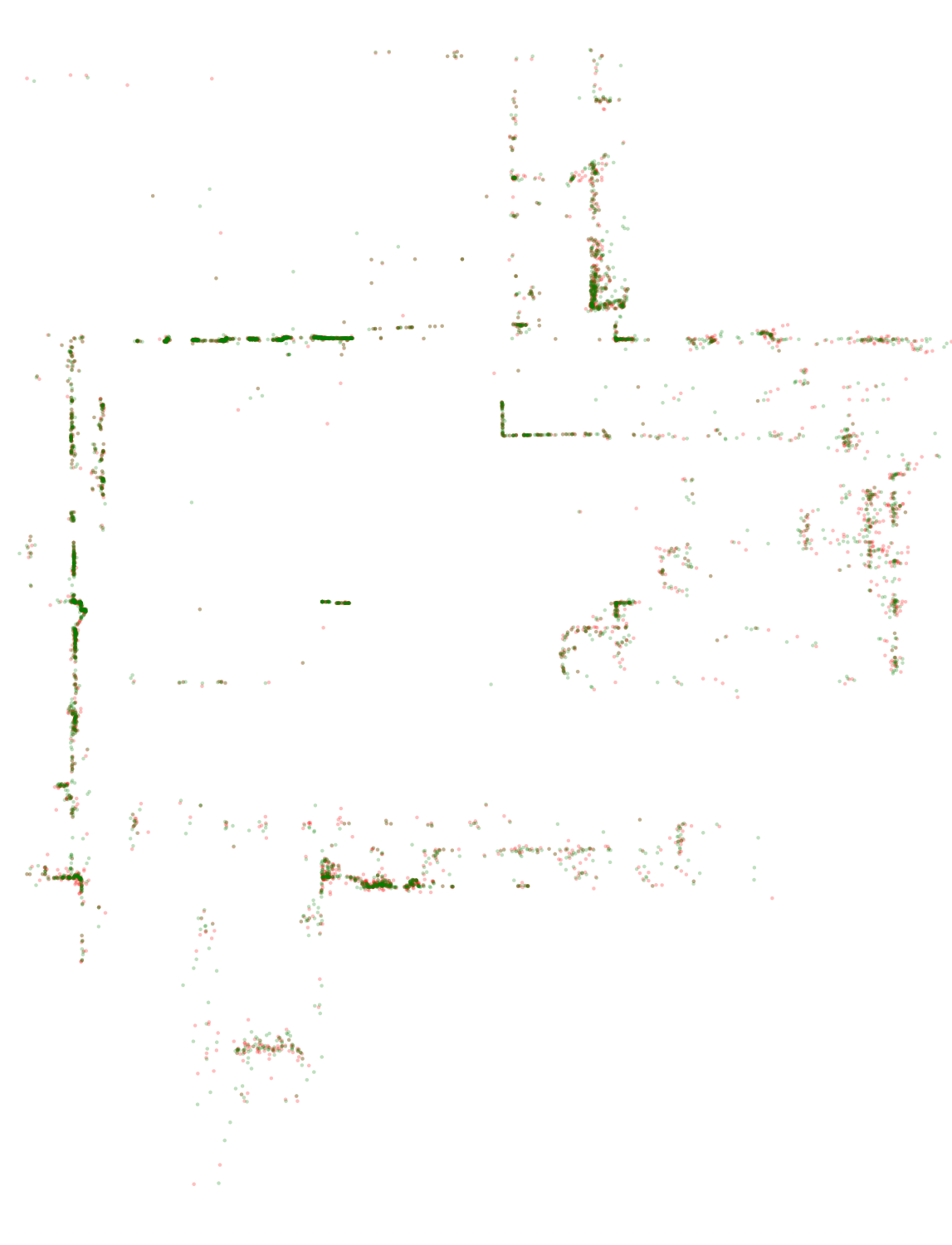} \\
		D2-Net & & R2D2 \\
		\includegraphics[width=.30\textwidth]{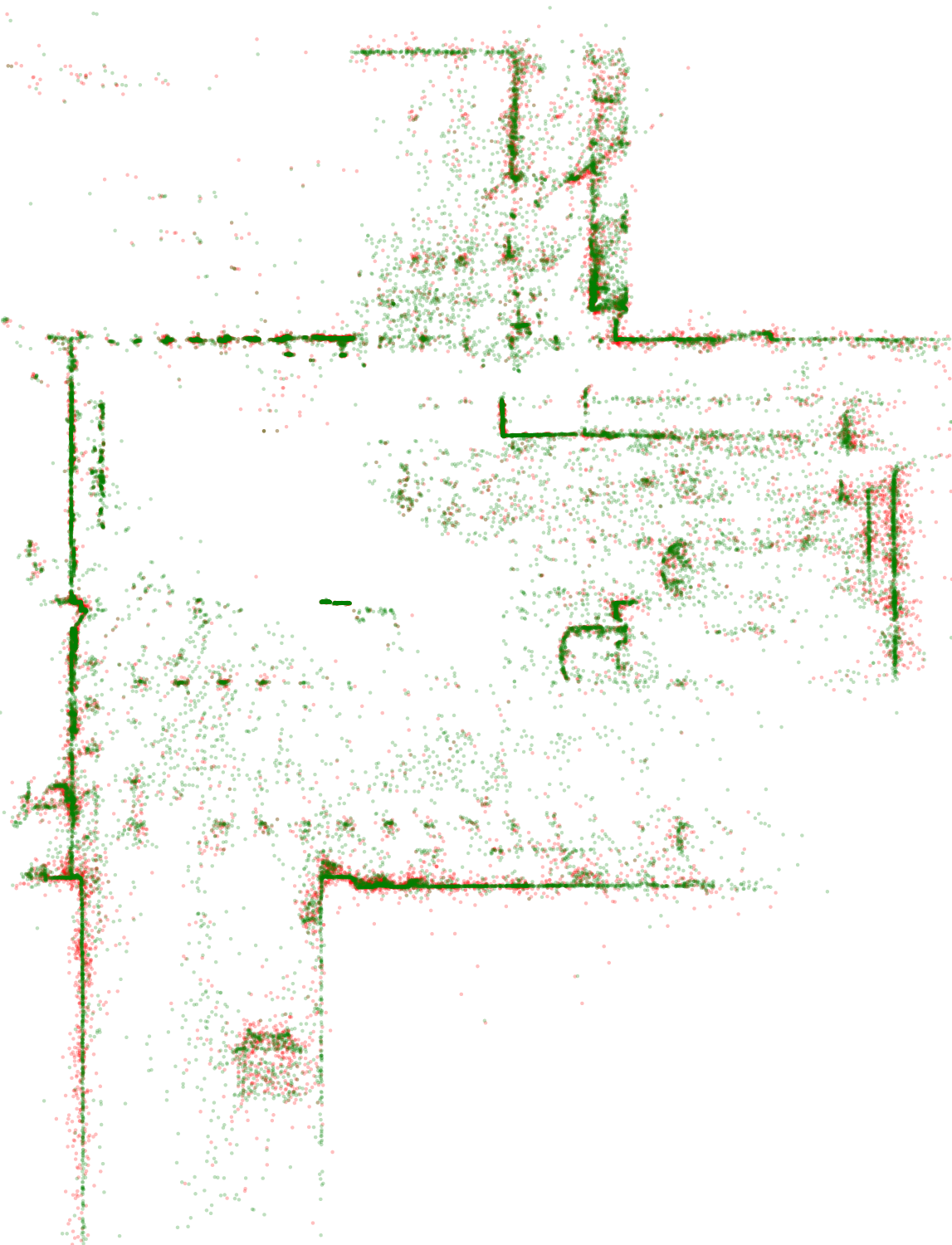} & & \includegraphics[width=.30\textwidth]{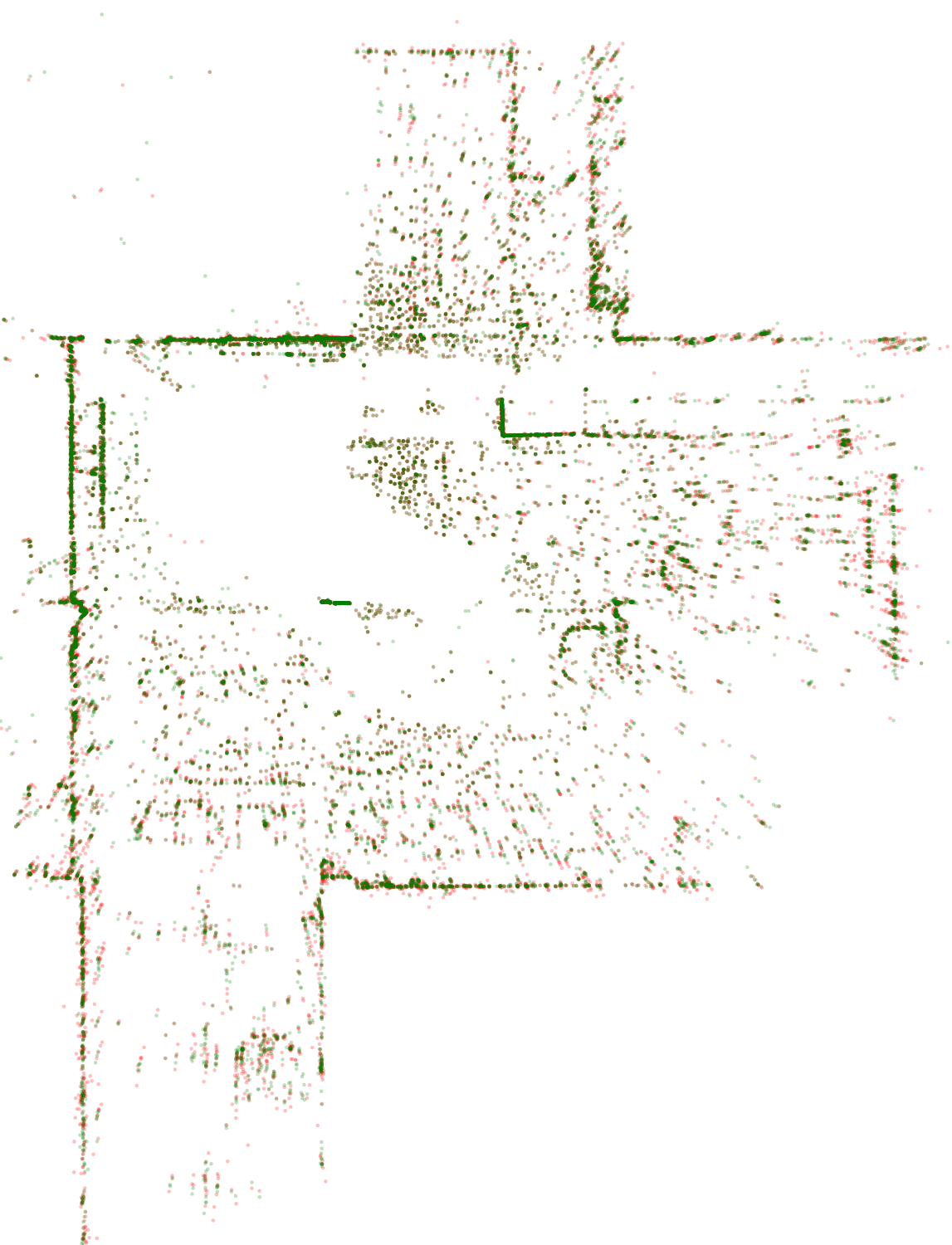} \\
		SuperPoint & & Key.Net \\
		\includegraphics[width=.30\textwidth]{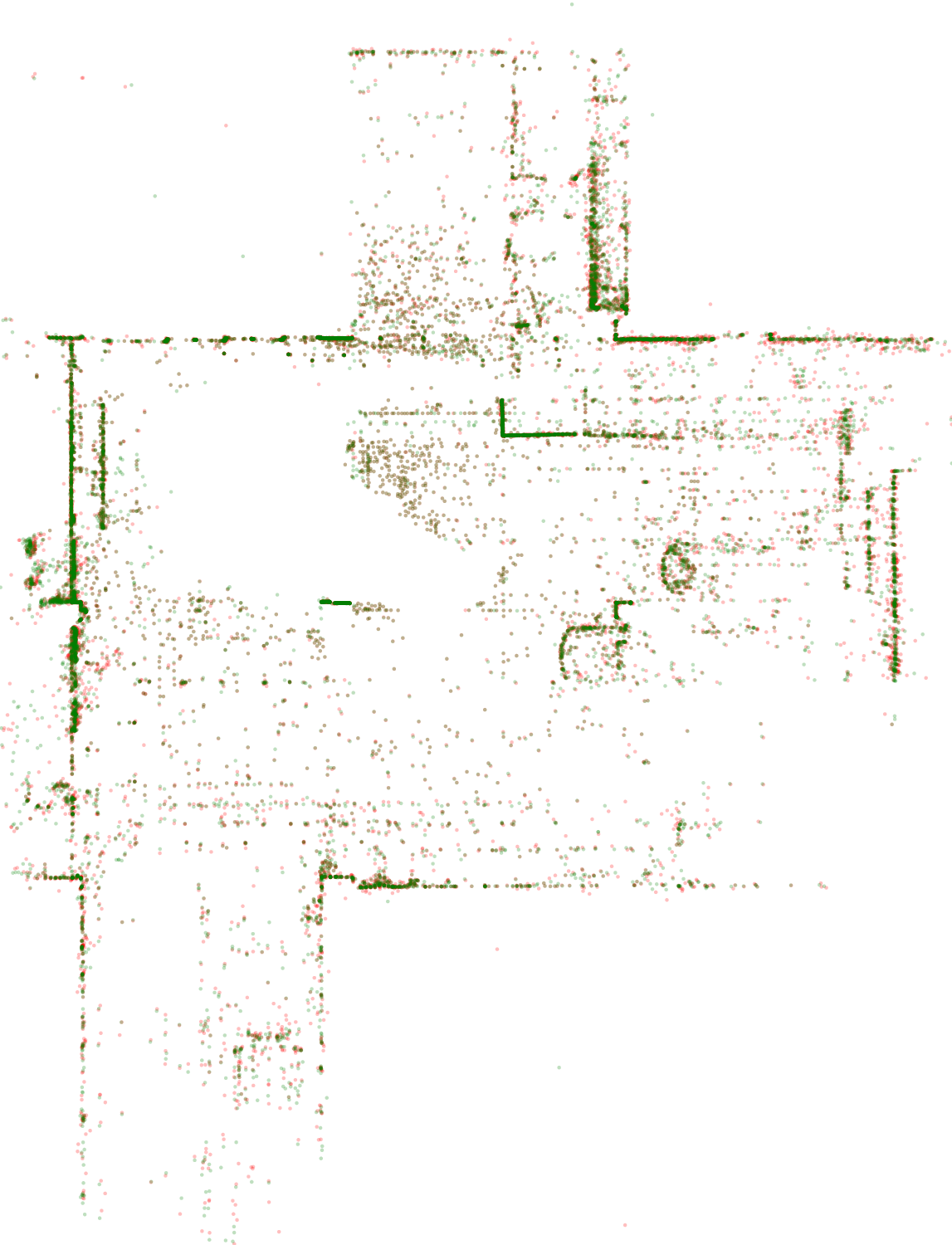} & & \includegraphics[width=.30\textwidth]{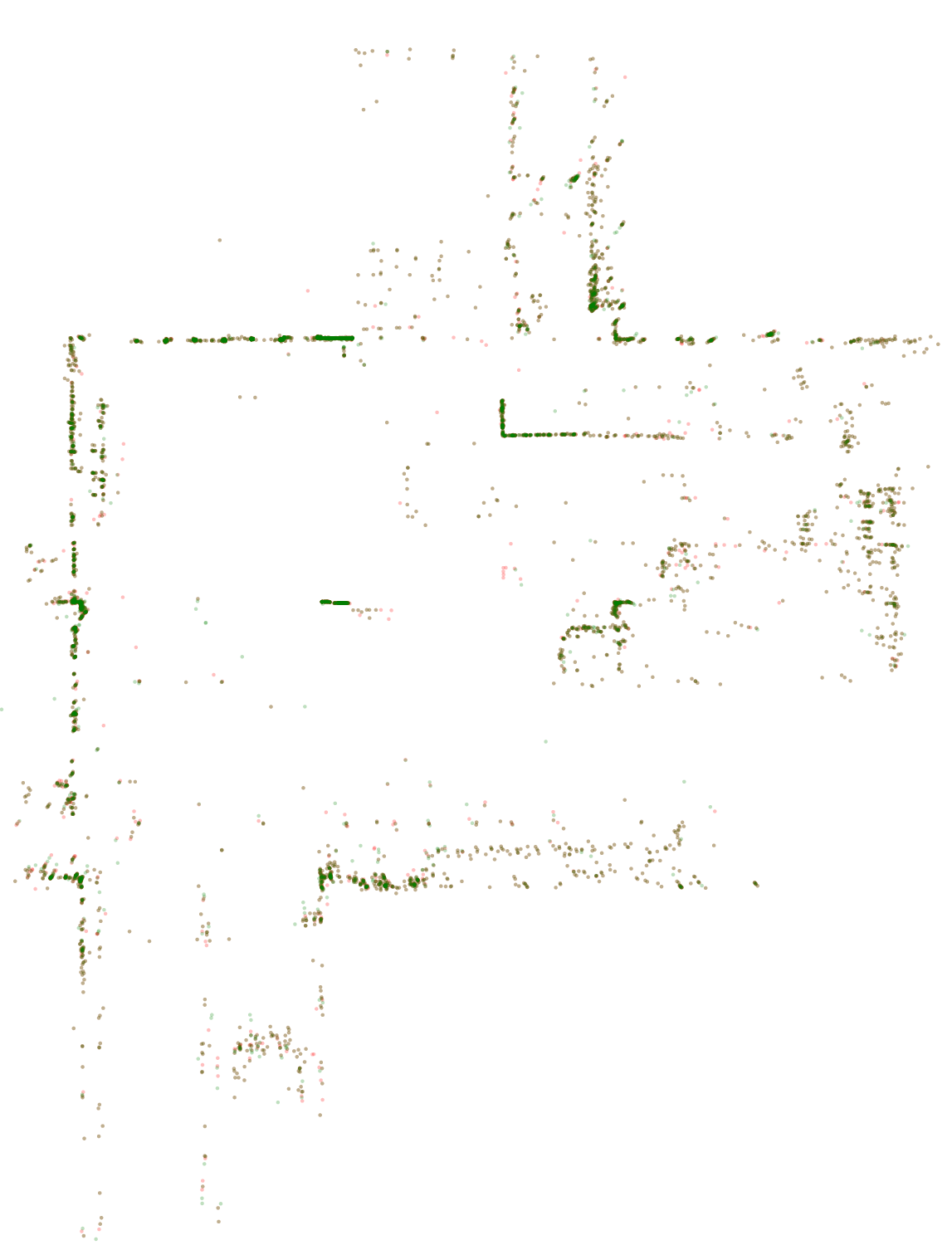} \\
	\end{tabular}
	\caption{{\bf Delivery Area.} We show top-down partial views of point clouds triangulated on the Courtyard scene. We overlap the point-cloud obtained from \textcolor{green}{refined keypoints} and the point-cloud from \textcolor{red}{raw keypoints}. The noise levels are drastically reduced nearby planar surfaces. Best viewed on a monitor.}
	\label{fig:delivery_area}
\end{figure}

Due to space constraints, in the main paper, we only reported the average results on indoor and outdoor scenes for the ETH3D triangulation evaluation~\cite{Schoeps2017A}.
Tables~\ref{tab:eth3d-in-indep}~and~\ref{tab:eth3d-out-indep} show the results for each of the $13$ datasets.
For the learned features, the results with refinement are always better.
For SIFT, the only scene where the results after refinement are worse is Meadow; this is a textureless scene where SIFT has troubles correctly matching features.
Due to the low number of matches passed to COLMAP, its triangulation results are very sensitive to small changes in the input.
Some qualitative examples are shown in Figures~\ref{fig:courtyard} and~\ref{fig:delivery_area}.
A short video with additional examples is available at \url{https://www.youtube.com/watch?v=eH4UNwXLsyk}.

\section{Ablation study}
\label{sec:ablation}

In this section, an ablation study for the proposed refinement procedure will be presented. We will first start by studying the effect of training data on the two-view refinement network. Secondly, we will study how each step of the multi-view refinement influences the final result.

\subsection{Two-view refinement}

\begin{figure}[t]
	\centering
	\begin{minipage}{0.5\textwidth}
		\centering
		\tiny
		\begin{tabular}{c c c c}
			\toprule
			\multirowcell{2}{Layer} & \multirowcell{2}{Batch \\ Norm.} & \multirowcell{2}{ReLU} & \multirowcell{2}{Output \\ shape} \\ 
			& & & \\ \midrule
			\texttt{input}, RGB & & & $33 \times 33 \times 3$ \\ \midrule
			\texttt{conv1\_1}, $3 \times 3$ & & \checkmark & $33 \times 33 \times 64$ \\
			\texttt{conv1\_2}, $3 \times 3$ & & \checkmark & $33 \times 33 \times 64$ \\ \midrule
			\texttt{max\_pool1}, $3 \times 3$, stride $2$ & & & $17 \times 17 \times 64$ \\ \midrule
			\texttt{conv2\_1}, $3 \times 3$ & & \checkmark & $17 \times 17 \times 128$ \\
			\texttt{conv2\_2}, $3 \times 3$ & & \checkmark & $17 \times 17 \times 128$ \\ \midrule
			\texttt{correlation} & & & $17 \times 17 \times 289$ \\ \midrule
			\texttt{reg\_conv1}, $5 \times 5$ & \checkmark & \checkmark & $13 \times 13 \times 128$ \\
			\texttt{reg\_conv2}, $5 \times 5$ & \checkmark & \checkmark & $9 \times 9 \times 128$ \\
			\texttt{reg\_conv3}, $5 \times 5$ & \checkmark & \checkmark & $5 \times 5 \times 64$ \\
			\texttt{reg\_conv4}, $5 \times 5$ & \checkmark & \checkmark & $1 \times 1 \times 64$ \\
			\texttt{reg\_fc} & & & $2$ \\
			\bottomrule
		\end{tabular}
	\end{minipage}
	~
	\begin{minipage}{0.45\textwidth}
		\centering
		\includegraphics[width=.6\textwidth]{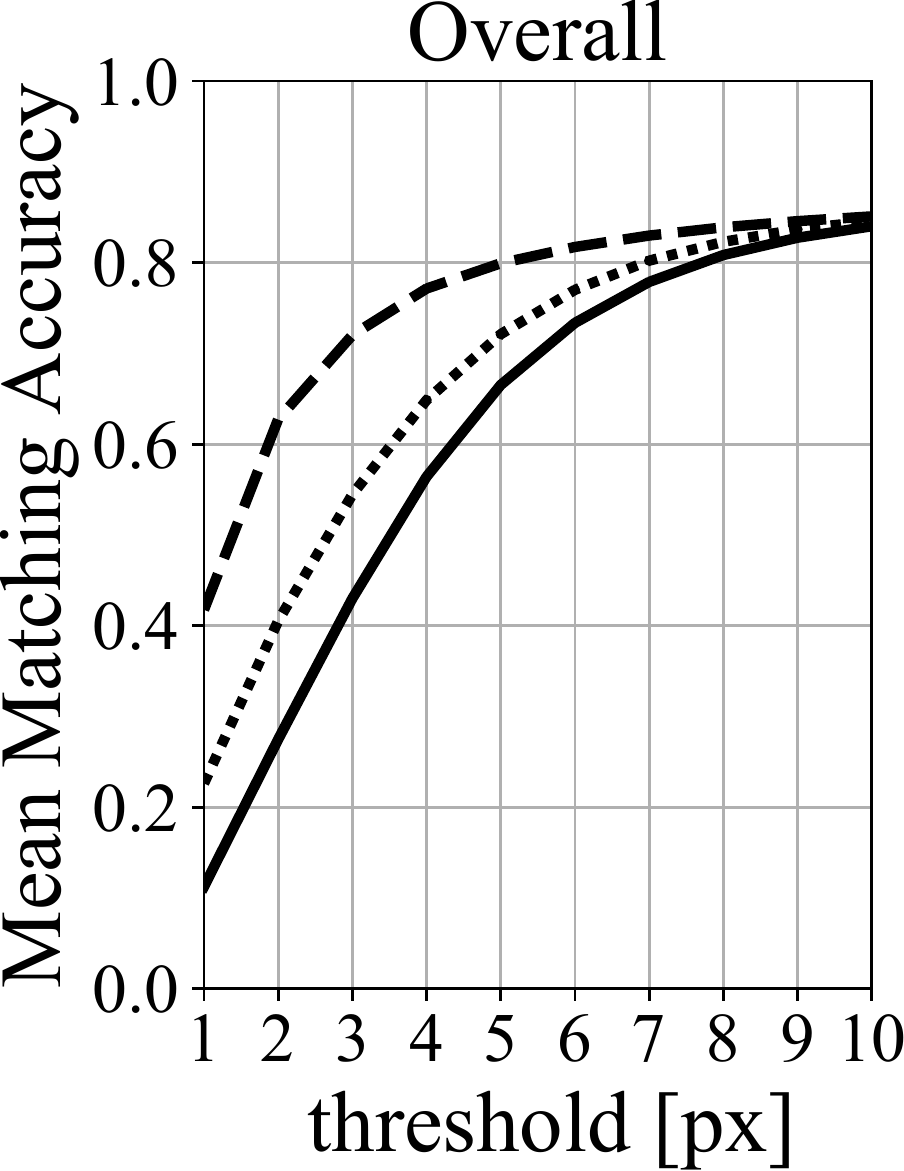}
	\end{minipage}
	\caption{{\bf Two-view refinement.} {\it Left -- architecture:} We use a slightly modified version of VGG16 up to \texttt{conv2\_2} for feature extraction. The results of the dense matching are processed by a sequence of convolutional and fully connected layers. {\it Right -- ablation:} The results for D2-Net without refinement are reported by a solid line. We compare a network trained on synthetic pairs only (dotted) and one trained with both synthetic and real data (dashed).}
	\label{tab:architecture}\label{fig:hpatches_sequences_ablation}
\end{figure}

The architecture used for the two-view refinement between tentative matches is described in Table~\ref{tab:architecture}. For the layers with batch normalization, we place it before the non-linearity (\ie, the order is convolution followed by batch normalization and finally non-linearity) as suggested in the reference paper~\cite{Ioffe2015Batch}.

\setlength{\tabcolsep}{1.5pt}
\setlength{\tabcolsep}{\tabcolsepdefault}

For this ablation study, we focus on the HPatches Sequences dataset~\cite{Balntas2017HPatches}, because it allows to isolate the network output. Given a tentative match $u, v$, we run a forward pass of the patch alignment network to predict $d_{u \rightarrow v}$ and use $u$ and $v + d_{u \rightarrow v}$ as keypoint locations. As can be seen in Figure~\ref{fig:hpatches_sequences_ablation}, training only with synthetic data (\ie, pairs consisting of a patch and a warped version of itself) is not sufficient to achieve the final performance. By using real pairs extracted from the MegaDepth dataset~\cite{Li2018MegaDepth}, we allow the network to learn different illumination conditions as well as occlusions / large viewpoint changes.

\subsection{Multi-view refinement}

We use the largest dataset with ground-truth data available (Facade from ETH3D~\cite{Schoeps2017A}) to study the relevance of the following steps of our pipeline: graph partitioning, inter-edges, $3 \times 3$ displacement grid. The ablation results are summarized in Table~\ref{tab:facade_ablation}. For the purpose of this section, we define the set of intra-edges connecting nodes within a track as $E_\text{intra} = \{ (u \rightarrow v) \in E | t_u = t_v \}$ and the set of inter-edges connecting nodes of different tracks as $E_\text{inter} = \{ (u \rightarrow v) \in E | t_u \neq t_v \}$ on the entire graph $G$ (without graph-cut).

\setlength{\tabcolsep}{1.5pt}
\begin{table}[t]
	\caption{{\bf Multi-view refinement ablation study.} Reconstruction statistics are reported for the Facade scene of ETH3D~\cite{Schoeps2017A} consisting of $76$ images for different formulations of the multi-view optimization problem.}
	\label{tab:facade_ablation}
	\centering
	\tiny
	\begin{tabular}[t]{l l | c c c | c c c | c | c | c}
		\toprule
		\multirow{2}{*}{Method} & & \multicolumn{3}{c |}{Comp. (\%)} & \multicolumn{3}{c |}{Accuracy (\%)} & \multirowcell{2}{Track \\ length} &  \multirowcell{2}{Reproj. \\ error} & \multirowcell{2}{Optim. \\ runtime} \\
		& & $1$cm & $2$cm & $5$cm & $1$cm & $2$cm & $5$cm & & \\ \midrule
		\multirow{8}{*}{SIFT} & no refinement & $0.06$ & $0.36$ & $3.08$ & $36.04$ & $52.10$ & $73.28$ & $5.42$ & $1.07$ & \\ \cmidrule{2-11}
		& no graph partitioning & $0.09$ & $0.50$ & $3.85$ & $44.52$ & $62.26$ & $82.74$ & $5.86$ & $0.81$ & $49.7$s \\ \cmidrule{2-11}
		& intra-edges & $0.09$ & $0.51$ & $3.84$ & $45.16$ & $62.56$ & $82.44$ & $5.80$ & $0.80$ & $10.2$s \\
		& + inter-edges & $0.09$ & $0.50$ & $3.83$ & $45.46$ & $62.58$ & $82.65$ & $5.82$ & $0.80$ & $54.1$s\\
		& + graph-cut {\it (full)} & $0.09$ & $0.50$ & $3.82$ & $45.19$ & $62.32$ & $82.38$ & $5.81$ & $0.80$ & $13.3$s \\ \cmidrule{2-11}
		& {\it full} (constant flow) & $0.09$ & $0.49$ & $3.72$ & $44.12$ & $61.33$ & $80.65$ & $5.75$ & $0.84$ & $11.9$s \\ \midrule
		\multirow{8}{*}{D2-Net} & no refinement & $0.02$ & $0.18$ & $2.26$ & $7.56$ & $14.21$ & $29.90$ & $3.20$ & $1.60$ & \\ \cmidrule{2-11}
		& no graph partitioning & $0.11$ & $0.71$ & $5.50$ & $28.20$ & $43.64$ & $67.50$ & $5.64$ & $1.09$ & $317.7$s \\ \cmidrule{2-11}
		& intra-edges & $0.16$ & $1.01$ & $8.17$ & $34.85$ & $53.05$ & $75.80$ & $5.05$ & $0.85$ & $20.0$s \\
		& + inter-edges & $0.16$ & $1.01$ & $8.16$ & $34.88$ & $53.18$ & $75.90$ & $5.06$ & $0.85$ & $223.6$s\\
		& + graph-cut {\it (full)} & $0.16$ & $1.01$ & $8.18$ & $34.86$ & $53.32$ & $76.02$ & $5.06$ & $0.85$ & $31.1$s \\ \cmidrule{2-11}
		& {\it full} (constant flow) & $0.13$ & $0.87$ & $7.53$ & $29.37$ & $46.05$ & $69.51$ & $4.78$ & $0.99$ & $28.1$s \\ \midrule
		\multirow{8}{*}{SuperPoint} & no refinement & $0.07$ & $0.49$ & $4.95$ & $18.82$ & $32.21$ & $54.72$ & $4.21$ & $1.54$ & \\ \cmidrule{2-11}
		& no graph partitioning & $0.09$ & $0.62$ & $5.54$ & $25.77$ & $41.67$ & $66.16$ & $4.95$ & $1.28$ & $246.7$s \\ \cmidrule{2-11}
		& intra-edges & $0.14$ & $0.90$ & $7.21$ & $35.16$ & $53.12$ & $74.72$ & $5.23$ & $0.94$ & $32.4$s \\
		& + inter-edges & $0.14$ & $0.89$ & $7.21$ & $35.73$ & $53.36$ & $75.32$ & $5.31$ & $0.93$ & $255.8$s \\
		& + graph-cut {\it (full)} & $0.14$ & $0.90$ & $7.23$ & $35.73$ & $53.49$ & $75.48$ & $5.25$ & $0.94$ & $47.6$s\\ \cmidrule{2-11}
		& {\it full} (constant flow) & $0.12$ & $0.78$ & $6.66$ & $30.41$ & $47.57$ & $69.90$ & $5.12$ & $1.06$ & $43.2$s \\
		\bottomrule
	\end{tabular}
\end{table}
\setlength{\tabcolsep}{\tabcolsepdefault}

Without any graph partitioning, the optimization can be formulated as:
\begin{equation}
	\begin{split}
		\min_{x_p} & \sum_{(u \rightarrow v) \in E} s_{u \rightarrow v} \rho(\lVert \bar{x}_v - \bar{x}_u - T_{u \rightarrow v}(\bar{x}_u) \rVert^2) \\
		\text{s.t.} & \lVert \bar{x}_p \rVert_1 = \lVert x_p - x_p^0 \rVert_1 \leq K, \forall p \enspace .
	\end{split}
	\label{eq:no_partitioning}
\end{equation}
Despite the long track length, the reprojection error is generally larger and the point clouds are less accurate - this is mainly due to wrong tentative matches. Moreover, this formulation has one of the highest optimization runtimes.

After partitioning the graph into tracks, one could ignore the inter-edges:
\begin{equation}
	\begin{split}
		\min_{x_p} & \sum_{(u \rightarrow v) \in E_{\text{intra}}} s_{u \rightarrow v} \rho(\lVert \bar{x}_v - \bar{x}_u - T_{u \rightarrow v}(\bar{x}_u) \rVert^2) \\
		\text{s.t.} & \lVert \bar{x}_p \rVert_1 = \lVert x_p - x_p^0 \rVert_1 \leq K, \forall p \enspace .
	\end{split}
	\label{eq:intra}
\end{equation}
This formulation can be solved independently for each track and is thus the fastest. However, detectors often fire multiple times for the same visual feature. Since we restrict the tracks to only contain one feature from each image, these multiple detections will never be merged into a single point.

To address this, the inter-edges must be considered:
\begin{equation}
	\begin{split}
		\min_{x_p} & \sum_{(u \rightarrow v) \in E_\text{intra}} s_{u \rightarrow v} \rho(\lVert \bar{x}_v - \bar{x}_u - T_{u \rightarrow v}(\bar{x}_u) \rVert^2) + \\
		& \sum_{(u \rightarrow v) \in E_\text{inter}} s_{u \rightarrow v} \psi(\lVert \bar{x}_v - \bar{x}_u - T_{u \rightarrow v}(\bar{x}_u) \rVert^2) \\
		\text{s.t.} & \lVert \bar{x}_p \rVert_1 = \lVert x_p - x_p^0 \rVert_1 \leq K, \forall p \enspace .
	\end{split}
	\label{eq:intra_inter}
\end{equation}
The main issue with this formulation is its runtime due to having the same number of residuals as Equation~\ref{eq:no_partitioning}. However, it generally achieves similar accuracy and reprojection error to Equation~\ref{eq:intra} while having a better track length.

By using recursive graph cut to split the connected components into smaller sets and solving on each remaining component independently, we strike a balance between the performance of Equation~\ref{eq:intra_inter} and the efficiency of Equation~\ref{eq:intra}.

While the constant flow assumption also improves the performance of local features, it is not sufficient to explain all structures. The $3 \times 3$ deformation grid is better suited and achieves a superior performance across the board.

\begin{figure}[t]
	\centering
	\includegraphics[width=0.75\textwidth]{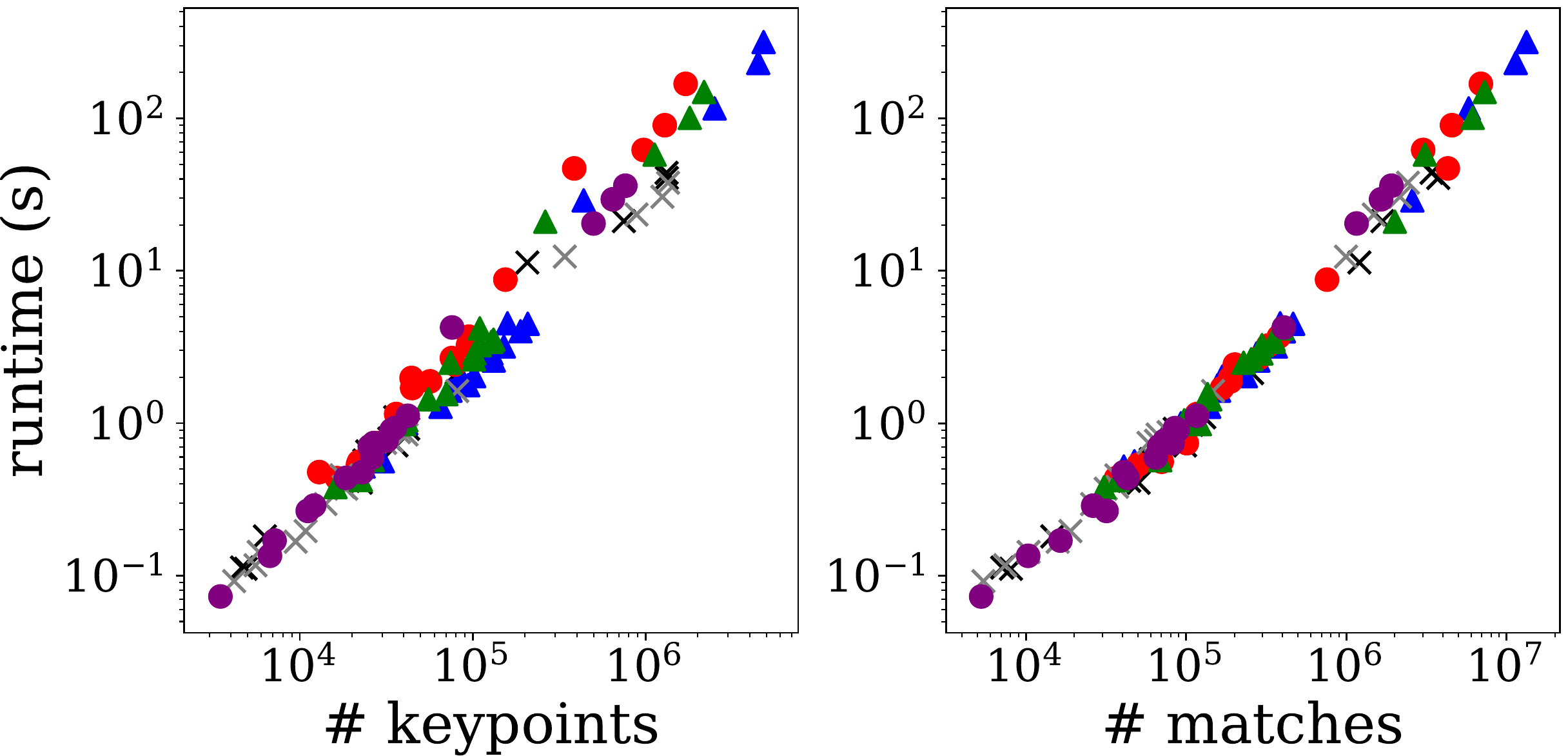}
	\caption{{\bf Graph optimization runtime.} The runtime is plotted as a function of the number of keypoints and matches. We respect the color coding from the main paper: \textcolor{black}{SIFT}~\cite{Lowe2004Distinctive}, \textcolor{gray}{SURF}~\cite{Bay2006SURF}, \textcolor{blue}{D2-Net}~\cite{Dusmanu2019D2}, \textcolor{green}{R2D2}~\cite{Revaud2019R2D2}, \textcolor{red}{SuperPoint}~\cite{Detone2018SuperPoint}, and \textcolor{purple}{Key.Net}~\cite{Barroso-Laguna2019KeyNet}.}
	\label{fig:runtime}
\end{figure}

The runtime of the proposed graph optimization procedure is shown in Figure~\ref{fig:runtime} for all datasets of our evaluation. The top-right points correspond to the internet reconstruction from the Local Feature Evaluation Benchmark~\cite{Schonberger2017Comparative} (Madrid Metropolis, Gendarmenmarkt, Tower of London). For these datasets, the runtime remains low ($1-5$ minutes depending on the method) compared to the runtime of the sparse 3D reconstruction ($15-30$ minutes).

\section{Query refinement}
\label{sec:query}

For the localization experiments, we used the tentative matches $\{u_1, u_2, \dots\}$ of each query feature $q$ to refine its location. First, all matches corresponding to non-triangulated features are discarded since they cannot be used for PnP. For each remaining match $u_i \leftrightarrow q$, let $\pi_i$ be the 3D point associated to $u_i$ and $\hat{u}_i$ be the reprojection of $\pi_i$ to the image of $u_i$.

Since these matches are purely based on appearance, the points $u_i$ might correspond to different 3D points of the partial model. Each matching 3D location $\Pi$ is considered as an independent hypothesis. Given that the reprojected locations are fixed, the optimization problem can be simplified by considering only one-directional $u_i \rightarrow q$ edges:
\begin{equation}
	\begin{split}
		\min_{x_q} & \sum_{u_i \text{ s.t. } \pi_i = \Pi} s_{u_i \rightarrow q} \rho(\lVert \bar{x}_q - d_{\hat{u}_i \rightarrow q} \rVert^2) \\
		\text{s.t.} & \lVert \bar{x}_q \rVert_1 = \lVert x_q - x_q^0 \rVert_1 \leq K \enspace .
	\end{split}
\end{equation}
The central point flow in the above formulation is considered from the reprojected feature $\hat{u}_i$ in a view of the partial model to the query feature $q$.

After removing the robustifier and supposing that the two-view displacements are always smaller than $K$, the problem can be rewritten as follows:
\begin{equation}
	\begin{split}
		\min_{x_q} & \sum_{u_i \text{ s.t. } \pi_i = \Pi} s_{u_i \rightarrow q} \lVert \bar{x}_q - d_{\hat{u}_i \rightarrow q} \rVert^2 \enspace .
	\end{split}
\end{equation}
This formulation has a closed-form solution:
\begin{equation}
	x_q^\Pi = x_q^0 + \frac{\sum\limits_{u_i \text{ s.t. } \pi_i = \Pi} s_{u_i \rightarrow q} d_{\hat{u}_i \rightarrow q}}{\sum\limits_{u_i \text{ s.t. } \pi_i = \Pi} s_{u_i \rightarrow q}} \enspace .
\end{equation}

Thus, for each query feature $q$ with triangulated tentative matches, we obtain one or more refined 2D-3D correspondences $(x_q^\Pi, \Pi)$ which can be used for pose estimation.

\section{Training dataset}
\label{sec:training}

As mentioned in the main paper, several steps were taken to improve the quality of the training data extracted from MegaDepth~\cite{Li2018MegaDepth}.

\PAR{Scene filtering.} We discarded 16 scenes due to inconsistencies between sparse and dense reconstructions. This was done automatically using the following heuristic: from each scene, $100000$ random pairs of matching 2D observations part of the 3D model were selected; for each such pair $(k_1, k_2)$, the Multi-View Stereo (MVS) depth was used to warp $k_1$ to the other image obtaining $\hat{k}_2$; a keypoint is inconsistent if its reprojection $\hat{k}_2$ is more than $12$ pixels away from its feature position $k_2$, \ie, $\lvert \hat{k}_2 - k_2 \rvert > 12$. The following scenes were removed for having a low number of consistent points: \texttt{0000}, \texttt{0002}, \texttt{0011}, \texttt{0020}, \texttt{0033}, \texttt{0050}, \texttt{0103}, \texttt{0105}, \texttt{0143}, \texttt{0176}, \texttt{0177}, \texttt{0265}, \texttt{0366}, \texttt{0474}, \texttt{0860}, \texttt{4541}.

\PAR{Depth consistency.} We enforce depth consistency to make sure that the central pixel is not occluded. The MVS depth $D_1$ of a source image is used to back-project a keypoint $k_1$ to 3D and obtain $p$. We then reproject this 3D point to the target image to obtain $\hat{k}_2$ and depth $d$. The depth consistency verifies that the MVS depth from the second image $D_2$ is consistent with the 3D point $p$, \ie, $\lvert D_2(\hat{k}_2) - d \rvert < 10^{-2}$.

\setlength{\tabcolsep}{1pt}
\begin{table}[p]
	\caption{{\bf ETH3D triangulation evaluation - Indoors.} We report triangulation statistics on each indoor dataset for methods with and without refinement.}
	\label{tab:eth3d-in-indep}
	\centering
	\tiny
	\begin{tabular}[t]{c | l | c c c | c c c | l | c c c | c c c}
		\toprule
		\multirow{2}{*}{Dataset} & \multirow{2}{*}{Method} & \multicolumn{3}{c |}{Comp. (\%)} & \multicolumn{3}{c |}{Accuracy (\%)} & \multirow{2}{*}{Method} & \multicolumn{3}{c |}{Comp. (\%)} & \multicolumn{3}{c}{Accuracy (\%)} \\
		& & $1$cm & $2$cm & $5$cm & $1$cm & $2$cm & $5$cm & & $1$cm & $2$cm & $5$cm & $1$cm & $2$cm & $5$cm \\ \midrule
		\multirowcell{8}{\textit{Deliv. Area}\\$44$ images} & SIFT & $0.06$ & $0.34$ & $2.29$ & $61.59$ & $74.40$ & $86.98$ & SURF & $0.03$ & $0.20$ & $1.35$ & $53.91$ & $70.15$ & $83.18$ \\
		& SIFT + ref. & $\mathbf{0.09}$ & $\mathbf{0.44}$ & $\mathbf{2.66}$ & $\mathbf{71.65}$ & $\mathbf{82.47}$ & $\mathbf{91.64}$ & SURF + ref. & $\mathbf{0.06}$ & $\mathbf{0.30}$ & $\mathbf{1.76}$ & $\mathbf{68.67}$ & $\mathbf{80.26}$ & $\mathbf{89.72}$ \\ \cmidrule{2-15}
		& D2-Net & $0.08$ & $0.53$ & $3.53$ & $30.99$ & $47.16$ & $67.35$ & R2D2 & $0.17$ & $0.86$ & $5.26$ & $52.09$ & $66.80$ & $82.29$ \\
		& D2-Net + ref. & $\mathbf{0.40}$ & $\mathbf{1.93}$ & $\mathbf{9.87}$ & $\mathbf{65.00}$ & $\mathbf{77.26}$ & $\mathbf{88.51}$ & R2D2 + ref. & $\mathbf{0.27}$ & $\mathbf{1.11}$ & $\mathbf{5.81}$ & $\mathbf{70.57}$ & $\mathbf{81.63}$ & $\mathbf{91.29}$ \\ \cmidrule{2-15}
		& S & $0.15$ & $0.80$ & $5.36$ & $56.80$ & $71.42$ & $85.10$ & Key.Net & $0.05$ & $0.28$ & $1.78$ & $52.08$ & $69.11$ & $85.33$ \\
		& SP + ref. & $\mathbf{0.22}$ & $\mathbf{1.07}$ & $\mathbf{6.36}$ & $\mathbf{74.39}$ & $\mathbf{85.36}$ & $\mathbf{93.89}$ & Key.Net + ref. & $\mathbf{0.09}$ & $\mathbf{0.38}$ & $\mathbf{2.15}$ & $\mathbf{74.01}$ & $\mathbf{84.16}$ & $\mathbf{92.56}$ \\ \midrule
		\multirowcell{8}{\textit{Kicker}\\$31$ images} & SIFT & $0.27$ & $1.29$ & $5.64$ & $71.78$ & $82.69$ & $91.63$ & SURF & $0.22$ & $1.08$ & $4.78$ & $65.20$ & $77.94$ & $90.31$ \\
		& SIFT + ref. & $\mathbf{0.33}$ & $\mathbf{1.44}$ & $\mathbf{5.92}$ & $\mathbf{77.32}$ & $\mathbf{86.61}$ & $\mathbf{93.90}$ & SURF + ref. & $\mathbf{0.31}$ & $\mathbf{1.34}$ & $\mathbf{5.31}$ & $\mathbf{77.43}$ & $\mathbf{86.12}$ & $\mathbf{93.82}$ \\ \cmidrule{2-15}
		& D2-Net & $0.20$ & $1.16$ & $6.18$ & $38.41$ & $56.54$ & $75.83$ & R2D2 & $0.46$ & $1.87$ & $8.41$ & $68.08$ & $80.30$ & $89.91$ \\
		& D2-Net + ref. & $\mathbf{0.87}$ & $\mathbf{3.51}$ & $\mathbf{11.20}$ & $\mathbf{69.53}$ & $\mathbf{79.72}$ & $\mathbf{88.16}$ & R2D2 + ref. & $\mathbf{0.56}$ & $\mathbf{2.12}$ & $\mathbf{8.89}$ & $\mathbf{75.12}$ & $\mathbf{84.28}$ & $\mathbf{91.48}$ \\ \cmidrule{2-15}
		& SP & $0.44$ & $2.08$ & $9.24$ & $67.88$ & $79.43$ & $89.01$ & Key.Net & $0.18$ & $0.84$ & $4.28$ & $62.94$ & $79.51$ & $90.44$ \\
		& SP + ref. & $\mathbf{0.57}$ & $\mathbf{2.46}$ & $\mathbf{10.05}$ & $\mathbf{79.23}$ & $\mathbf{87.04}$ & $\mathbf{92.01}$ & Key.Net + ref. & $\mathbf{0.25}$ & $\mathbf{1.07}$ & $\mathbf{4.85}$ & $\mathbf{72.73}$ & $\mathbf{84.31}$ & $\mathbf{92.92}$ \\ \midrule
		\multirowcell{8}{\textit{Office}\\$26$ images} & SIFT & $0.11$ & $0.53$ & $\mathbf{2.72}$ & $75.48$ & $84.81$ & $93.30$ & SURF & $0.06$ & $0.26$ & $1.36$ & $70.50$ & $\mathbf{86.47}$ & $95.17$ \\
		& SIFT + ref. & $\mathbf{0.12}$ & $\mathbf{0.55}$ & $2.64$ & $\mathbf{77.27}$ & $\mathbf{86.98}$ & $\mathbf{94.69}$ & SURF + ref. & $\mathbf{0.07}$ & $\mathbf{0.34}$ & $\mathbf{1.58}$ & $\mathbf{70.72}$ & $85.57$ & $\mathbf{96.01}$ \\ \cmidrule{2-15}
		& D2-Net & $0.12$ & $0.76$ & $3.76$ & $38.78$ & $57.62$ & $82.58$ & R2D2 & $0.33$ & $1.45$ & $6.02$ & $54.97$ & $70.53$ & $87.52$ \\
		& D2-Net + ref. & $\mathbf{0.54}$ & $\mathbf{2.08}$ & $\mathbf{6.21}$ & $\mathbf{65.46}$ & $\mathbf{79.30}$ & $\mathbf{91.47}$ & R2D2 + ref. & $\mathbf{0.42}$ & $\mathbf{1.66}$ & $\mathbf{6.53}$ & $\mathbf{61.07}$ & $\mathbf{75.67}$ & $\mathbf{89.38}$ \\ \cmidrule{2-15}
		& SP & $0.27$ & $1.19$ & $5.27$ & $75.36$ & $85.47$ & $95.46$ & Key.Net & $0.11$ & $0.53$ & $2.73$ & $63.14$ & $77.22$ & $90.43$ \\
		& SP + ref. & $\mathbf{0.34}$ & $\mathbf{1.37}$ & $\mathbf{5.46}$ & $\mathbf{84.05}$ & $\mathbf{91.69}$ & $\mathbf{96.96}$ & Key.Net + ref. & $\mathbf{0.17}$ & $\mathbf{0.71}$ & $\mathbf{3.19}$ & $\mathbf{80.63}$ & $\mathbf{90.29}$ & $\mathbf{95.87}$ \\ \midrule
		\multirowcell{8}{\textit{Pipes}\\$14$ images} & SIFT & $0.06$ & $0.27$ & $1.11$ & $73.23$ & $80.52$ & $87.53$ & SURF & $0.02$ & $0.10$ & $0.52$ & $66.90$ & $74.19$ & $90.30$ \\
		& SIFT + ref. & $\mathbf{0.08}$ & $\mathbf{0.34}$ & $\mathbf{1.50}$ & $\mathbf{80.66}$ & $\mathbf{86.61}$ & $\mathbf{93.52}$ & SURF + ref. & $\mathbf{0.03}$ & $\mathbf{0.14}$ & $\mathbf{0.64}$ & $\mathbf{77.65}$ & $\mathbf{84.04}$ & $\mathbf{91.84}$ \\ \cmidrule{2-15}
		& D2-Net & $0.14$ & $0.76$ & $3.53$ & $54.80$ & $76.15$ & $91.93$ & R2D2 & $0.22$ & $0.97$ & $4.83$ & $68.50$ & $79.42$ & $87.92$ \\
		& D2-Net + ref. & $\mathbf{0.59}$ & $\mathbf{2.08}$ & $\mathbf{5.69}$ & $\mathbf{87.10}$ & $\mathbf{93.23}$ & $\mathbf{97.50}$ & R2D2 + ref. & $\mathbf{0.31}$ & $\mathbf{1.19}$ & $\mathbf{5.21}$ & $\mathbf{75.22}$ & $\mathbf{82.71}$ & $\mathbf{88.68}$ \\ \cmidrule{2-15}
		& SP & $0.41$ & $1.77$ & $7.30$ & $85.31$ & $90.70$ & $\mathbf{96.23}$ & Key.Net & $0.05$ & $0.24$ & $1.27$ & $76.85$ & $87.68$ & $93.31$ \\
		& SP + ref. & $\mathbf{0.55}$ & $\mathbf{2.17}$ & $\mathbf{8.25}$ & $\mathbf{91.15}$ & $\mathbf{94.15}$ & $96.07$ & Key.Net + ref. & $\mathbf{0.07}$ & $\mathbf{0.30}$ & $\mathbf{1.55}$ & $\mathbf{82.15}$ & $\mathbf{92.89}$ & $\mathbf{95.36}$ \\ \midrule
		\multirowcell{8}{\textit{Relief}\\$31$ images} & SIFT & $0.30$ & $1.35$ & $5.19$ & $81.88$ & $91.02$ & $96.29$ & SURF & $0.09$ & $0.46$ & $2.20$ & $73.72$ & $86.39$ & $94.79$ \\
		& SIFT + ref. & $\mathbf{0.35}$ & $\mathbf{1.46}$ & $\mathbf{5.43}$ & $\mathbf{86.59}$ & $\mathbf{92.80}$ & $\mathbf{96.61}$ & SURF + ref. & $\mathbf{0.13}$ & $\mathbf{0.55}$ & $\mathbf{2.40}$ & $\mathbf{83.11}$ & $\mathbf{89.60}$ & $\mathbf{95.07}$ \\ \cmidrule{2-15}
		& D2-Net & $0.45$ & $2.51$ & $9.29$ & $46.72$ & $67.65$ & $88.16$ & R2D2 & $0.52$ & $2.16$ & $9.86$ & $71.12$ & $85.64$ & $95.50$ \\
		& D2-Net + ref. & $\mathbf{1.82}$ & $\mathbf{6.45}$ & $\mathbf{16.58}$ & $\mathbf{87.71}$ & $\mathbf{92.03}$ & $\mathbf{95.33}$ & R2D2 + ref. & $\mathbf{0.70}$ & $\mathbf{2.48}$ & $\mathbf{10.45}$ & $\mathbf{87.07}$ & $\mathbf{93.13}$ & $\mathbf{96.89}$ \\ \cmidrule{2-15}
		& SP & $0.49$ & $2.25$ & $9.17$ & $77.73$ & $88.05$ & $95.52$ & Key.Net & $0.14$ & $0.66$ & $3.23$ & $65.82$ & $80.47$ & $92.78$ \\
		& SP + ref. & $\mathbf{0.60}$ & $\mathbf{2.49}$ & $\mathbf{9.75}$ & $\mathbf{91.01}$ & $\mathbf{94.82}$ & $\mathbf{97.07}$ & Key.Net + ref. & $\mathbf{0.18}$ & $\mathbf{0.74}$ & $\mathbf{3.41}$ & $\mathbf{83.26}$ & $\mathbf{89.70}$ & $\mathbf{94.99}$ \\ \midrule
		\multirowcell{8}{\textit{Relief 2}\\$31$ images} & SIFT & $0.16$ & $0.80$ & $3.74$ & $76.67$ & $86.48$ & $93.35$ & SURF & $0.05$ & $0.29$ & $1.41$ & $64.15$ & $82.25$ & $93.02$ \\
		& SIFT + ref. & $\mathbf{0.20}$ & $\mathbf{0.89}$ & $\mathbf{4.00}$ & $\mathbf{83.77}$ & $\mathbf{91.19}$ & $\mathbf{95.64}$ & SURF + ref. & $\mathbf{0.08}$ & $\mathbf{0.37}$ & $\mathbf{1.62}$ & $\mathbf{80.24}$ & $\mathbf{89.38}$ & $\mathbf{94.64}$ \\ \cmidrule{2-15}
		& D2-Net & $0.25$ & $1.48$ & $7.63$ & $46.03$ & $64.57$ & $84.66$ & R2D2 & $0.49$ & $2.10$ & $10.16$ & $74.70$ & $86.28$ & $94.43$ \\
		& D2-Net + ref. & $\mathbf{1.36}$ & $\mathbf{5.24}$ & $\mathbf{16.12}$ & $\mathbf{86.58}$ & $\mathbf{91.56}$ & $\mathbf{95.01}$ & R2D2 + ref. & $\mathbf{0.67}$ & $\mathbf{2.47}$ & $\mathbf{10.84}$ & $\mathbf{88.42}$ & $\mathbf{93.04}$ & $\mathbf{96.73}$ \\ \cmidrule{2-15}
		& SP & $0.32$ & $1.58$ & $7.80$ & $77.21$ & $88.20$ & $94.85$ & Key.Net & $0.11$ & $0.58$ & $3.00$ & $59.26$ & $76.71$ & $93.30$ \\
		& SP + ref. & $\mathbf{0.41}$ & $\mathbf{1.83}$ & $\mathbf{8.42}$ & $\mathbf{89.62}$ & $\mathbf{94.49}$ & $\mathbf{97.05}$ & Key.Net + ref. & $\mathbf{0.16}$ & $\mathbf{0.70}$ & $\mathbf{3.32}$ & $\mathbf{79.91}$ & $\mathbf{90.00}$ & $\mathbf{95.35}$ \\ \midrule
		\multirowcell{8}{\textit{Terrains}\\$42$ images} & SIFT & $0.44$ & $1.46$ & $4.60$ & $89.51$ & $93.47$ & $\mathbf{96.76}$ & SURF & $0.11$ & $0.46$ & $2.14$ & $70.22$ & $75.98$ & $80.49$ \\
		& SIFT + ref. & $\mathbf{0.50}$ & $\mathbf{1.60}$ & $\mathbf{5.01}$ & $\mathbf{90.14}$ & $\mathbf{93.81}$ & $96.29$ & SURF + ref. & $\mathbf{0.15}$ & $\mathbf{0.58}$ & $\mathbf{2.55}$ & $\mathbf{76.13}$ & $\mathbf{82.15}$ & $\mathbf{85.45}$ \\ \cmidrule{2-15}
		& D2-Net & $1.99$ & $5.59$ & $15.11$ & $72.96$ & $84.66$ & $92.26$ & R2D2 & $1.49$ & $4.87$ & $15.15$ & $77.45$ & $85.81$ & $92.74$ \\
		& D2-Net + ref. & $\mathbf{4.51}$ & $\mathbf{10.40}$ & $\mathbf{25.10}$ & $\mathbf{88.34}$ & $\mathbf{92.13}$ & $\mathbf{95.37}$ & R2D2 + ref. & $\mathbf{1.71}$ & $\mathbf{5.21}$ & $\mathbf{15.81}$ & $\mathbf{85.44}$ & $\mathbf{89.72}$ & $\mathbf{93.33}$ \\ \cmidrule{2-15}
		& SP & $2.07$ & $5.82$ & $17.89$ & $86.51$ & $93.60$ & $96.93$ & Key.Net & $0.51$ & $1.64$ & $4.77$ & $85.47$ & $92.35$ & $95.69$ \\
		& SP + ref. & $\mathbf{2.30}$ & $\mathbf{6.20}$ & $\mathbf{18.58}$ & $\mathbf{92.77}$ & $\mathbf{95.80}$ & $\mathbf{97.74}$ & Key.Net + ref. & $\mathbf{0.58}$ & $\mathbf{1.78}$ & $\mathbf{5.05}$ & $\mathbf{90.91}$ & $\mathbf{93.30}$ & $\mathbf{96.09}$ \\ \bottomrule
	\end{tabular}
\end{table}
\setlength{\tabcolsep}{\tabcolsepdefault}

\setlength{\tabcolsep}{1pt}
\begin{table}[p]
	\caption{{\bf ETH3D triangulation evaluation - Outdoors.} We report triangulation statistics on each outdoor dataset for methods with and without refinement.}
	\label{tab:eth3d-out-indep}
	\centering
	\tiny
	\begin{tabular}[t]{c | l | c c c | c c c | l | c c c | c c c}
		\toprule
		\multirow{2}{*}{Dataset} & \multirow{2}{*}{Method} & \multicolumn{3}{c |}{Comp. (\%)} & \multicolumn{3}{c |}{Accuracy (\%)} & \multirow{2}{*}{Method} & \multicolumn{3}{c |}{Comp. (\%)} & \multicolumn{3}{c}{Accuracy (\%)} \\
		& & $1$cm & $2$cm & $5$cm & $1$cm & $2$cm & $5$cm & & $1$cm & $2$cm & $5$cm & $1$cm & $2$cm & $5$cm \\ \midrule
		\multirowcell{8}{\textit{Courtyard}\\$38$ images} & SIFT & $0.08$ & $0.47$ & $3.72$ & $67.94$ & $81.80$ & $92.04$ & SURF & $0.06$ & $0.31$ & $1.88$ & $66.40$ & $80.04$ & $89.58$ \\
		& SIFT + ref. & $\mathbf{0.10}$ & $\mathbf{0.56}$ & $\mathbf{4.03}$ & $\mathbf{75.17}$ & $\mathbf{86.01}$ & $\mathbf{94.00}$ & SURF + ref. & $\mathbf{0.08}$ & $\mathbf{0.41}$ & $\mathbf{2.25}$ & $\mathbf{79.96}$ & $\mathbf{87.53}$ & $\mathbf{94.03}$ \\ \cmidrule{2-15}
		& D2-Net & $0.03$ & $0.24$ & $2.07$ & $22.63$ & $38.53$ & $61.33$ & R2D2 & $0.07$ & $0.37$ & $2.73$ & $45.72$ & $62.08$ & $79.61$ \\
		& D2-Net + ref. & $\mathbf{0.21}$ & $\mathbf{1.14}$ & $\mathbf{5.98}$ & $\mathbf{66.78}$ & $\mathbf{79.04}$ & $\mathbf{89.40}$ & R2D2 + ref. & $\mathbf{0.10}$ & $\mathbf{0.52}$ & $\mathbf{3.33}$ & $\mathbf{63.91}$ & $\mathbf{78.18}$ & $\mathbf{90.29}$ \\ \cmidrule{2-15}
		& SP & $0.13$ & $0.79$ & $5.04$ & $45.36$ & $60.61$ & $77.84$ & Key.Net & $0.02$ & $0.12$ & $0.83$ & $41.60$ & $62.78$ & $79.38$ \\
		& SP + ref. & $\mathbf{0.21}$ & $\mathbf{1.12}$ & $\mathbf{6.68}$ & $\mathbf{63.98}$ & $\mathbf{77.69}$ & $\mathbf{88.95}$ & Key.Net + ref. & $\mathbf{0.03}$ & $\mathbf{0.16}$ & $\mathbf{0.99}$ & $\mathbf{63.54}$ & $\mathbf{77.83}$ & $\mathbf{89.96}$ \\ \midrule
		\multirowcell{8}{\textit{Electro}\\$45$ images} & SIFT & $0.03$ & $0.15$ & $0.94$ & $63.76$ & $78.46$ & $88.84$ & SURF & $0.01$ & $0.07$ & $0.48$ & $47.54$ & $65.22$ & $81.48$ \\
		& SIFT + ref. & $\mathbf{0.03}$ & $\mathbf{0.18}$ & $\mathbf{1.05}$ & $\mathbf{65.82}$ & $\mathbf{79.19}$ & $\mathbf{90.11}$ & SURF + ref. & $\mathbf{0.02}$ & $\mathbf{0.11}$ & $\mathbf{0.68}$ & $\mathbf{62.75}$ & $\mathbf{75.20}$ & $\mathbf{87.06}$ \\ \cmidrule{2-15}
		& D2-Net & $0.03$ & $0.19$ & $1.50$ & $30.30$ & $45.29$ & $66.46$ & R2D2 & $0.12$ & $0.57$ & $3.66$ & $57.32$ & $73.33$ & $87.98$ \\
		& D2-Net + ref. & $\mathbf{0.19}$ & $\mathbf{0.95}$ & $\mathbf{4.99}$ & $\mathbf{68.36}$ & $\mathbf{79.57}$ & $\mathbf{89.56}$ & R2D2 + ref. & $\mathbf{0.17}$ & $\mathbf{0.72}$ & $\mathbf{4.00}$ & $\mathbf{70.96}$ & $\mathbf{82.32}$ & $\mathbf{91.46}$ \\ \cmidrule{2-15}
		& SP & $0.06$ & $0.34$ & $2.45$ & $60.66$ & $75.89$ & $89.26$ & Key.Net & $0.02$ & $0.11$ & $0.83$ & $45.09$ & $65.80$ & $82.31$ \\
		& SP + ref. & $\mathbf{0.09}$ & $\mathbf{0.44}$ & $\mathbf{2.77}$ & $\mathbf{76.96}$ & $\mathbf{87.29}$ & $\mathbf{93.75}$ & Key.Net + ref. & $\mathbf{0.03}$ & $\mathbf{0.17}$ & $\mathbf{1.01}$ & $\mathbf{65.93}$ & $\mathbf{81.83}$ & $\mathbf{91.56}$ \\ \midrule
		\multirowcell{8}{\textit{Facade}\\$76$ images} & SIFT & $0.06$ & $0.36$ & $3.08$ & $36.04$ & $52.10$ & $73.28$ & SURF & $0.05$ & $0.36$ & $3.18$ & $25.17$ & $41.25$ & $63.75$ \\
		& SIFT + ref. & $\mathbf{0.09}$ & $\mathbf{0.50}$ & $\mathbf{3.82}$ & $\mathbf{45.19}$ & $\mathbf{62.32}$ & $\mathbf{82.38}$ & SURF + ref. & $\mathbf{0.11}$ & $\mathbf{0.66}$ & $\mathbf{4.71}$ & $\mathbf{43.41}$ & $\mathbf{63.28}$ & $\mathbf{83.43}$ \\ \cmidrule{2-15}
		& D2-Net & $0.02$ & $0.18$ & $2.26$ & $7.56$ & $14.21$ & $29.90$ & R2D2 & $0.05$ & $0.28$ & $2.17$ & $25.07$ & $40.83$ & $64.42$ \\
		& D2-Net + ref. & $\mathbf{0.16}$ & $\mathbf{1.01}$ & $\mathbf{8.18}$ & $\mathbf{34.86}$ & $\mathbf{53.32}$ & $\mathbf{76.02}$ & R2D2 + ref. & $\mathbf{0.08}$ & $\mathbf{0.42}$ & $\mathbf{2.91}$ & $\mathbf{37.34}$ & $\mathbf{56.66}$ & $\mathbf{78.81}$ \\ \cmidrule{2-15}
		& SP & $0.07$ & $0.49$ & $4.95$ & $18.82$ & $32.21$ & $54.72$ & Key.Net & $0.01$ & $0.06$ & $0.58$ & $15.21$ & $25.12$ & $49.91$ \\
		& SP + ref. & $\mathbf{0.14}$ & $\mathbf{0.90}$ & $\mathbf{7.23}$ & $\mathbf{35.73}$ & $\mathbf{53.49}$ & $\mathbf{75.48}$ & Key.Net + ref. & $\mathbf{0.01}$ & $\mathbf{0.08}$ & $\mathbf{0.74}$ & $\mathbf{29.77}$ & $\mathbf{43.53}$ & $\mathbf{71.33}$ \\ \midrule
		\multirowcell{8}{\textit{Meadow}\\$15$ images} & SIFT & $0.01$ & $0.04$ & $0.35$ & $\mathbf{60.25}$ & $\mathbf{78.01}$ & $\mathbf{89.47}$ & SURF & $0.00$ & $0.01$ & $0.10$ & $30.77$ & $63.64$ & $\mathbf{84.62}$ \\
		& SIFT + ref. & $\mathbf{0.01}$ & $\mathbf{0.05}$ & $\mathbf{0.40}$ & $49.26$ & $73.95$ & $87.12$ & SURF + ref. & $\mathbf{0.00}$ & $\mathbf{0.02}$ & $\mathbf{0.13}$ & $\mathbf{55.56}$ & $\mathbf{65.31}$ & $80.70$ \\ \cmidrule{2-15}
		& D2-Net & $0.00$ & $0.03$ & $0.35$ & $21.89$ & $34.05$ & $57.35$ & R2D2 & $0.02$ & $0.14$ & $0.95$ & $50.23$ & $70.77$ & $87.10$ \\
		& D2-Net + ref. & $\mathbf{0.03}$ & $\mathbf{0.17}$ & $\mathbf{1.19}$ & $\mathbf{49.89}$ & $\mathbf{62.62}$ & $\mathbf{77.82}$ & R2D2 + ref. & $\mathbf{0.03}$ & $\mathbf{0.17}$ & $\mathbf{1.05}$ & $\mathbf{63.15}$ & $\mathbf{81.45}$ & $\mathbf{91.74}$ \\ \cmidrule{2-15}
		& SP & $0.02$ & $0.12$ & $1.06$ & $51.05$ & $68.91$ & $\mathbf{88.18}$ & Key.Net & $0.00$ & $0.01$ & $0.06$ & $46.67$ & $56.25$ & $64.71$ \\
		& SP + ref. & $\mathbf{0.03}$ & $\mathbf{0.16}$ & $\mathbf{1.21}$ & $\mathbf{66.67}$ & $\mathbf{78.85}$ & $88.02$ & Key.Net + ref. & $\mathbf{0.00}$ & $\mathbf{0.01}$ & $\mathbf{0.07}$ & $\mathbf{51.72}$ & $\mathbf{64.52}$ & $\mathbf{85.71}$ \\ \midrule
		\multirowcell{8}{\textit{Playground}\\$38$ images} & SIFT & $0.15$ & $0.80$ & $4.86$ & $66.57$ & $78.10$ & $90.58$ & SURF & $0.03$ & $0.18$ & $1.14$ & $57.25$ & $73.61$ & $86.05$ \\
		& SIFT + ref. & $\mathbf{0.18}$ & $\mathbf{0.91}$ & $\mathbf{5.27}$ & $\mathbf{70.70}$ & $\mathbf{81.76}$ & $\mathbf{91.73}$ & SURF + ref. & $\mathbf{0.06}$ & $\mathbf{0.27}$ & $\mathbf{1.57}$ & $\mathbf{74.60}$ & $\mathbf{83.76}$ & $\mathbf{92.70}$ \\ \cmidrule{2-15}
		& D2-Net & $0.05$ & $0.31$ & $2.42$ & $28.01$ & $46.88$ & $69.61$ & R2D2 & $0.26$ & $1.28$ & $7.71$ & $63.69$ & $78.08$ & $91.31$ \\
		& D2-Net + ref. & $\mathbf{0.46}$ & $\mathbf{2.01}$ & $\mathbf{8.19}$ & $\mathbf{71.63}$ & $\mathbf{83.73}$ & $\mathbf{93.60}$ & R2D2 + ref. & $\mathbf{0.37}$ & $\mathbf{1.58}$ & $\mathbf{8.29}$ & $\mathbf{78.03}$ & $\mathbf{88.76}$ & $\mathbf{96.53}$ \\ \cmidrule{2-15}
		& SP & $0.19$ & $0.97$ & $5.63$ & $59.09$ & $72.42$ & $86.01$ & Key.Net & $0.03$ & $0.15$ & $1.26$ & $45.61$ & $59.18$ & $80.10$ \\
		& SP + ref. & $\mathbf{0.28}$ & $\mathbf{1.29}$ & $\mathbf{6.83}$ & $\mathbf{70.30}$ & $\mathbf{79.84}$ & $\mathbf{90.09}$ & Key.Net + ref. & $\mathbf{0.04}$ & $\mathbf{0.22}$ & $\mathbf{1.54}$ & $\mathbf{64.06}$ & $\mathbf{78.65}$ & $\mathbf{91.32}$ \\ \midrule
		\multirowcell{8}{\textit{Terrace}\\$23$ images} & SIFT & $0.04$ & $0.20$ & $1.66$ & $55.32$ & $70.28$ & $83.23$ & SURF & $0.01$ & $0.06$ & $0.56$ & $38.13$ & $54.91$ & $72.80$ \\
		& SIFT + ref. & $\mathbf{0.05}$ & $\mathbf{0.26}$ & $\mathbf{1.93}$ & $\mathbf{63.53}$ & $\mathbf{78.10}$ & $\mathbf{88.41}$ & SURF + ref. & $\mathbf{0.02}$ & $\mathbf{0.10}$ & $\mathbf{0.75}$ & $\mathbf{61.00}$ & $\mathbf{72.97}$ & $\mathbf{84.68}$ \\ \cmidrule{2-15}
		& D2-Net & $0.02$ & $0.19$ & $2.21$ & $17.73$ & $31.53$ & $55.85$ & R2D2 & $0.12$ & $0.64$ & $4.46$ & $50.46$ & $69.33$ & $86.43$ \\
		& D2-Net + ref. & $\mathbf{0.22}$ & $\mathbf{1.24}$ & $\mathbf{8.26}$ & $\mathbf{62.92}$ & $\mathbf{75.78}$ & $\mathbf{87.34}$ & R2D2 + ref. & $\mathbf{0.19}$ & $\mathbf{0.84}$ & $\mathbf{4.90}$ & $\mathbf{69.73}$ & $\mathbf{81.24}$ & $\mathbf{91.69}$ \\ \cmidrule{2-15}
		& SP & $0.10$ & $0.56$ & $4.04$ & $63.03$ & $77.40$ & $88.71$ & Key.Net & $0.02$ & $0.10$ & $0.96$ & $41.31$ & $58.29$ & $77.42$ \\
		& SP + ref. & $\mathbf{0.14}$ & $\mathbf{0.72}$ & $\mathbf{4.75}$ & $\mathbf{77.76}$ & $\mathbf{87.87}$ & $\mathbf{93.94}$ & Key.Net + ref. & $\mathbf{0.03}$ & $\mathbf{0.14}$ & $\mathbf{1.13}$ & $\mathbf{58.70}$ & $\mathbf{70.11}$ & $\mathbf{83.46}$ \\ \bottomrule
	\end{tabular}
\end{table}
\setlength{\tabcolsep}{\tabcolsepdefault}

\clearpage \newpage
{
	\small
	\bibliographystyle{splncs04}
	\bibliography{shortstrings,paper}
}
\end{document}